\documentclass[10pt,twocolumn,letterpaper]{article}

\usepackage[pagenumbers]{cvpr}   %

\usepackage{pifont}
\usepackage{tabularx} %
\newcommand{\cmark}{\ding{51}}
\newcommand{\xmark}{\ding{55}}
\usepackage{fontawesome5}
\newcommand{\frozen}{\,{\scriptsize\textcolor[HTML]{3B72B5}{\faSnowflake}}}
\newcommand{\tuned}{\,{\scriptsize\textcolor[HTML]{D9622B}{\faFire}}}
\newcommand{\scratch}{\,{\scriptsize\textcolor[HTML]{2A8C7E}{\faSeedling}}}

\makeatletter
\renewcommand\paragraph{\@startsection{paragraph}{4}{\z@}{1.2ex \@plus 0.3ex \@minus 0.2ex}{-1em}{\normalfont\normalsize\bfseries}}
\makeatother

\newcommand{\SyncRmsMin}{0.9}
\newcommand{\SyncRmsMax}{10.7}
\newcommand{\SyncRmsMedian}{2.8}

\newcommand{\SyncInlierMedian}{94}
\newcommand{\SyncGlassesPpmMin}{5}
\newcommand{\SyncGlassesPpmMax}{50}
\newcommand{\SyncBoardPpmMin}{4}
\newcommand{\SyncBoardPpmMax}{30}

\newcommand{\VocabVerbs}{43}
\newcommand{\VocabVerbClasses}{17}
\newcommand{\VocabVerbKeptPct}{97.6}
\newcommand{\VocabMinParticipants}{6}
\newcommand{\VocabMinSeconds}{100}
\newcommand{\VocabNames}{97}
\newcommand{\VocabNamesMapped}{89}

\newcommand{\VocabEkNouns}{70}

\newcommand{\VocabNounClasses}{22}
\newcommand{\VocabActionClasses}{36}

\newcommand{\RNAAVF}{13.0}

\newcommand{\RNAAVFive}{62.8}

\newcommand{\RNAAonEgoMicIMUVFive}{+1.7 [+0.8, +2.7]}
\newcommand{\RNAAonEgoMicIMUVOne}{+2.7 [+1.4, +4.1]}

\newcommand{\RNAAonMicIMUVFive}{+2.5 [+1.4, +3.9]}
\newcommand{\RNAAonMicIMUVOne}{+4.0 [+2.3, +5.5]}

\newcommand{\RNAntEgo}{42.9}
\newcommand{\RNDetEgo}{3.2}

\newcommand{\RNDetWatch}{4.6}

\newcommand{\RNEgoAAGainVOneShort}{+6.8}

\newcommand{\RNEgoExoGainVFive}{$-$6.5 [$-$13.5, +0.1]}

\newcommand{\RNEgoExoGainVOneShort}{$-$6.3}

\newcommand{\RNEgoVF}{16.2}

\newcommand{\RNEgoVFive}{58.7}
\newcommand{\RNEgoVOne}{27.3}

\newcommand{\RNEgoWGGainActOne}{+4.1 [+0.9, +8.0]}
\newcommand{\RNEgoWGGainVFOne}{+4.5 [+3.5, +5.5]}

\newcommand{\RNEgoWGGainVFive}{+10.3 [+6.1, +14.2]}

\newcommand{\RNEgoWGGainVOneShort}{+11.6}
\newcommand{\RNEgoWGVF}{20.6}
\newcommand{\RNEgoWGVFive}{68.9}
\newcommand{\RNEgoWGVOne}{38.9}

\newcommand{\RNHandAAAll}{+4.3 [+2.1, +6.9]}

\newcommand{\RNHandAANone}{+7.7 [+3.5, +11.8]}

\newcommand{\RNHandWGAll}{+6.9 [+2.3, +12.1]}

\newcommand{\RNHandWGNone}{+12.7 [+7.6, +17.7]}

\newcommand{\RNIMUVF}{11.7}
\newcommand{\RNIMUVFive}{61.0}
\newcommand{\RNIMUVOne}{26.4}

\newcommand{\RNMicVF}{14.0}
\newcommand{\RNMicVFive}{61.7}
\newcommand{\RNMicVOne}{28.1}

\newcommand{\RNSegAllAOne}{28.7}

\newcommand{\RNSegWearVFive}{72.5}

\newcommand{\RSAAVFive}{65.2}

\newcommand{\RSAAonMicIMUVF}{+1.3 [+0.3, +2.3]}

\newcommand{\RSAAonMicIMUVOne}{+2.0 [+1.1, +2.9]}

\newcommand{\RSAntEgo}{51.6}

\newcommand{\RSDetExo}{13.7}

\newcommand{\RSEgoAAGAAGainVFive}{+4.0 [+2.8, +5.1]}

\newcommand{\RSEgoAAGainVFive}{+2.1 [+0.9, +3.3]}
\newcommand{\RSEgoAAGainVFiveShort}{+2.1}

\newcommand{\RSEgoAAGainVOneShort}{+2.0}

\newcommand{\RSEgoActOne}{63.4}

\newcommand{\RSEgoExoGainVFOne}{+1.9 [+0.8, +3.0]}

\newcommand{\RSEgoExoGainVFive}{+3.1 [+1.7, +4.6]}
\newcommand{\RSEgoExoGainVFiveShort}{+3.1}

\newcommand{\RSEgoExoGainVOneShort}{+3.2}

\newcommand{\RSEgoExoVFive}{73.8}

\newcommand{\RSEgoVF}{23.3}

\newcommand{\RSEgoVFive}{70.7}
\newcommand{\RSEgoVOne}{40.2}

\newcommand{\RSEgoWGGainVFOne}{+1.9 [+0.7, +2.8]}

\newcommand{\RSEgoWGGainVFive}{+5.8 [+4.2, +7.5]}

\newcommand{\RSEgoWGGainVOneShort}{+5.6}
\newcommand{\RSEgoWGVF}{25.2}
\newcommand{\RSEgoWGVFive}{76.5}
\newcommand{\RSEgoWGVOne}{45.8}

\newcommand{\RSHandAAAll}{+0.9 [+0.1, +2.0]}

\newcommand{\RSHandAANone}{+2.7 [+1.2, +4.4]}

\newcommand{\RSHandShareAll}{19}
\newcommand{\RSHandShareNone}{53}
\newcommand{\RSHandShareSome}{28}
\newcommand{\RSHandWGAll}{+3.7 [+2.1, +5.8]}

\newcommand{\RSHandWGNone}{+7.0 [+4.8, +9.3]}

\newcommand{\RSIMUVF}{13.8}
\newcommand{\RSIMUVFive}{63.9}
\newcommand{\RSIMUVOne}{29.6}

\newcommand{\RSMicVF}{18.9}
\newcommand{\RSMicVFive}{70.3}
\newcommand{\RSMicVOne}{38.9}

\newcommand{\RSSegAllAOne}{36.8}

\newcommand{\RSSegEgoVFive}{84.2}

\newcommand{\RSSegN}{3,491}

\newcommand{\RSSegWearVFive}{79.7}

\newcommand{\ResultsSegmentRowsMain}{%
Ego video & 45.5 & 84.2 & 39.8 & 36.2 & 39.3 & 79.4 & 31.6 & 28.0 \\
Ego + AA & 45.5 & 84.2 & \textbf{40.6} & 36.1 & 39.5 & 79.7 & 32.8 & 28.1 \\
Ego + all wearables & \textbf{45.9} & \textbf{84.3} & 40.0 & \textbf{36.8} & \textbf{40.8} & \textbf{80.0} & \textbf{33.4} & \textbf{28.7} \\
\midrule
Ego + exo (ref.) & 45.7 & 84.3 & 40.7 & 36.9 & 40.0 & 80.1 & 31.1 & 25.9 \\
\midrule
Wearables, no camera & 38.1 & 79.7 & 33.2 & 27.9 & 31.9 & 72.5 & 25.8 & 18.3 \\}
\newcommand{\ResultsDetectRowsMain}{%
Ego video & 12.2 & \textbf{51.6} & 3.2 & \textbf{42.9} \\
Watch, no camera & 8.4 & 42.6 & 4.6 & 37.2 \\
Ego + watch & \textbf{12.6} & 49.6 & \textbf{4.7} & 42.8 \\
Ego + watch + glasses & 12.1 & 49.7 & 3.6 & 41.6 \\
\midrule
Ego + exo (ref.) & 13.7 & 52.3 & 3.7 & 37.6 \\}

\newcommand{\SuppWindowRows}{%
\tbase Ego video & 40.2 & 23.3 & 60.3 & 27.3 & 16.2 & 51.7 \\
\addlinespace[2pt]
\tgrp{7}{Wearables alone, no camera} \\
Watch AA & 35.6 & 17.3 & 48.4 & 31.7 & 13.0 & 43.4 \\
Watch IMU & 29.6 & 13.8 & 41.8 & 26.4 & 11.7 & 38.5 \\
Watch mic & 38.9 & 18.9 & 51.0 & 28.1 & 14.0 & 43.4 \\
Watch mic + IMU & 41.1 & 19.9 & 53.0 & 33.6 & 15.6 & 47.4 \\
Whole watch & 43.1 & 21.2 & 55.2 & 37.6 & 16.8 & 49.0 \\
Glasses AA & 30.4 & 14.6 & 44.6 & 15.4 & 9.9 & 33.8 \\
Glasses mic & 26.2 & 14.2 & 42.6 & 21.5 & 11.3 & 37.7 \\
\addlinespace[2pt]
\tgrp{7}{Added to egocentric video} \\
+ watch AA & \textbf{42.1} & 23.8 & 59.8 & \textbf{34.1} & \textbf{18.2} & 53.6 \\
+ watch IMU & \textbf{42.8} & 23.3 & 59.7 & \textbf{31.9} & 16.9 & 51.9 \\
+ watch mic & \textbf{43.2} & 24.1 & 60.1 & \textbf{33.5} & \textbf{18.9} & \textbf{54.1} \\
+ watch mic + IMU & \textbf{44.9} & 24.2 & 59.7 & \textbf{36.9} & \textbf{19.7} & \textbf{54.4} \\
+ whole watch & \textbf{45.3} & 24.4 & 60.0 & \textbf{39.6} & \textbf{20.4} & \textbf{54.6} \\
+ glasses AA & \textbf{42.2} & 23.6 & 59.9 & \textbf{29.8} & \textbf{17.7} & 52.1 \\
+ glasses mic & 40.7 & 24.0 & 59.2 & \textbf{30.8} & \textbf{18.2} & 52.9 \\
\tours + watch + glasses & \textbf{45.8} & \textbf{25.2} & 59.7 & \textbf{38.9} & \textbf{20.6} & \textbf{54.9} \\
\addlinespace[2pt]
\tgrp{7}{Exocentric reference} \\
\tref{Exo alone} & \tref{38.6} & \tref{22.3} & \tref{59.9} & \tref{12.1} & \tref{8.0} & \tref{27.2} \\
\tref{+ exo} & \tref{43.3} & \tref{25.2} & \tref{63.0} & \tref{21.0} & \tref{13.7} & \tref{48.3} \\
\tref{+ exo + whole watch} & \tref{46.9} & \tref{26.3} & \tref{62.2} & \tref{32.6} & \tref{18.2} & \tref{53.7} \\}
\newcommand{\SuppVisibilityRows}{%
\tgrp{7}{(a) Verb top-5} \\
\tbase Ego video (accuracy) & 69.2 & 71.1 & 74.2 & 55.9 & 60.8 & 63.0 \\
+ watch AA & \tpos{+2.7} & \tpos{+1.7} & \tpos{+0.9} & \tpos{+7.7} & \tpos{+4.9} & \tpos{+4.3} \\
+ watch IMU & \tpos{+3.3} & \tpos{+1.4} & +0.4 & \tpos{+6.5} & \tpos{+4.2} & +2.4 \\
+ watch mic + IMU & \tpos{+6.2} & \tpos{+4.2} & \tpos{+2.0} & \tpos{+10.9} & \tpos{+6.7} & \tpos{+6.0} \\
+ whole watch & \tpos{+6.3} & \tpos{+4.2} & \tpos{+2.3} & \tpos{+12.6} & \tpos{+8.4} & \tpos{+7.6} \\
+ glasses AA & \tpos{+2.9} & \tpos{+2.5} & \tpos{+1.7} & \tpos{+4.2} & \tpos{+2.2} & \tpos{+2.6} \\
\tours + watch + glasses & \tpos{+7.0} & \tpos{+5.1} & \tpos{+3.7} & \tpos{+12.7} & \tpos{+8.0} & \tpos{+6.9} \\
\tref{+ exo} & \tref{+3.7} & \tref{+2.6} & \tref{+2.2} & \tref{$-$3.4} & \tref{$-$8.5} & \tref{$-$12.5} \\
\addlinespace[2pt]
\tgrp{7}{(b) Verb top-1} \\
\tbase Ego video (accuracy) & 36.9 & 43.6 & 44.0 & 24.5 & 29.0 & 32.2 \\
+ watch AA & \tpos{+2.6} & +1.5 & +0.9 & \tpos{+7.1} & \tpos{+7.1} & \tpos{+5.8} \\
+ watch IMU & \tpos{+2.9} & \tpos{+2.1} & +2.5 & \tpos{+4.5} & \tpos{+6.0} & \tpos{+3.3} \\
+ watch mic + IMU & \tpos{+5.8} & \tpos{+3.7} & \tpos{+3.6} & \tpos{+9.8} & \tpos{+9.9} & \tpos{+8.8} \\
+ whole watch & \tpos{+6.4} & \tpos{+3.9} & \tpos{+3.6} & \tpos{+12.8} & \tpos{+12.6} & \tpos{+10.4} \\
+ glasses AA & \tpos{+2.0} & \tpos{+1.8} & \tpos{+2.4} & \tpos{+2.5} & +2.0 & +3.3 \\
\tours + watch + glasses & \tpos{+6.6} & \tpos{+4.3} & \tpos{+4.9} & \tpos{+11.7} & \tpos{+11.6} & \tpos{+11.6} \\
\tref{+ exo} & \tref{+3.6} & \tref{+1.5} & \tref{+4.5} & \tref{$-$6.2} & \tref{$-$5.5} & \tref{$-$7.5} \\}
\newcommand{\SuppShuffleRows}{%
\tgrp{7}{(a) Verbs} \\
Whole watch, no camera & 21.2 & 18.1 & \tneg{$-$3.0} & 16.8 & 14.1 & \tneg{$-$2.6} \\
Ego + watch AA & 23.8 & 20.5 & \tneg{$-$3.3} & 18.2 & 14.7 & \tneg{$-$3.5} \\
Ego + whole watch & 24.4 & 23.4 & \tneg{$-$0.9} & 20.4 & 18.7 & \tneg{$-$1.7} \\
\tours Ego + watch + glasses & 25.2 & 23.3 & \tneg{$-$1.9} & 20.6 & 18.5 & \tneg{$-$2.1} \\
\addlinespace[2pt]
\tgrp{7}{(b) Activities} \\
Whole watch, no camera & 55.2 & 48.2 & \tneg{$-$7.0} & 49.0 & 41.8 & \tneg{$-$7.2} \\
Ego + watch AA & 59.8 & 53.2 & \tneg{$-$6.6} & 53.6 & 46.5 & \tneg{$-$7.1} \\
Ego + whole watch & 60.0 & 58.0 & \tneg{$-$2.0} & 54.6 & 52.2 & \tneg{$-$2.4} \\
\tours Ego + watch + glasses & 59.7 & 56.0 & \tneg{$-$3.7} & 54.9 & 51.2 & \tneg{$-$3.7} \\}
\newcommand{\SuppComponentRows}{%
\tgrp{7}{Watch active acoustics} \\
Probe on frozen features\frozen & 59.4 & 29.4 & 14.8 & 53.4 & 22.4 & 10.8 \\
Fine-tuned network\tuned & 64.5 & \textbf{36.4} & 16.2 & \textbf{64.8} & \textbf{33.6} & 12.3 \\
\tours Both, averaged (used)\frozen\tuned & \textbf{65.2} & 35.6 & \textbf{17.3} & 62.8 & 31.7 & \textbf{13.0} \\
\addlinespace[2pt]
\tgrp{7}{Watch IMU} \\
Fine-tuned network\tuned & \textbf{64.4} & \textbf{30.8} & 13.1 & 60.7 & \textbf{27.3} & 10.9 \\
\tours Both, averaged (used)\frozen\tuned & 63.9 & 29.6 & \textbf{13.8} & \textbf{61.0} & 26.4 & \textbf{11.7} \\
\addlinespace[2pt]
\tgrp{7}{Watch microphone} \\
Fine-tuned network\tuned & \textbf{70.5} & \textbf{39.6} & 16.9 & 58.8 & 27.0 & 11.7 \\
\tours Both, averaged (used)\frozen\tuned & 70.3 & 38.9 & \textbf{18.9} & \textbf{61.7} & \textbf{28.1} & \textbf{14.0} \\}
\newcommand{\SuppSegmentRows}{%
\tgrp{7}{Seen participants} \\
\tbase Ego video & 45.5 & 84.2 & 39.8 & 77.6 & 36.2 & 68.6 \\
+ AA (watch + glasses) & 45.5 & 84.2 & \textbf{40.6} & 77.9 & 36.1 & \textbf{68.8} \\
\tours + watch + glasses & \textbf{45.9} & \textbf{84.3} & 40.0 & \textbf{78.1} & \textbf{36.8} & 68.6 \\
\tref{+ exo} & \tref{45.7} & \tref{84.3} & \tref{40.7} & \tref{77.9} & \tref{36.9} & \tref{68.3} \\
Watch + glasses, no camera & 38.1 & 79.7 & 33.2 & 70.3 & 27.9 & 59.4 \\
AA alone, no camera & 34.8 & 76.3 & 27.8 & 65.1 & 23.4 & 52.8 \\
\tref{Exo alone} & \tref{37.6} & \tref{79.9} & \tref{34.8} & \tref{71.4} & \tref{29.6} & \tref{60.6} \\
\addlinespace[2pt]
\tgrp{7}{New participants} \\
\tbase Ego video & 39.3 & 79.4 & 31.6 & 69.6 & 28.0 & 58.4 \\
+ AA (watch + glasses) & 39.5 & 79.7 & 32.8 & 70.9 & 28.1 & 58.6 \\
\tours + watch + glasses & \textbf{40.8} & \textbf{80.0} & \textbf{33.4} & \textbf{71.0} & \textbf{28.7} & \textbf{60.1} \\
\tref{+ exo} & \tref{40.0} & \tref{80.1} & \tref{31.1} & \tref{70.8} & \tref{25.9} & \tref{58.4} \\
Watch + glasses, no camera & 31.9 & 72.5 & 25.8 & 59.3 & 18.3 & 49.9 \\
AA alone, no camera & 29.2 & 69.3 & 20.1 & 54.4 & 15.0 & 42.4 \\
\tref{Exo alone} & \tref{21.9} & \tref{62.8} & \tref{16.5} & \tref{52.0} & \tref{9.7} & \tref{34.8} \\}
\newcommand{\SuppDetectRows}{%
\tgrp{5}{(a) Five activities} \\
\tbase Ego video & 12.2\tsd{0.4} & 39.7\tsd{0.6} & 3.2\tsd{0.2} & 24.5\tsd{0.2} \\
Watch AA & 5.2\tsd{0.3} & 29.5\tsd{0.1} & 3.1\tsd{0.2} & 25.7\tsd{0.2} \\
Watch IMU & 5.3\tsd{0.1} & 29.7\tsd{0.7} & 2.7\tsd{0.2} & 27.2\tsd{0.5} \\
Whole watch & 8.4\tsd{0.2} & 35.2\tsd{0.3} & 4.6\tsd{0.4} & 29.6\tsd{1.5} \\
Ego + watch AA & 11.7\tsd{0.3} & 39.7\tsd{0.8} & 3.8\tsd{0.3} & 24.8\tsd{0.9} \\
Ego + watch mic + IMU & 12.2\tsd{0.1} & 39.7\tsd{0.8} & 3.8\tsd{0.4} & 25.1\tsd{0.9} \\
Ego + whole watch & 12.6\tsd{0.5} & 39.7\tsd{0.5} & 4.7\tsd{0.7} & 25.2\tsd{1.5} \\
\tours Ego + watch + glasses & 12.1\tsd{0.3} & 39.9\tsd{0.6} & 3.6\tsd{0.2} & 24.6\tsd{0.3} \\
\tref{Ego + exo} & \tref{13.7\tsd{0.5}} & \tref{41.1\tsd{0.3}} & \tref{3.7\tsd{0.3}} & \tref{25.5\tsd{1.0}} \\
\addlinespace[2pt]
\tgrp{5}{(b) 17 verbs} \\
\tbase Ego video & 4.5\tsd{0.3} & 16.3\tsd{0.1} & 1.8\tsd{0.2} & 10.1\tsd{0.5} \\
Watch AA & 1.8\tsd{0.0} & 10.6\tsd{0.1} & 1.0\tsd{0.1} & 8.3\tsd{0.3} \\
Watch IMU & 1.8\tsd{0.1} & 11.3\tsd{0.2} & 1.3\tsd{0.1} & 9.9\tsd{0.4} \\
Whole watch & 3.2\tsd{0.3} & 14.0\tsd{0.3} & 1.8\tsd{0.0} & 10.8\tsd{0.2} \\
Ego + watch AA & 4.9\tsd{0.1} & 16.9\tsd{0.4} & 1.9\tsd{0.1} & 10.6\tsd{0.3} \\
Ego + watch mic + IMU & 5.2\tsd{0.1} & 16.9\tsd{0.3} & 2.3\tsd{0.3} & 10.1\tsd{0.4} \\
Ego + whole watch & 5.1\tsd{0.1} & 17.1\tsd{0.1} & 2.3\tsd{0.3} & 11.0\tsd{0.1} \\
\tours Ego + watch + glasses & 5.2\tsd{0.3} & 17.7\tsd{0.3} & 2.4\tsd{0.1} & 11.2\tsd{0.5} \\
\tref{Ego + exo} & \tref{5.8\tsd{0.2}} & \tref{17.9\tsd{0.2}} & \tref{2.4\tsd{0.3}} & \tref{11.0\tsd{0.6}} \\}
\newcommand{\SuppAnticipationRows}{%
\tbase Ego video & 51.6 & 48.2 & 10.5 & 42.9 & 38.6 & 5.4 \\
Watch AA & 35.2 & -- & -- & 33.0 & -- & -- \\
Watch mic + IMU & 43.9 & -- & -- & 36.3 & -- & -- \\
Whole watch & 42.6 & -- & -- & 37.2 & -- & -- \\
Ego + watch AA & 49.5 & 46.6 & 9.9 & 40.1 & 38.9 & 6.2 \\
Ego + watch mic + IMU & 50.4 & 49.5 & 10.0 & 42.4 & 38.8 & 5.5 \\
Ego + whole watch & 49.6 & 48.6 & 10.4 & 42.8 & 39.8 & 6.3 \\
\tours Ego + watch + glasses & 49.7 & 47.9 & 9.8 & 41.6 & 39.1 & 5.7 \\
\tref{Ego + exo} & \tref{52.3} & \tref{48.8} & \tref{10.5} & \tref{37.6} & \tref{36.4} & \tref{3.1} \\}
\newcommand{\RSAntN}{3,464}

\makeatletter
\newcommand{\MB@get}[4]{\ifcsname #1/#2/#3/#4\endcsname\csname #1/#2/#3/#4\endcsname\else\errmessage{#1: no value #2/#3/#4}\fi}
\newcommand{\MBv}[3]{\MB@get{MBv}{#1}{#2}{#3}}
\newcommand{\MBg}[3]{\MB@get{MBg}{#1}{#2}{#3}}
\newcommand{\MBgl}[3]{\MB@get{MBgl}{#1}{#2}{#3}}
\newcommand{\MBgh}[3]{\MB@get{MBgh}{#1}{#2}{#3}}
\makeatother
\newcommand{\MBSeenWindows}{30{,}774}

\newcommand{\MBParticipants}{15}

\expandafter\def\csname MBv/seen/actsonic/top5\endcsname{70.6}
\expandafter\def\csname MBv/seen/actsonic/top1\endcsname{37.2}
\expandafter\def\csname MBv/seen/actsonic/macro_f1\endcsname{8.0}
\expandafter\def\csname MBv/seen/actsonic_board/top5\endcsname{72.2}
\expandafter\def\csname MBv/seen/actsonic_board/top1\endcsname{33.2}
\expandafter\def\csname MBv/seen/actsonic_board/macro_f1\endcsname{6.8}
\expandafter\def\csname MBv/seen/asf_board_probe/top5\endcsname{66.8}
\expandafter\def\csname MBv/seen/asf_board_probe/top1\endcsname{33.3}
\expandafter\def\csname MBv/seen/asf_board_probe/macro_f1\endcsname{12.5}
\expandafter\def\csname MBv/seen/asf_glasses_probe/top5\endcsname{67.7}
\expandafter\def\csname MBv/seen/asf_glasses_probe/top1\endcsname{37.2}
\expandafter\def\csname MBv/seen/asf_glasses_probe/macro_f1\endcsname{14.7}
\expandafter\def\csname MBv/seen/asf_probe/top5\endcsname{67.2}
\expandafter\def\csname MBv/seen/asf_probe/top1\endcsname{36.4}
\expandafter\def\csname MBv/seen/asf_probe/macro_f1\endcsname{14.3}
\expandafter\def\csname MBv/seen/avt_exo_probe/top5\endcsname{73.7}
\expandafter\def\csname MBv/seen/avt_exo_probe/top1\endcsname{46.0}
\expandafter\def\csname MBv/seen/avt_exo_probe/macro_f1\endcsname{19.0}
\expandafter\def\csname MBv/seen/avt_probe/top5\endcsname{71.8}
\expandafter\def\csname MBv/seen/avt_probe/top1\endcsname{43.2}
\expandafter\def\csname MBv/seen/avt_probe/macro_f1\endcsname{19.6}
\expandafter\def\csname MBv/seen/clip_l14_exo_probe/top5\endcsname{74.1}
\expandafter\def\csname MBv/seen/clip_l14_exo_probe/top1\endcsname{45.1}
\expandafter\def\csname MBv/seen/clip_l14_exo_probe/macro_f1\endcsname{19.4}
\expandafter\def\csname MBv/seen/clip_l14_probe/top5\endcsname{73.2}
\expandafter\def\csname MBv/seen/clip_l14_probe/top1\endcsname{43.3}
\expandafter\def\csname MBv/seen/clip_l14_probe/macro_f1\endcsname{19.1}
\expandafter\def\csname MBv/seen/deepconvlstm/top5\endcsname{70.3}
\expandafter\def\csname MBv/seen/deepconvlstm/top1\endcsname{33.4}
\expandafter\def\csname MBv/seen/deepconvlstm/macro_f1\endcsname{12.7}
\expandafter\def\csname MBv/seen/egovlp_exo_probe/top5\endcsname{73.7}
\expandafter\def\csname MBv/seen/egovlp_exo_probe/top1\endcsname{45.1}
\expandafter\def\csname MBv/seen/egovlp_exo_probe/macro_f1\endcsname{19.4}
\expandafter\def\csname MBv/seen/egovlp_probe/top5\endcsname{73.7}
\expandafter\def\csname MBv/seen/egovlp_probe/top1\endcsname{45.5}
\expandafter\def\csname MBv/seen/egovlp_probe/macro_f1\endcsname{22.7}
\expandafter\def\csname MBv/seen/fx_ego+exo/top5\endcsname{76.9}
\expandafter\def\csname MBv/seen/fx_ego+exo/top1\endcsname{47.3}
\expandafter\def\csname MBv/seen/fx_ego+exo/macro_f1\endcsname{22.0}
\expandafter\def\csname MBv/seen/fx_ego+exo+watch+glasses/top5\endcsname{78.3}
\expandafter\def\csname MBv/seen/fx_ego+exo+watch+glasses/top1\endcsname{46.1}
\expandafter\def\csname MBv/seen/fx_ego+exo+watch+glasses/macro_f1\endcsname{16.1}
\expandafter\def\csname MBv/seen/fx_ego+exo+watch_noecho+glasses/top5\endcsname{78.7}
\expandafter\def\csname MBv/seen/fx_ego+exo+watch_noecho+glasses/top1\endcsname{46.5}
\expandafter\def\csname MBv/seen/fx_ego+exo+watch_noecho+glasses/macro_f1\endcsname{16.8}
\expandafter\def\csname MBv/seen/fx_ego+g_echo/top5\endcsname{73.5}
\expandafter\def\csname MBv/seen/fx_ego+g_echo/top1\endcsname{39.7}
\expandafter\def\csname MBv/seen/fx_ego+g_echo/macro_f1\endcsname{12.6}
\expandafter\def\csname MBv/seen/fx_ego+glasses/top5\endcsname{73.8}
\expandafter\def\csname MBv/seen/fx_ego+glasses/top1\endcsname{41.0}
\expandafter\def\csname MBv/seen/fx_ego+glasses/macro_f1\endcsname{13.3}
\expandafter\def\csname MBv/seen/fx_ego+h_mic/top5\endcsname{73.6}
\expandafter\def\csname MBv/seen/fx_ego+h_mic/top1\endcsname{44.1}
\expandafter\def\csname MBv/seen/fx_ego+h_mic/macro_f1\endcsname{19.2}
\expandafter\def\csname MBv/seen/fx_ego+h_mic+watch/top5\endcsname{75.8}
\expandafter\def\csname MBv/seen/fx_ego+h_mic+watch/top1\endcsname{45.8}
\expandafter\def\csname MBv/seen/fx_ego+h_mic+watch/macro_f1\endcsname{20.1}
\expandafter\def\csname MBv/seen/fx_ego+h_mic+watch_noecho/top5\endcsname{76.3}
\expandafter\def\csname MBv/seen/fx_ego+h_mic+watch_noecho/top1\endcsname{45.5}
\expandafter\def\csname MBv/seen/fx_ego+h_mic+watch_noecho/macro_f1\endcsname{20.7}
\expandafter\def\csname MBv/seen/fx_ego+ib_imu/top5\endcsname{73.3}
\expandafter\def\csname MBv/seen/fx_ego+ib_imu/top1\endcsname{43.8}
\expandafter\def\csname MBv/seen/fx_ego+ib_imu/macro_f1\endcsname{19.0}
\expandafter\def\csname MBv/seen/fx_ego+w_echo/top5\endcsname{74.7}
\expandafter\def\csname MBv/seen/fx_ego+w_echo/top1\endcsname{42.8}
\expandafter\def\csname MBv/seen/fx_ego+w_echo/macro_f1\endcsname{18.1}
\expandafter\def\csname MBv/seen/fx_ego+w_echo+g_echo/top5\endcsname{74.3}
\expandafter\def\csname MBv/seen/fx_ego+w_echo+g_echo/top1\endcsname{40.5}
\expandafter\def\csname MBv/seen/fx_ego+w_echo+g_echo/macro_f1\endcsname{12.7}
\expandafter\def\csname MBv/seen/fx_ego+w_echo+w_mic+w_imu/top5\endcsname{76.1}
\expandafter\def\csname MBv/seen/fx_ego+w_echo+w_mic+w_imu/top1\endcsname{44.9}
\expandafter\def\csname MBv/seen/fx_ego+w_echo+w_mic+w_imu/macro_f1\endcsname{19.6}
\expandafter\def\csname MBv/seen/fx_ego+w_imu/top5\endcsname{74.2}
\expandafter\def\csname MBv/seen/fx_ego+w_imu/top1\endcsname{43.5}
\expandafter\def\csname MBv/seen/fx_ego+w_imu/macro_f1\endcsname{21.2}
\expandafter\def\csname MBv/seen/fx_ego+w_imu_dcl/top5\endcsname{75.4}
\expandafter\def\csname MBv/seen/fx_ego+w_imu_dcl/top1\endcsname{44.5}
\expandafter\def\csname MBv/seen/fx_ego+w_imu_dcl/macro_f1\endcsname{19.0}
\expandafter\def\csname MBv/seen/fx_ego+w_mic/top5\endcsname{73.8}
\expandafter\def\csname MBv/seen/fx_ego+w_mic/top1\endcsname{42.1}
\expandafter\def\csname MBv/seen/fx_ego+w_mic/macro_f1\endcsname{20.0}
\expandafter\def\csname MBv/seen/fx_ego+w_mic+w_imu/top5\endcsname{75.7}
\expandafter\def\csname MBv/seen/fx_ego+w_mic+w_imu/top1\endcsname{43.9}
\expandafter\def\csname MBv/seen/fx_ego+w_mic+w_imu/macro_f1\endcsname{21.6}
\expandafter\def\csname MBv/seen/fx_ego+watch/top5\endcsname{74.8}
\expandafter\def\csname MBv/seen/fx_ego+watch/top1\endcsname{44.1}
\expandafter\def\csname MBv/seen/fx_ego+watch/macro_f1\endcsname{20.3}
\expandafter\def\csname MBv/seen/fx_ego+watch+glasses/top5\endcsname{75.5}
\expandafter\def\csname MBv/seen/fx_ego+watch+glasses/top1\endcsname{42.1}
\expandafter\def\csname MBv/seen/fx_ego+watch+glasses/macro_f1\endcsname{14.0}
\expandafter\def\csname MBv/seen/fx_ego+watch_noecho/top5\endcsname{75.6}
\expandafter\def\csname MBv/seen/fx_ego+watch_noecho/top1\endcsname{44.2}
\expandafter\def\csname MBv/seen/fx_ego+watch_noecho/macro_f1\endcsname{21.2}
\expandafter\def\csname MBv/seen/fx_ego+watch_noecho+glasses/top5\endcsname{75.8}
\expandafter\def\csname MBv/seen/fx_ego+watch_noecho+glasses/top1\endcsname{42.8}
\expandafter\def\csname MBv/seen/fx_ego+watch_noecho+glasses/macro_f1\endcsname{14.6}
\expandafter\def\csname MBv/seen/fx_ego_ft+exo/top5\endcsname{77.2}
\expandafter\def\csname MBv/seen/fx_ego_ft+exo/top1\endcsname{48.7}
\expandafter\def\csname MBv/seen/fx_ego_ft+exo/macro_f1\endcsname{23.6}
\expandafter\def\csname MBv/seen/fx_ego_ft+w_echo/top5\endcsname{75.6}
\expandafter\def\csname MBv/seen/fx_ego_ft+w_echo/top1\endcsname{45.1}
\expandafter\def\csname MBv/seen/fx_ego_ft+w_echo/macro_f1\endcsname{19.4}
\expandafter\def\csname MBv/seen/fx_ego_ft+watch/top5\endcsname{75.6}
\expandafter\def\csname MBv/seen/fx_ego_ft+watch/top1\endcsname{46.2}
\expandafter\def\csname MBv/seen/fx_ego_ft+watch/macro_f1\endcsname{21.4}
\expandafter\def\csname MBv/seen/fx_ego_ft+watch_noecho/top5\endcsname{75.9}
\expandafter\def\csname MBv/seen/fx_ego_ft+watch_noecho/top1\endcsname{45.6}
\expandafter\def\csname MBv/seen/fx_ego_ft+watch_noecho/macro_f1\endcsname{21.7}
\expandafter\def\csname MBv/seen/fx_ib_ego+ib_imu/top5\endcsname{72.6}
\expandafter\def\csname MBv/seen/fx_ib_ego+ib_imu/top1\endcsname{43.7}
\expandafter\def\csname MBv/seen/fx_ib_ego+ib_imu/macro_f1\endcsname{18.0}
\expandafter\def\csname MBv/seen/fx_ib_ego+ib_mic+ib_imu/top5\endcsname{74.7}
\expandafter\def\csname MBv/seen/fx_ib_ego+ib_mic+ib_imu/top1\endcsname{45.6}
\expandafter\def\csname MBv/seen/fx_ib_ego+ib_mic+ib_imu/macro_f1\endcsname{18.7}
\expandafter\def\csname MBv/seen/fx_w_echo+w_mic+w_imu/top5\endcsname{74.4}
\expandafter\def\csname MBv/seen/fx_w_echo+w_mic+w_imu/top1\endcsname{38.5}
\expandafter\def\csname MBv/seen/fx_w_echo+w_mic+w_imu/macro_f1\endcsname{16.8}
\expandafter\def\csname MBv/seen/fx_watch+glasses/top5\endcsname{72.8}
\expandafter\def\csname MBv/seen/fx_watch+glasses/top1\endcsname{37.2}
\expandafter\def\csname MBv/seen/fx_watch+glasses/macro_f1\endcsname{10.9}
\expandafter\def\csname MBv/seen/fxc_ego+exo/top5\endcsname{76.8}
\expandafter\def\csname MBv/seen/fxc_ego+exo/top1\endcsname{47.3}
\expandafter\def\csname MBv/seen/fxc_ego+exo/macro_f1\endcsname{22.1}
\expandafter\def\csname MBv/seen/fxc_ego+g_echo/top5\endcsname{74.0}
\expandafter\def\csname MBv/seen/fxc_ego+g_echo/top1\endcsname{43.9}
\expandafter\def\csname MBv/seen/fxc_ego+g_echo/macro_f1\endcsname{15.9}
\expandafter\def\csname MBv/seen/fxc_ego+h_mic/top5\endcsname{73.5}
\expandafter\def\csname MBv/seen/fxc_ego+h_mic/top1\endcsname{44.2}
\expandafter\def\csname MBv/seen/fxc_ego+h_mic/macro_f1\endcsname{19.3}
\expandafter\def\csname MBv/seen/fxc_ego+w_echo/top5\endcsname{75.4}
\expandafter\def\csname MBv/seen/fxc_ego+w_echo/top1\endcsname{44.5}
\expandafter\def\csname MBv/seen/fxc_ego+w_echo/macro_f1\endcsname{18.8}
\expandafter\def\csname MBv/seen/fxc_ego+w_imu/top5\endcsname{74.3}
\expandafter\def\csname MBv/seen/fxc_ego+w_imu/top1\endcsname{43.7}
\expandafter\def\csname MBv/seen/fxc_ego+w_imu/macro_f1\endcsname{21.2}
\expandafter\def\csname MBv/seen/fxc_ego+w_mic/top5\endcsname{73.7}
\expandafter\def\csname MBv/seen/fxc_ego+w_mic/top1\endcsname{42.3}
\expandafter\def\csname MBv/seen/fxc_ego+w_mic/macro_f1\endcsname{19.8}
\expandafter\def\csname MBv/seen/fxc_ego+watch/top5\endcsname{74.9}
\expandafter\def\csname MBv/seen/fxc_ego+watch/top1\endcsname{44.3}
\expandafter\def\csname MBv/seen/fxc_ego+watch/macro_f1\endcsname{20.2}
\expandafter\def\csname MBv/seen/fxc_ego+watch+glasses/top5\endcsname{76.2}
\expandafter\def\csname MBv/seen/fxc_ego+watch+glasses/top1\endcsname{45.6}
\expandafter\def\csname MBv/seen/fxc_ego+watch+glasses/macro_f1\endcsname{17.9}
\expandafter\def\csname MBv/seen/fxc_ego+watch_noecho/top5\endcsname{75.6}
\expandafter\def\csname MBv/seen/fxc_ego+watch_noecho/top1\endcsname{44.4}
\expandafter\def\csname MBv/seen/fxc_ego+watch_noecho/macro_f1\endcsname{21.1}
\expandafter\def\csname MBv/seen/fxc_egovlp/top5\endcsname{73.7}
\expandafter\def\csname MBv/seen/fxc_egovlp/top1\endcsname{45.5}
\expandafter\def\csname MBv/seen/fxc_egovlp/macro_f1\endcsname{22.7}
\expandafter\def\csname MBv/seen/fxc_lavila/top5\endcsname{74.7}
\expandafter\def\csname MBv/seen/fxc_lavila/top1\endcsname{46.3}
\expandafter\def\csname MBv/seen/fxc_lavila/macro_f1\endcsname{23.3}
\expandafter\def\csname MBv/seen/fxc_lavila+egovlp/top5\endcsname{76.1}
\expandafter\def\csname MBv/seen/fxc_lavila+egovlp/top1\endcsname{47.7}
\expandafter\def\csname MBv/seen/fxc_lavila+egovlp/macro_f1\endcsname{24.0}
\expandafter\def\csname MBv/seen/fxc_omnivore/top5\endcsname{71.5}
\expandafter\def\csname MBv/seen/fxc_omnivore/top1\endcsname{42.3}
\expandafter\def\csname MBv/seen/fxc_omnivore/macro_f1\endcsname{19.2}
\expandafter\def\csname MBv/seen/gecho_stats_probe/top5\endcsname{62.5}
\expandafter\def\csname MBv/seen/gecho_stats_probe/top1\endcsname{28.6}
\expandafter\def\csname MBv/seen/gecho_stats_probe/macro_f1\endcsname{7.1}
\expandafter\def\csname MBv/seen/glassstats_probe/top5\endcsname{67.2}
\expandafter\def\csname MBv/seen/glassstats_probe/top1\endcsname{32.3}
\expandafter\def\csname MBv/seen/glassstats_probe/macro_f1\endcsname{9.9}
\expandafter\def\csname MBv/seen/gmel_stats_probe/top5\endcsname{64.8}
\expandafter\def\csname MBv/seen/gmel_stats_probe/top1\endcsname{31.4}
\expandafter\def\csname MBv/seen/gmel_stats_probe/macro_f1\endcsname{9.2}
\expandafter\def\csname MBv/seen/imagebind_audio_glasses_probe/top5\endcsname{68.1}
\expandafter\def\csname MBv/seen/imagebind_audio_glasses_probe/top1\endcsname{37.6}
\expandafter\def\csname MBv/seen/imagebind_audio_glasses_probe/macro_f1\endcsname{15.9}
\expandafter\def\csname MBv/seen/imagebind_audio_probe/top5\endcsname{67.7}
\expandafter\def\csname MBv/seen/imagebind_audio_probe/top1\endcsname{37.3}
\expandafter\def\csname MBv/seen/imagebind_audio_probe/macro_f1\endcsname{15.4}
\expandafter\def\csname MBv/seen/imagebind_imu_probe/top5\endcsname{64.2}
\expandafter\def\csname MBv/seen/imagebind_imu_probe/top1\endcsname{34.0}
\expandafter\def\csname MBv/seen/imagebind_imu_probe/macro_f1\endcsname{11.9}
\expandafter\def\csname MBv/seen/imagebind_vision_probe/top5\endcsname{70.9}
\expandafter\def\csname MBv/seen/imagebind_vision_probe/top1\endcsname{41.6}
\expandafter\def\csname MBv/seen/imagebind_vision_probe/macro_f1\endcsname{18.5}
\expandafter\def\csname MBv/seen/imu_stats_probe/top5\endcsname{65.9}
\expandafter\def\csname MBv/seen/imu_stats_probe/top1\endcsname{32.3}
\expandafter\def\csname MBv/seen/imu_stats_probe/macro_f1\endcsname{10.8}
\expandafter\def\csname MBv/seen/lavila_exo_probe/top5\endcsname{73.6}
\expandafter\def\csname MBv/seen/lavila_exo_probe/top1\endcsname{44.8}
\expandafter\def\csname MBv/seen/lavila_exo_probe/macro_f1\endcsname{20.9}
\expandafter\def\csname MBv/seen/lavila_probe/top5\endcsname{74.7}
\expandafter\def\csname MBv/seen/lavila_probe/top1\endcsname{46.3}
\expandafter\def\csname MBv/seen/lavila_probe/macro_f1\endcsname{23.3}
\expandafter\def\csname MBv/seen/omnivore_exo_probe/top5\endcsname{74.8}
\expandafter\def\csname MBv/seen/omnivore_exo_probe/top1\endcsname{44.1}
\expandafter\def\csname MBv/seen/omnivore_exo_probe/macro_f1\endcsname{20.8}
\expandafter\def\csname MBv/seen/omnivore_probe/top5\endcsname{71.5}
\expandafter\def\csname MBv/seen/omnivore_probe/top1\endcsname{42.3}
\expandafter\def\csname MBv/seen/omnivore_probe/macro_f1\endcsname{19.2}
\expandafter\def\csname MBv/seen/slowfast_exo_probe/top5\endcsname{72.9}
\expandafter\def\csname MBv/seen/slowfast_exo_probe/top1\endcsname{44.8}
\expandafter\def\csname MBv/seen/slowfast_exo_probe/macro_f1\endcsname{20.8}
\expandafter\def\csname MBv/seen/slowfast_probe/top5\endcsname{72.2}
\expandafter\def\csname MBv/seen/slowfast_probe/top1\endcsname{42.4}
\expandafter\def\csname MBv/seen/slowfast_probe/macro_f1\endcsname{19.4}
\expandafter\def\csname MBv/seen/ssast/top5\endcsname{70.3}
\expandafter\def\csname MBv/seen/ssast/top1\endcsname{37.7}
\expandafter\def\csname MBv/seen/ssast/macro_f1\endcsname{16.6}
\expandafter\def\csname MBv/seen/tim/top5\endcsname{77.4}
\expandafter\def\csname MBv/seen/tim/top1\endcsname{44.2}
\expandafter\def\csname MBv/seen/tim/macro_f1\endcsname{18.3}
\expandafter\def\csname MBv/seen/v3_echo_egofeat_wrist/top5\endcsname{76.2}
\expandafter\def\csname MBv/seen/v3_echo_egofeat_wrist/top1\endcsname{39.1}
\expandafter\def\csname MBv/seen/v3_echo_egofeat_wrist/macro_f1\endcsname{20.0}
\expandafter\def\csname MBv/seen/v3_echo_mel_imu_egofeat_wrist/top5\endcsname{76.5}
\expandafter\def\csname MBv/seen/v3_echo_mel_imu_egofeat_wrist/top1\endcsname{39.9}
\expandafter\def\csname MBv/seen/v3_echo_mel_imu_egofeat_wrist/macro_f1\endcsname{20.7}
\expandafter\def\csname MBv/seen/v3_echo_mel_imu_wrist/top5\endcsname{69.7}
\expandafter\def\csname MBv/seen/v3_echo_mel_imu_wrist/top1\endcsname{31.5}
\expandafter\def\csname MBv/seen/v3_echo_mel_imu_wrist/macro_f1\endcsname{14.9}
\expandafter\def\csname MBv/seen/v3_echo_wrist/top5\endcsname{71.2}
\expandafter\def\csname MBv/seen/v3_echo_wrist/top1\endcsname{31.6}
\expandafter\def\csname MBv/seen/v3_echo_wrist/macro_f1\endcsname{13.8}
\expandafter\def\csname MBv/seen/v3_egofeat_wrist/top5\endcsname{76.1}
\expandafter\def\csname MBv/seen/v3_egofeat_wrist/top1\endcsname{39.1}
\expandafter\def\csname MBv/seen/v3_egofeat_wrist/macro_f1\endcsname{20.1}
\expandafter\def\csname MBv/seen/v3_imu_egofeat_wrist/top5\endcsname{76.3}
\expandafter\def\csname MBv/seen/v3_imu_egofeat_wrist/top1\endcsname{39.6}
\expandafter\def\csname MBv/seen/v3_imu_egofeat_wrist/macro_f1\endcsname{20.5}
\expandafter\def\csname MBv/seen/v3_imu_wrist/top5\endcsname{64.7}
\expandafter\def\csname MBv/seen/v3_imu_wrist/top1\endcsname{23.5}
\expandafter\def\csname MBv/seen/v3_imu_wrist/macro_f1\endcsname{13.2}
\expandafter\def\csname MBv/seen/v3_mel_imu_egofeat_wrist/top5\endcsname{76.3}
\expandafter\def\csname MBv/seen/v3_mel_imu_egofeat_wrist/top1\endcsname{40.0}
\expandafter\def\csname MBv/seen/v3_mel_imu_egofeat_wrist/macro_f1\endcsname{21.0}
\expandafter\def\csname MBv/seen/v3_mel_imu_wrist/top5\endcsname{71.1}
\expandafter\def\csname MBv/seen/v3_mel_imu_wrist/top1\endcsname{32.4}
\expandafter\def\csname MBv/seen/v3_mel_imu_wrist/macro_f1\endcsname{16.7}
\expandafter\def\csname MBv/seen/v3_mel_wrist/top5\endcsname{62.4}
\expandafter\def\csname MBv/seen/v3_mel_wrist/top1\endcsname{24.2}
\expandafter\def\csname MBv/seen/v3_mel_wrist/macro_f1\endcsname{12.3}
\expandafter\def\csname MBv/seen/videomae_ft/top5\endcsname{73.1}
\expandafter\def\csname MBv/seen/videomae_ft/top1\endcsname{45.2}
\expandafter\def\csname MBv/seen/videomae_ft/macro_f1\endcsname{22.0}
\expandafter\def\csname MBv/seen/watchstats_noecho_probe/top5\endcsname{67.7}
\expandafter\def\csname MBv/seen/watchstats_noecho_probe/top1\endcsname{36.9}
\expandafter\def\csname MBv/seen/watchstats_noecho_probe/macro_f1\endcsname{13.2}
\expandafter\def\csname MBv/seen/watchstats_probe/top5\endcsname{69.1}
\expandafter\def\csname MBv/seen/watchstats_probe/top1\endcsname{38.1}
\expandafter\def\csname MBv/seen/watchstats_probe/macro_f1\endcsname{14.0}
\expandafter\def\csname MBv/seen/wecho_stats_probe/top5\endcsname{63.6}
\expandafter\def\csname MBv/seen/wecho_stats_probe/top1\endcsname{34.1}
\expandafter\def\csname MBv/seen/wecho_stats_probe/macro_f1\endcsname{9.0}
\expandafter\def\csname MBv/seen/wmel_stats_probe/top5\endcsname{63.1}
\expandafter\def\csname MBv/seen/wmel_stats_probe/top1\endcsname{32.3}
\expandafter\def\csname MBv/seen/wmel_stats_probe/macro_f1\endcsname{9.5}
\expandafter\def\csname MBv/seen/sec5-ego/top5\endcsname{70.7}
\expandafter\def\csname MBv/seen/sec5-ego/top1\endcsname{40.2}
\expandafter\def\csname MBv/seen/sec5-ego/macro_f1\endcsname{23.3}
\expandafter\def\csname MBv/seen/sec5-mic/top5\endcsname{70.3}
\expandafter\def\csname MBv/seen/sec5-mic/top1\endcsname{38.8}
\expandafter\def\csname MBv/seen/sec5-mic/macro_f1\endcsname{18.9}
\expandafter\def\csname MBv/seen/sec5-imu/top5\endcsname{63.9}
\expandafter\def\csname MBv/seen/sec5-imu/top1\endcsname{29.6}
\expandafter\def\csname MBv/seen/sec5-imu/macro_f1\endcsname{13.7}
\expandafter\def\csname MBv/seen/sec5-gmic/top5\endcsname{58.4}
\expandafter\def\csname MBv/seen/sec5-gmic/top1\endcsname{26.2}
\expandafter\def\csname MBv/seen/sec5-gmic/macro_f1\endcsname{14.2}
\expandafter\def\csname MBv/seen/sec5-exo/top5\endcsname{68.9}
\expandafter\def\csname MBv/seen/sec5-exo/top1\endcsname{38.6}
\expandafter\def\csname MBv/seen/sec5-exo/macro_f1\endcsname{22.3}
\expandafter\def\csname MBv/seen/sec5-ego-watch/top5\endcsname{75.7}
\expandafter\def\csname MBv/seen/sec5-ego-watch/top1\endcsname{45.3}
\expandafter\def\csname MBv/seen/sec5-ego-watch/macro_f1\endcsname{24.3}
\expandafter\def\csname MBv/seen/sec5-all/top5\endcsname{76.5}
\expandafter\def\csname MBv/seen/sec5-all/top1\endcsname{45.8}
\expandafter\def\csname MBv/seen/sec5-all/macro_f1\endcsname{25.2}
\expandafter\def\csname MBg/seen/lavila/ego2/top5\endcsname{+1.5}
\expandafter\def\csname MBgl/seen/lavila/ego2/top5\endcsname{+0.8}
\expandafter\def\csname MBgh/seen/lavila/ego2/top5\endcsname{+2.2}
\expandafter\def\csname MBg/seen/lavila/ego2/top1\endcsname{+0.4}
\expandafter\def\csname MBgl/seen/lavila/ego2/top1\endcsname{$-$2.2}
\expandafter\def\csname MBgh/seen/lavila/ego2/top1\endcsname{+2.8}
\expandafter\def\csname MBg/seen/lavila/ego2/macro_f1\endcsname{+3.0}
\expandafter\def\csname MBgl/seen/lavila/ego2/macro_f1\endcsname{+1.4}
\expandafter\def\csname MBgh/seen/lavila/ego2/macro_f1\endcsname{+4.4}
\expandafter\def\csname MBg/seen/sec5/aa/top5\endcsname{+2.1}
\expandafter\def\csname MBgl/seen/sec5/aa/top5\endcsname{+1.0}
\expandafter\def\csname MBgh/seen/sec5/aa/top5\endcsname{+3.4}
\expandafter\def\csname MBg/seen/sec5/aa/top1\endcsname{+2.0}
\expandafter\def\csname MBgl/seen/sec5/aa/top1\endcsname{+0.7}
\expandafter\def\csname MBgh/seen/sec5/aa/top1\endcsname{+3.3}
\expandafter\def\csname MBg/seen/sec5/aa/macro_f1\endcsname{+0.5}
\expandafter\def\csname MBgl/seen/sec5/aa/macro_f1\endcsname{$-$0.2}
\expandafter\def\csname MBgh/seen/sec5/aa/macro_f1\endcsname{+1.0}
\expandafter\def\csname MBg/seen/lavila/aa/top5\endcsname{+1.5}
\expandafter\def\csname MBgl/seen/lavila/aa/top5\endcsname{+0.6}
\expandafter\def\csname MBgh/seen/lavila/aa/top5\endcsname{+2.4}
\expandafter\def\csname MBg/seen/lavila/aa/top1\endcsname{+0.4}
\expandafter\def\csname MBgl/seen/lavila/aa/top1\endcsname{$-$1.0}
\expandafter\def\csname MBgh/seen/lavila/aa/top1\endcsname{+1.8}
\expandafter\def\csname MBg/seen/lavila/aa/macro_f1\endcsname{+1.1}
\expandafter\def\csname MBgl/seen/lavila/aa/macro_f1\endcsname{$-$0.1}
\expandafter\def\csname MBgh/seen/lavila/aa/macro_f1\endcsname{+1.9}
\expandafter\def\csname MBg/seen/sec5/micimu/top5\endcsname{+4.8}
\expandafter\def\csname MBgl/seen/sec5/micimu/top5\endcsname{+3.3}
\expandafter\def\csname MBgh/seen/sec5/micimu/top5\endcsname{+6.3}
\expandafter\def\csname MBg/seen/sec5/micimu/top1\endcsname{+4.8}
\expandafter\def\csname MBgl/seen/sec5/micimu/top1\endcsname{+2.9}
\expandafter\def\csname MBgh/seen/sec5/micimu/top1\endcsname{+6.3}
\expandafter\def\csname MBg/seen/sec5/micimu/macro_f1\endcsname{+0.9}
\expandafter\def\csname MBgl/seen/sec5/micimu/macro_f1\endcsname{$-$0.1}
\expandafter\def\csname MBgh/seen/sec5/micimu/macro_f1\endcsname{+1.8}
\expandafter\def\csname MBg/seen/lavila/micimu/top5\endcsname{+3.4}
\expandafter\def\csname MBgl/seen/lavila/micimu/top5\endcsname{+2.2}
\expandafter\def\csname MBgh/seen/lavila/micimu/top5\endcsname{+4.4}
\expandafter\def\csname MBg/seen/lavila/micimu/top1\endcsname{+1.9}
\expandafter\def\csname MBgl/seen/lavila/micimu/top1\endcsname{+0.3}
\expandafter\def\csname MBgh/seen/lavila/micimu/top1\endcsname{+3.2}
\expandafter\def\csname MBg/seen/lavila/micimu/macro_f1\endcsname{+1.2}
\expandafter\def\csname MBgl/seen/lavila/micimu/macro_f1\endcsname{$-$0.2}
\expandafter\def\csname MBgh/seen/lavila/micimu/macro_f1\endcsname{+2.4}
\expandafter\def\csname MBg/seen/sec5/watch/top5\endcsname{+5.0}
\expandafter\def\csname MBgl/seen/sec5/watch/top5\endcsname{+3.3}
\expandafter\def\csname MBgh/seen/sec5/watch/top5\endcsname{+6.6}
\expandafter\def\csname MBg/seen/sec5/watch/top1\endcsname{+5.2}
\expandafter\def\csname MBgl/seen/sec5/watch/top1\endcsname{+2.8}
\expandafter\def\csname MBgh/seen/sec5/watch/top1\endcsname{+7.3}
\expandafter\def\csname MBg/seen/sec5/watch/macro_f1\endcsname{+1.0}
\expandafter\def\csname MBgl/seen/sec5/watch/macro_f1\endcsname{$-$0.1}
\expandafter\def\csname MBgh/seen/sec5/watch/macro_f1\endcsname{+1.8}
\expandafter\def\csname MBg/seen/lavila/watch/top5\endcsname{+3.1}
\expandafter\def\csname MBgl/seen/lavila/watch/top5\endcsname{+1.7}
\expandafter\def\csname MBgh/seen/lavila/watch/top5\endcsname{+4.6}
\expandafter\def\csname MBg/seen/lavila/watch/top1\endcsname{+2.1}
\expandafter\def\csname MBgl/seen/lavila/watch/top1\endcsname{+0.6}
\expandafter\def\csname MBgh/seen/lavila/watch/top1\endcsname{+3.6}
\expandafter\def\csname MBg/seen/lavila/watch/macro_f1\endcsname{+1.6}
\expandafter\def\csname MBgl/seen/lavila/watch/macro_f1\endcsname{+0.1}
\expandafter\def\csname MBgh/seen/lavila/watch/macro_f1\endcsname{+2.5}
\expandafter\def\csname MBg/seen/sec5/gaa/top5\endcsname{+2.6}
\expandafter\def\csname MBgl/seen/sec5/gaa/top5\endcsname{+1.6}
\expandafter\def\csname MBgh/seen/sec5/gaa/top5\endcsname{+3.4}
\expandafter\def\csname MBg/seen/sec5/gaa/top1\endcsname{+2.0}
\expandafter\def\csname MBgl/seen/sec5/gaa/top1\endcsname{+0.9}
\expandafter\def\csname MBgh/seen/sec5/gaa/top1\endcsname{+2.9}
\expandafter\def\csname MBg/seen/sec5/gaa/macro_f1\endcsname{+0.3}
\expandafter\def\csname MBgl/seen/sec5/gaa/macro_f1\endcsname{$-$0.4}
\expandafter\def\csname MBgh/seen/sec5/gaa/macro_f1\endcsname{+1.0}
\expandafter\def\csname MBg/seen/lavila/gaa/top5\endcsname{+1.5}
\expandafter\def\csname MBgl/seen/lavila/gaa/top5\endcsname{+0.5}
\expandafter\def\csname MBgh/seen/lavila/gaa/top5\endcsname{+2.4}
\expandafter\def\csname MBg/seen/lavila/gaa/top1\endcsname{$-$0.4}
\expandafter\def\csname MBgl/seen/lavila/gaa/top1\endcsname{$-$2.0}
\expandafter\def\csname MBgh/seen/lavila/gaa/top1\endcsname{+1.1}
\expandafter\def\csname MBg/seen/lavila/gaa/macro_f1\endcsname{$-$0.5}
\expandafter\def\csname MBgl/seen/lavila/gaa/macro_f1\endcsname{$-$1.9}
\expandafter\def\csname MBgh/seen/lavila/gaa/macro_f1\endcsname{+0.6}
\expandafter\def\csname MBg/seen/sec5/all/top5\endcsname{+5.9}
\expandafter\def\csname MBgl/seen/sec5/all/top5\endcsname{+4.3}
\expandafter\def\csname MBgh/seen/sec5/all/top5\endcsname{+7.5}
\expandafter\def\csname MBg/seen/sec5/all/top1\endcsname{+5.6}
\expandafter\def\csname MBgl/seen/sec5/all/top1\endcsname{+3.3}
\expandafter\def\csname MBgh/seen/sec5/all/top1\endcsname{+7.3}
\expandafter\def\csname MBg/seen/sec5/all/macro_f1\endcsname{+1.9}
\expandafter\def\csname MBgl/seen/sec5/all/macro_f1\endcsname{+0.7}
\expandafter\def\csname MBgh/seen/sec5/all/macro_f1\endcsname{+2.8}
\expandafter\def\csname MBg/seen/lavila/all/top5\endcsname{+3.0}
\expandafter\def\csname MBgl/seen/lavila/all/top5\endcsname{+1.6}
\expandafter\def\csname MBgh/seen/lavila/all/top5\endcsname{+4.5}
\expandafter\def\csname MBg/seen/lavila/all/top1\endcsname{+1.4}
\expandafter\def\csname MBgl/seen/lavila/all/top1\endcsname{$-$0.6}
\expandafter\def\csname MBgh/seen/lavila/all/top1\endcsname{+3.2}
\expandafter\def\csname MBg/seen/lavila/all/macro_f1\endcsname{+1.8}
\expandafter\def\csname MBgl/seen/lavila/all/macro_f1\endcsname{+0.3}
\expandafter\def\csname MBgh/seen/lavila/all/macro_f1\endcsname{+3.0}
\expandafter\def\csname MBg/seen/sec5/exo/top5\endcsname{+3.1}
\expandafter\def\csname MBgl/seen/sec5/exo/top5\endcsname{+1.7}
\expandafter\def\csname MBgh/seen/sec5/exo/top5\endcsname{+4.7}
\expandafter\def\csname MBg/seen/sec5/exo/top1\endcsname{+3.2}
\expandafter\def\csname MBgl/seen/sec5/exo/top1\endcsname{+1.2}
\expandafter\def\csname MBgh/seen/sec5/exo/top1\endcsname{+5.3}
\expandafter\def\csname MBg/seen/sec5/exo/macro_f1\endcsname{+1.9}
\expandafter\def\csname MBgl/seen/sec5/exo/macro_f1\endcsname{+0.8}
\expandafter\def\csname MBgh/seen/sec5/exo/macro_f1\endcsname{+3.0}
\expandafter\def\csname MBg/seen/lavila/exo/top5\endcsname{+2.5}
\expandafter\def\csname MBgl/seen/lavila/exo/top5\endcsname{+1.3}
\expandafter\def\csname MBgh/seen/lavila/exo/top5\endcsname{+3.8}
\expandafter\def\csname MBg/seen/lavila/exo/top1\endcsname{+1.1}
\expandafter\def\csname MBgl/seen/lavila/exo/top1\endcsname{$-$1.3}
\expandafter\def\csname MBgh/seen/lavila/exo/top1\endcsname{+3.5}
\expandafter\def\csname MBg/seen/lavila/exo/macro_f1\endcsname{+3.4}
\expandafter\def\csname MBgl/seen/lavila/exo/macro_f1\endcsname{+1.4}
\expandafter\def\csname MBgh/seen/lavila/exo/macro_f1\endcsname{+5.1}
\expandafter\def\csname MBv/seen/prior/top5\endcsname{72.3}
\expandafter\def\csname MBv/seen/prior/top1\endcsname{25.5}
\expandafter\def\csname MBv/seen/prior/macro_f1\endcsname{2.4}
\expandafter\def\csname MBv/seen/sec5-ego-prior/top5\endcsname{77.7}
\expandafter\def\csname MBv/seen/sec5-ego-prior/top1\endcsname{48.0}
\expandafter\def\csname MBv/seen/sec5-ego-prior/macro_f1\endcsname{20.1}
\expandafter\def\csname MBg/seen/sec5/prior/top5\endcsname{+7.0}
\expandafter\def\csname MBgl/seen/sec5/prior/top5\endcsname{+5.0}
\expandafter\def\csname MBgh/seen/sec5/prior/top5\endcsname{+9.1}
\expandafter\def\csname MBg/seen/sec5/prior/top1\endcsname{+7.8}
\expandafter\def\csname MBgl/seen/sec5/prior/top1\endcsname{+4.3}
\expandafter\def\csname MBgh/seen/sec5/prior/top1\endcsname{+11.3}
\expandafter\def\csname MBg/seen/sec5/prior/macro_f1\endcsname{$-$3.2}
\expandafter\def\csname MBgl/seen/sec5/prior/macro_f1\endcsname{$-$4.9}
\expandafter\def\csname MBgh/seen/sec5/prior/macro_f1\endcsname{$-$1.6}
\expandafter\def\csname MBv/new/actsonic/top5\endcsname{70.8}
\expandafter\def\csname MBv/new/actsonic/top1\endcsname{33.6}
\expandafter\def\csname MBv/new/actsonic/macro_f1\endcsname{7.1}
\expandafter\def\csname MBv/new/actsonic_board/top5\endcsname{71.6}
\expandafter\def\csname MBv/new/actsonic_board/top1\endcsname{29.0}
\expandafter\def\csname MBv/new/actsonic_board/macro_f1\endcsname{5.6}
\expandafter\def\csname MBv/new/asf_board_probe/top5\endcsname{61.6}
\expandafter\def\csname MBv/new/asf_board_probe/top1\endcsname{26.6}
\expandafter\def\csname MBv/new/asf_board_probe/macro_f1\endcsname{9.5}
\expandafter\def\csname MBv/new/asf_glasses_probe/top5\endcsname{57.4}
\expandafter\def\csname MBv/new/asf_glasses_probe/top1\endcsname{27.0}
\expandafter\def\csname MBv/new/asf_glasses_probe/macro_f1\endcsname{10.3}
\expandafter\def\csname MBv/new/asf_probe/top5\endcsname{59.3}
\expandafter\def\csname MBv/new/asf_probe/top1\endcsname{26.9}
\expandafter\def\csname MBv/new/asf_probe/macro_f1\endcsname{10.7}
\expandafter\def\csname MBv/new/avt_exo_probe/top5\endcsname{37.2}
\expandafter\def\csname MBv/new/avt_exo_probe/top1\endcsname{10.2}
\expandafter\def\csname MBv/new/avt_exo_probe/macro_f1\endcsname{6.2}
\expandafter\def\csname MBv/new/avt_probe/top5\endcsname{61.0}
\expandafter\def\csname MBv/new/avt_probe/top1\endcsname{26.4}
\expandafter\def\csname MBv/new/avt_probe/macro_f1\endcsname{12.1}
\expandafter\def\csname MBv/new/clip_l14_exo_probe/top5\endcsname{42.5}
\expandafter\def\csname MBv/new/clip_l14_exo_probe/top1\endcsname{12.3}
\expandafter\def\csname MBv/new/clip_l14_exo_probe/macro_f1\endcsname{7.4}
\expandafter\def\csname MBv/new/clip_l14_probe/top5\endcsname{66.4}
\expandafter\def\csname MBv/new/clip_l14_probe/top1\endcsname{35.3}
\expandafter\def\csname MBv/new/clip_l14_probe/macro_f1\endcsname{14.6}
\expandafter\def\csname MBv/new/deepconvlstm/top5\endcsname{67.8}
\expandafter\def\csname MBv/new/deepconvlstm/top1\endcsname{27.1}
\expandafter\def\csname MBv/new/deepconvlstm/macro_f1\endcsname{9.8}
\expandafter\def\csname MBv/new/egovlp_exo_probe/top5\endcsname{46.3}
\expandafter\def\csname MBv/new/egovlp_exo_probe/top1\endcsname{13.6}
\expandafter\def\csname MBv/new/egovlp_exo_probe/macro_f1\endcsname{7.5}
\expandafter\def\csname MBv/new/egovlp_probe/top5\endcsname{68.3}
\expandafter\def\csname MBv/new/egovlp_probe/top1\endcsname{37.2}
\expandafter\def\csname MBv/new/egovlp_probe/macro_f1\endcsname{17.2}
\expandafter\def\csname MBv/new/fx_ego+exo/top5\endcsname{57.9}
\expandafter\def\csname MBv/new/fx_ego+exo/top1\endcsname{24.4}
\expandafter\def\csname MBv/new/fx_ego+exo/macro_f1\endcsname{11.6}
\expandafter\def\csname MBv/new/fx_ego+exo+watch+glasses/top5\endcsname{71.7}
\expandafter\def\csname MBv/new/fx_ego+exo+watch+glasses/top1\endcsname{33.9}
\expandafter\def\csname MBv/new/fx_ego+exo+watch+glasses/macro_f1\endcsname{10.8}
\expandafter\def\csname MBv/new/fx_ego+exo+watch_noecho+glasses/top5\endcsname{71.2}
\expandafter\def\csname MBv/new/fx_ego+exo+watch_noecho+glasses/top1\endcsname{33.4}
\expandafter\def\csname MBv/new/fx_ego+exo+watch_noecho+glasses/macro_f1\endcsname{11.1}
\expandafter\def\csname MBv/new/fx_ego+g_echo/top5\endcsname{68.8}
\expandafter\def\csname MBv/new/fx_ego+g_echo/top1\endcsname{32.2}
\expandafter\def\csname MBv/new/fx_ego+g_echo/macro_f1\endcsname{9.2}
\expandafter\def\csname MBv/new/fx_ego+glasses/top5\endcsname{69.5}
\expandafter\def\csname MBv/new/fx_ego+glasses/top1\endcsname{33.0}
\expandafter\def\csname MBv/new/fx_ego+glasses/macro_f1\endcsname{9.5}
\expandafter\def\csname MBv/new/fx_ego+h_mic/top5\endcsname{65.1}
\expandafter\def\csname MBv/new/fx_ego+h_mic/top1\endcsname{35.1}
\expandafter\def\csname MBv/new/fx_ego+h_mic/macro_f1\endcsname{14.6}
\expandafter\def\csname MBv/new/fx_ego+h_mic+watch/top5\endcsname{70.2}
\expandafter\def\csname MBv/new/fx_ego+h_mic+watch/top1\endcsname{37.4}
\expandafter\def\csname MBv/new/fx_ego+h_mic+watch/macro_f1\endcsname{14.6}
\expandafter\def\csname MBv/new/fx_ego+h_mic+watch_noecho/top5\endcsname{68.6}
\expandafter\def\csname MBv/new/fx_ego+h_mic+watch_noecho/top1\endcsname{36.9}
\expandafter\def\csname MBv/new/fx_ego+h_mic+watch_noecho/macro_f1\endcsname{15.4}
\expandafter\def\csname MBv/new/fx_ego+ib_imu/top5\endcsname{67.4}
\expandafter\def\csname MBv/new/fx_ego+ib_imu/top1\endcsname{36.8}
\expandafter\def\csname MBv/new/fx_ego+ib_imu/macro_f1\endcsname{14.1}
\expandafter\def\csname MBv/new/fx_ego+w_echo/top5\endcsname{70.2}
\expandafter\def\csname MBv/new/fx_ego+w_echo/top1\endcsname{34.9}
\expandafter\def\csname MBv/new/fx_ego+w_echo/macro_f1\endcsname{13.6}
\expandafter\def\csname MBv/new/fx_ego+w_echo+g_echo/top5\endcsname{71.2}
\expandafter\def\csname MBv/new/fx_ego+w_echo+g_echo/top1\endcsname{33.3}
\expandafter\def\csname MBv/new/fx_ego+w_echo+g_echo/macro_f1\endcsname{9.5}
\expandafter\def\csname MBv/new/fx_ego+w_echo+w_mic+w_imu/top5\endcsname{71.5}
\expandafter\def\csname MBv/new/fx_ego+w_echo+w_mic+w_imu/top1\endcsname{36.7}
\expandafter\def\csname MBv/new/fx_ego+w_echo+w_mic+w_imu/macro_f1\endcsname{14.2}
\expandafter\def\csname MBv/new/fx_ego+w_imu/top5\endcsname{67.4}
\expandafter\def\csname MBv/new/fx_ego+w_imu/top1\endcsname{34.8}
\expandafter\def\csname MBv/new/fx_ego+w_imu/macro_f1\endcsname{15.1}
\expandafter\def\csname MBv/new/fx_ego+w_imu_dcl/top5\endcsname{70.4}
\expandafter\def\csname MBv/new/fx_ego+w_imu_dcl/top1\endcsname{36.4}
\expandafter\def\csname MBv/new/fx_ego+w_imu_dcl/macro_f1\endcsname{13.7}
\expandafter\def\csname MBv/new/fx_ego+w_mic/top5\endcsname{66.2}
\expandafter\def\csname MBv/new/fx_ego+w_mic/top1\endcsname{33.7}
\expandafter\def\csname MBv/new/fx_ego+w_mic/macro_f1\endcsname{14.0}
\expandafter\def\csname MBv/new/fx_ego+w_mic+w_imu/top5\endcsname{69.1}
\expandafter\def\csname MBv/new/fx_ego+w_mic+w_imu/top1\endcsname{35.6}
\expandafter\def\csname MBv/new/fx_ego+w_mic+w_imu/macro_f1\endcsname{15.1}
\expandafter\def\csname MBv/new/fx_ego+watch/top5\endcsname{70.5}
\expandafter\def\csname MBv/new/fx_ego+watch/top1\endcsname{36.7}
\expandafter\def\csname MBv/new/fx_ego+watch/macro_f1\endcsname{14.5}
\expandafter\def\csname MBv/new/fx_ego+watch+glasses/top5\endcsname{71.6}
\expandafter\def\csname MBv/new/fx_ego+watch+glasses/top1\endcsname{34.4}
\expandafter\def\csname MBv/new/fx_ego+watch+glasses/macro_f1\endcsname{9.8}
\expandafter\def\csname MBv/new/fx_ego+watch_noecho/top5\endcsname{68.4}
\expandafter\def\csname MBv/new/fx_ego+watch_noecho/top1\endcsname{35.8}
\expandafter\def\csname MBv/new/fx_ego+watch_noecho/macro_f1\endcsname{15.2}
\expandafter\def\csname MBv/new/fx_ego+watch_noecho+glasses/top5\endcsname{71.5}
\expandafter\def\csname MBv/new/fx_ego+watch_noecho+glasses/top1\endcsname{34.0}
\expandafter\def\csname MBv/new/fx_ego+watch_noecho+glasses/macro_f1\endcsname{10.2}
\expandafter\def\csname MBv/new/fx_ego_ft+exo/top5\endcsname{55.0}
\expandafter\def\csname MBv/new/fx_ego_ft+exo/top1\endcsname{20.6}
\expandafter\def\csname MBv/new/fx_ego_ft+exo/macro_f1\endcsname{10.4}
\expandafter\def\csname MBv/new/fx_ego_ft+w_echo/top5\endcsname{70.0}
\expandafter\def\csname MBv/new/fx_ego_ft+w_echo/top1\endcsname{32.2}
\expandafter\def\csname MBv/new/fx_ego_ft+w_echo/macro_f1\endcsname{13.0}
\expandafter\def\csname MBv/new/fx_ego_ft+watch/top5\endcsname{70.4}
\expandafter\def\csname MBv/new/fx_ego_ft+watch/top1\endcsname{34.3}
\expandafter\def\csname MBv/new/fx_ego_ft+watch/macro_f1\endcsname{14.2}
\expandafter\def\csname MBv/new/fx_ego_ft+watch_noecho/top5\endcsname{69.2}
\expandafter\def\csname MBv/new/fx_ego_ft+watch_noecho/top1\endcsname{32.1}
\expandafter\def\csname MBv/new/fx_ego_ft+watch_noecho/macro_f1\endcsname{14.5}
\expandafter\def\csname MBv/new/fx_ib_ego+ib_imu/top5\endcsname{64.0}
\expandafter\def\csname MBv/new/fx_ib_ego+ib_imu/top1\endcsname{34.9}
\expandafter\def\csname MBv/new/fx_ib_ego+ib_imu/macro_f1\endcsname{12.9}
\expandafter\def\csname MBv/new/fx_ib_ego+ib_mic+ib_imu/top5\endcsname{68.6}
\expandafter\def\csname MBv/new/fx_ib_ego+ib_mic+ib_imu/top1\endcsname{38.5}
\expandafter\def\csname MBv/new/fx_ib_ego+ib_mic+ib_imu/macro_f1\endcsname{14.6}
\expandafter\def\csname MBv/new/fx_w_echo+w_mic+w_imu/top5\endcsname{69.4}
\expandafter\def\csname MBv/new/fx_w_echo+w_mic+w_imu/top1\endcsname{30.0}
\expandafter\def\csname MBv/new/fx_w_echo+w_mic+w_imu/macro_f1\endcsname{12.1}
\expandafter\def\csname MBv/new/fx_watch+glasses/top5\endcsname{70.2}
\expandafter\def\csname MBv/new/fx_watch+glasses/top1\endcsname{30.5}
\expandafter\def\csname MBv/new/fx_watch+glasses/macro_f1\endcsname{8.0}
\expandafter\def\csname MBv/new/fxc_ego+exo/top5\endcsname{64.2}
\expandafter\def\csname MBv/new/fxc_ego+exo/top1\endcsname{33.8}
\expandafter\def\csname MBv/new/fxc_ego+exo/macro_f1\endcsname{14.2}
\expandafter\def\csname MBv/new/fxc_ego+g_echo/top5\endcsname{68.3}
\expandafter\def\csname MBv/new/fxc_ego+g_echo/top1\endcsname{35.7}
\expandafter\def\csname MBv/new/fxc_ego+g_echo/macro_f1\endcsname{11.7}
\expandafter\def\csname MBv/new/fxc_ego+h_mic/top5\endcsname{65.2}
\expandafter\def\csname MBv/new/fxc_ego+h_mic/top1\endcsname{35.1}
\expandafter\def\csname MBv/new/fxc_ego+h_mic/macro_f1\endcsname{14.6}
\expandafter\def\csname MBv/new/fxc_ego+w_echo/top5\endcsname{69.7}
\expandafter\def\csname MBv/new/fxc_ego+w_echo/top1\endcsname{36.6}
\expandafter\def\csname MBv/new/fxc_ego+w_echo/macro_f1\endcsname{14.0}
\expandafter\def\csname MBv/new/fxc_ego+w_imu/top5\endcsname{67.3}
\expandafter\def\csname MBv/new/fxc_ego+w_imu/top1\endcsname{34.7}
\expandafter\def\csname MBv/new/fxc_ego+w_imu/macro_f1\endcsname{15.1}
\expandafter\def\csname MBv/new/fxc_ego+w_mic/top5\endcsname{66.3}
\expandafter\def\csname MBv/new/fxc_ego+w_mic/top1\endcsname{34.3}
\expandafter\def\csname MBv/new/fxc_ego+w_mic/macro_f1\endcsname{14.0}
\expandafter\def\csname MBv/new/fxc_ego+watch/top5\endcsname{70.4}
\expandafter\def\csname MBv/new/fxc_ego+watch/top1\endcsname{36.9}
\expandafter\def\csname MBv/new/fxc_ego+watch/macro_f1\endcsname{14.5}
\expandafter\def\csname MBv/new/fxc_ego+watch+glasses/top5\endcsname{72.0}
\expandafter\def\csname MBv/new/fxc_ego+watch+glasses/top1\endcsname{38.2}
\expandafter\def\csname MBv/new/fxc_ego+watch+glasses/macro_f1\endcsname{12.8}
\expandafter\def\csname MBv/new/fxc_ego+watch_noecho/top5\endcsname{68.4}
\expandafter\def\csname MBv/new/fxc_ego+watch_noecho/top1\endcsname{35.9}
\expandafter\def\csname MBv/new/fxc_ego+watch_noecho/macro_f1\endcsname{15.1}
\expandafter\def\csname MBv/new/fxc_egovlp/top5\endcsname{68.3}
\expandafter\def\csname MBv/new/fxc_egovlp/top1\endcsname{37.2}
\expandafter\def\csname MBv/new/fxc_egovlp/macro_f1\endcsname{17.2}
\expandafter\def\csname MBv/new/fxc_lavila/top5\endcsname{69.9}
\expandafter\def\csname MBv/new/fxc_lavila/top1\endcsname{39.1}
\expandafter\def\csname MBv/new/fxc_lavila/macro_f1\endcsname{17.1}
\expandafter\def\csname MBv/new/fxc_lavila+egovlp/top5\endcsname{71.9}
\expandafter\def\csname MBv/new/fxc_lavila+egovlp/top1\endcsname{42.1}
\expandafter\def\csname MBv/new/fxc_lavila+egovlp/macro_f1\endcsname{18.1}
\expandafter\def\csname MBv/new/fxc_omnivore/top5\endcsname{64.0}
\expandafter\def\csname MBv/new/fxc_omnivore/top1\endcsname{33.9}
\expandafter\def\csname MBv/new/fxc_omnivore/macro_f1\endcsname{14.2}
\expandafter\def\csname MBv/new/gecho_stats_probe/top5\endcsname{61.9}
\expandafter\def\csname MBv/new/gecho_stats_probe/top1\endcsname{26.9}
\expandafter\def\csname MBv/new/gecho_stats_probe/macro_f1\endcsname{6.4}
\expandafter\def\csname MBv/new/glassstats_probe/top5\endcsname{62.3}
\expandafter\def\csname MBv/new/glassstats_probe/top1\endcsname{24.4}
\expandafter\def\csname MBv/new/glassstats_probe/macro_f1\endcsname{6.9}
\expandafter\def\csname MBv/new/gmel_stats_probe/top5\endcsname{62.6}
\expandafter\def\csname MBv/new/gmel_stats_probe/top1\endcsname{25.6}
\expandafter\def\csname MBv/new/gmel_stats_probe/macro_f1\endcsname{6.9}
\expandafter\def\csname MBv/new/imagebind_audio_glasses_probe/top5\endcsname{61.8}
\expandafter\def\csname MBv/new/imagebind_audio_glasses_probe/top1\endcsname{30.2}
\expandafter\def\csname MBv/new/imagebind_audio_glasses_probe/macro_f1\endcsname{12.3}
\expandafter\def\csname MBv/new/imagebind_audio_probe/top5\endcsname{61.9}
\expandafter\def\csname MBv/new/imagebind_audio_probe/top1\endcsname{30.8}
\expandafter\def\csname MBv/new/imagebind_audio_probe/macro_f1\endcsname{12.4}
\expandafter\def\csname MBv/new/imagebind_imu_probe/top5\endcsname{61.3}
\expandafter\def\csname MBv/new/imagebind_imu_probe/top1\endcsname{29.7}
\expandafter\def\csname MBv/new/imagebind_imu_probe/macro_f1\endcsname{9.7}
\expandafter\def\csname MBv/new/imagebind_vision_probe/top5\endcsname{59.9}
\expandafter\def\csname MBv/new/imagebind_vision_probe/top1\endcsname{30.6}
\expandafter\def\csname MBv/new/imagebind_vision_probe/macro_f1\endcsname{13.2}
\expandafter\def\csname MBv/new/imu_stats_probe/top5\endcsname{64.9}
\expandafter\def\csname MBv/new/imu_stats_probe/top1\endcsname{30.6}
\expandafter\def\csname MBv/new/imu_stats_probe/macro_f1\endcsname{9.6}
\expandafter\def\csname MBv/new/lavila_exo_probe/top5\endcsname{43.0}
\expandafter\def\csname MBv/new/lavila_exo_probe/top1\endcsname{14.0}
\expandafter\def\csname MBv/new/lavila_exo_probe/macro_f1\endcsname{7.9}
\expandafter\def\csname MBv/new/lavila_probe/top5\endcsname{69.9}
\expandafter\def\csname MBv/new/lavila_probe/top1\endcsname{39.1}
\expandafter\def\csname MBv/new/lavila_probe/macro_f1\endcsname{17.1}
\expandafter\def\csname MBv/new/omnivore_exo_probe/top5\endcsname{42.3}
\expandafter\def\csname MBv/new/omnivore_exo_probe/top1\endcsname{12.5}
\expandafter\def\csname MBv/new/omnivore_exo_probe/macro_f1\endcsname{7.3}
\expandafter\def\csname MBv/new/omnivore_probe/top5\endcsname{64.0}
\expandafter\def\csname MBv/new/omnivore_probe/top1\endcsname{33.9}
\expandafter\def\csname MBv/new/omnivore_probe/macro_f1\endcsname{14.2}
\expandafter\def\csname MBv/new/slowfast_exo_probe/top5\endcsname{38.9}
\expandafter\def\csname MBv/new/slowfast_exo_probe/top1\endcsname{10.7}
\expandafter\def\csname MBv/new/slowfast_exo_probe/macro_f1\endcsname{6.1}
\expandafter\def\csname MBv/new/slowfast_probe/top5\endcsname{63.4}
\expandafter\def\csname MBv/new/slowfast_probe/top1\endcsname{30.0}
\expandafter\def\csname MBv/new/slowfast_probe/macro_f1\endcsname{14.2}
\expandafter\def\csname MBv/new/ssast/top5\endcsname{56.7}
\expandafter\def\csname MBv/new/ssast/top1\endcsname{21.2}
\expandafter\def\csname MBv/new/ssast/macro_f1\endcsname{9.4}
\expandafter\def\csname MBv/new/tim/top5\endcsname{71.8}
\expandafter\def\csname MBv/new/tim/top1\endcsname{35.7}
\expandafter\def\csname MBv/new/tim/macro_f1\endcsname{12.8}
\expandafter\def\csname MBv/new/v3_echo_egofeat_wrist/top5\endcsname{71.5}
\expandafter\def\csname MBv/new/v3_echo_egofeat_wrist/top1\endcsname{32.1}
\expandafter\def\csname MBv/new/v3_echo_egofeat_wrist/macro_f1\endcsname{15.0}
\expandafter\def\csname MBv/new/v3_echo_mel_imu_egofeat_wrist/top5\endcsname{71.6}
\expandafter\def\csname MBv/new/v3_echo_mel_imu_egofeat_wrist/top1\endcsname{32.5}
\expandafter\def\csname MBv/new/v3_echo_mel_imu_egofeat_wrist/macro_f1\endcsname{14.9}
\expandafter\def\csname MBv/new/v3_echo_mel_imu_wrist/top5\endcsname{66.4}
\expandafter\def\csname MBv/new/v3_echo_mel_imu_wrist/top1\endcsname{26.6}
\expandafter\def\csname MBv/new/v3_echo_mel_imu_wrist/macro_f1\endcsname{11.6}
\expandafter\def\csname MBv/new/v3_echo_wrist/top5\endcsname{66.9}
\expandafter\def\csname MBv/new/v3_echo_wrist/top1\endcsname{24.8}
\expandafter\def\csname MBv/new/v3_echo_wrist/macro_f1\endcsname{10.4}
\expandafter\def\csname MBv/new/v3_egofeat_wrist/top5\endcsname{70.5}
\expandafter\def\csname MBv/new/v3_egofeat_wrist/top1\endcsname{29.7}
\expandafter\def\csname MBv/new/v3_egofeat_wrist/macro_f1\endcsname{14.2}
\expandafter\def\csname MBv/new/v3_imu_egofeat_wrist/top5\endcsname{72.3}
\expandafter\def\csname MBv/new/v3_imu_egofeat_wrist/top1\endcsname{32.0}
\expandafter\def\csname MBv/new/v3_imu_egofeat_wrist/macro_f1\endcsname{14.8}
\expandafter\def\csname MBv/new/v3_imu_wrist/top5\endcsname{59.7}
\expandafter\def\csname MBv/new/v3_imu_wrist/top1\endcsname{20.1}
\expandafter\def\csname MBv/new/v3_imu_wrist/macro_f1\endcsname{11.2}
\expandafter\def\csname MBv/new/v3_mel_imu_egofeat_wrist/top5\endcsname{73.2}
\expandafter\def\csname MBv/new/v3_mel_imu_egofeat_wrist/top1\endcsname{34.7}
\expandafter\def\csname MBv/new/v3_mel_imu_egofeat_wrist/macro_f1\endcsname{15.6}
\expandafter\def\csname MBv/new/v3_mel_imu_wrist/top5\endcsname{62.2}
\expandafter\def\csname MBv/new/v3_mel_imu_wrist/top1\endcsname{21.3}
\expandafter\def\csname MBv/new/v3_mel_imu_wrist/macro_f1\endcsname{11.2}
\expandafter\def\csname MBv/new/v3_mel_wrist/top5\endcsname{48.4}
\expandafter\def\csname MBv/new/v3_mel_wrist/top1\endcsname{14.1}
\expandafter\def\csname MBv/new/v3_mel_wrist/macro_f1\endcsname{7.7}
\expandafter\def\csname MBv/new/videomae_ft/top5\endcsname{64.7}
\expandafter\def\csname MBv/new/videomae_ft/top1\endcsname{29.2}
\expandafter\def\csname MBv/new/videomae_ft/macro_f1\endcsname{13.2}
\expandafter\def\csname MBv/new/watchstats_noecho_probe/top5\endcsname{64.5}
\expandafter\def\csname MBv/new/watchstats_noecho_probe/top1\endcsname{32.9}
\expandafter\def\csname MBv/new/watchstats_noecho_probe/macro_f1\endcsname{11.3}
\expandafter\def\csname MBv/new/watchstats_probe/top5\endcsname{64.9}
\expandafter\def\csname MBv/new/watchstats_probe/top1\endcsname{33.0}
\expandafter\def\csname MBv/new/watchstats_probe/macro_f1\endcsname{11.4}
\expandafter\def\csname MBv/new/wecho_stats_probe/top5\endcsname{54.8}
\expandafter\def\csname MBv/new/wecho_stats_probe/top1\endcsname{27.6}
\expandafter\def\csname MBv/new/wecho_stats_probe/macro_f1\endcsname{8.4}
\expandafter\def\csname MBv/new/wmel_stats_probe/top5\endcsname{61.4}
\expandafter\def\csname MBv/new/wmel_stats_probe/top1\endcsname{27.8}
\expandafter\def\csname MBv/new/wmel_stats_probe/macro_f1\endcsname{7.9}
\expandafter\def\csname MBv/new/sec5-ego/top5\endcsname{58.6}
\expandafter\def\csname MBv/new/sec5-ego/top1\endcsname{27.3}
\expandafter\def\csname MBv/new/sec5-ego/macro_f1\endcsname{16.2}
\expandafter\def\csname MBv/new/sec5-mic/top5\endcsname{61.7}
\expandafter\def\csname MBv/new/sec5-mic/top1\endcsname{28.1}
\expandafter\def\csname MBv/new/sec5-mic/macro_f1\endcsname{14.1}
\expandafter\def\csname MBv/new/sec5-imu/top5\endcsname{61.0}
\expandafter\def\csname MBv/new/sec5-imu/top1\endcsname{26.4}
\expandafter\def\csname MBv/new/sec5-imu/macro_f1\endcsname{11.7}
\expandafter\def\csname MBv/new/sec5-gmic/top5\endcsname{49.8}
\expandafter\def\csname MBv/new/sec5-gmic/top1\endcsname{21.5}
\expandafter\def\csname MBv/new/sec5-gmic/macro_f1\endcsname{11.3}
\expandafter\def\csname MBv/new/sec5-exo/top5\endcsname{38.7}
\expandafter\def\csname MBv/new/sec5-exo/top1\endcsname{12.1}
\expandafter\def\csname MBv/new/sec5-exo/macro_f1\endcsname{8.0}
\expandafter\def\csname MBv/new/sec5-ego-watch/top5\endcsname{69.1}
\expandafter\def\csname MBv/new/sec5-ego-watch/top1\endcsname{39.6}
\expandafter\def\csname MBv/new/sec5-ego-watch/macro_f1\endcsname{20.4}
\expandafter\def\csname MBv/new/sec5-all/top5\endcsname{68.9}
\expandafter\def\csname MBv/new/sec5-all/top1\endcsname{38.9}
\expandafter\def\csname MBv/new/sec5-all/macro_f1\endcsname{20.6}
\expandafter\def\csname MBg/new/lavila/ego2/top5\endcsname{$-$2.1}
\expandafter\def\csname MBgl/new/lavila/ego2/top5\endcsname{$-$3.9}
\expandafter\def\csname MBgh/new/lavila/ego2/top5\endcsname{+0.3}
\expandafter\def\csname MBg/new/lavila/ego2/top1\endcsname{$-$0.4}
\expandafter\def\csname MBgl/new/lavila/ego2/top1\endcsname{$-$3.0}
\expandafter\def\csname MBgh/new/lavila/ego2/top1\endcsname{+2.5}
\expandafter\def\csname MBg/new/lavila/ego2/macro_f1\endcsname{+2.0}
\expandafter\def\csname MBgl/new/lavila/ego2/macro_f1\endcsname{+0.2}
\expandafter\def\csname MBgh/new/lavila/ego2/macro_f1\endcsname{+3.8}
\expandafter\def\csname MBg/new/sec5/aa/top5\endcsname{+6.3}
\expandafter\def\csname MBgl/new/sec5/aa/top5\endcsname{+3.4}
\expandafter\def\csname MBgh/new/sec5/aa/top5\endcsname{+9.4}
\expandafter\def\csname MBg/new/sec5/aa/top1\endcsname{+6.8}
\expandafter\def\csname MBgl/new/sec5/aa/top1\endcsname{+4.3}
\expandafter\def\csname MBgh/new/sec5/aa/top1\endcsname{+9.0}
\expandafter\def\csname MBg/new/sec5/aa/macro_f1\endcsname{+2.0}
\expandafter\def\csname MBgl/new/sec5/aa/macro_f1\endcsname{+0.9}
\expandafter\def\csname MBgh/new/sec5/aa/macro_f1\endcsname{+3.1}
\expandafter\def\csname MBg/new/lavila/aa/top5\endcsname{+1.8}
\expandafter\def\csname MBgl/new/lavila/aa/top5\endcsname{+0.5}
\expandafter\def\csname MBgh/new/lavila/aa/top5\endcsname{+3.3}
\expandafter\def\csname MBg/new/lavila/aa/top1\endcsname{+3.0}
\expandafter\def\csname MBgl/new/lavila/aa/top1\endcsname{+0.7}
\expandafter\def\csname MBgh/new/lavila/aa/top1\endcsname{+5.8}
\expandafter\def\csname MBg/new/lavila/aa/macro_f1\endcsname{+1.4}
\expandafter\def\csname MBgl/new/lavila/aa/macro_f1\endcsname{+0.2}
\expandafter\def\csname MBgh/new/lavila/aa/macro_f1\endcsname{+2.4}
\expandafter\def\csname MBg/new/sec5/micimu/top5\endcsname{+8.8}
\expandafter\def\csname MBgl/new/sec5/micimu/top5\endcsname{+5.5}
\expandafter\def\csname MBgh/new/sec5/micimu/top5\endcsname{+12.3}
\expandafter\def\csname MBg/new/sec5/micimu/top1\endcsname{+9.6}
\expandafter\def\csname MBgl/new/sec5/micimu/top1\endcsname{+6.7}
\expandafter\def\csname MBgh/new/sec5/micimu/top1\endcsname{+12.2}
\expandafter\def\csname MBg/new/sec5/micimu/macro_f1\endcsname{+3.5}
\expandafter\def\csname MBgl/new/sec5/micimu/macro_f1\endcsname{+2.4}
\expandafter\def\csname MBgh/new/sec5/micimu/macro_f1\endcsname{+4.7}
\expandafter\def\csname MBg/new/lavila/micimu/top5\endcsname{+2.7}
\expandafter\def\csname MBgl/new/lavila/micimu/top5\endcsname{+0.9}
\expandafter\def\csname MBgh/new/lavila/micimu/top5\endcsname{+4.6}
\expandafter\def\csname MBg/new/lavila/micimu/top1\endcsname{+2.8}
\expandafter\def\csname MBgl/new/lavila/micimu/top1\endcsname{+0.8}
\expandafter\def\csname MBgh/new/lavila/micimu/top1\endcsname{+4.8}
\expandafter\def\csname MBg/new/lavila/micimu/macro_f1\endcsname{+2.3}
\expandafter\def\csname MBgl/new/lavila/micimu/macro_f1\endcsname{+1.1}
\expandafter\def\csname MBgh/new/lavila/micimu/macro_f1\endcsname{+3.2}
\expandafter\def\csname MBg/new/sec5/watch/top5\endcsname{+10.5}
\expandafter\def\csname MBgl/new/sec5/watch/top5\endcsname{+6.6}
\expandafter\def\csname MBgh/new/sec5/watch/top5\endcsname{+14.6}
\expandafter\def\csname MBg/new/sec5/watch/top1\endcsname{+12.3}
\expandafter\def\csname MBgl/new/sec5/watch/top1\endcsname{+8.7}
\expandafter\def\csname MBgh/new/sec5/watch/top1\endcsname{+15.4}
\expandafter\def\csname MBg/new/sec5/watch/macro_f1\endcsname{+4.3}
\expandafter\def\csname MBgl/new/sec5/watch/macro_f1\endcsname{+3.0}
\expandafter\def\csname MBgh/new/sec5/watch/macro_f1\endcsname{+5.4}
\expandafter\def\csname MBg/new/lavila/watch/top5\endcsname{+3.4}
\expandafter\def\csname MBgl/new/lavila/watch/top5\endcsname{+1.1}
\expandafter\def\csname MBgh/new/lavila/watch/top5\endcsname{+5.6}
\expandafter\def\csname MBg/new/lavila/watch/top1\endcsname{+4.0}
\expandafter\def\csname MBgl/new/lavila/watch/top1\endcsname{+1.6}
\expandafter\def\csname MBgh/new/lavila/watch/top1\endcsname{+6.6}
\expandafter\def\csname MBg/new/lavila/watch/macro_f1\endcsname{+2.8}
\expandafter\def\csname MBgl/new/lavila/watch/macro_f1\endcsname{+1.5}
\expandafter\def\csname MBgh/new/lavila/watch/macro_f1\endcsname{+3.6}
\expandafter\def\csname MBg/new/sec5/gaa/top5\endcsname{+3.4}
\expandafter\def\csname MBgl/new/sec5/gaa/top5\endcsname{+1.7}
\expandafter\def\csname MBgh/new/sec5/gaa/top5\endcsname{+5.3}
\expandafter\def\csname MBg/new/sec5/gaa/top1\endcsname{+2.5}
\expandafter\def\csname MBgl/new/sec5/gaa/top1\endcsname{+0.8}
\expandafter\def\csname MBgh/new/sec5/gaa/top1\endcsname{+3.8}
\expandafter\def\csname MBg/new/sec5/gaa/macro_f1\endcsname{+1.6}
\expandafter\def\csname MBgl/new/sec5/gaa/macro_f1\endcsname{+0.5}
\expandafter\def\csname MBgh/new/sec5/gaa/macro_f1\endcsname{+2.4}
\expandafter\def\csname MBg/new/lavila/gaa/top5\endcsname{$-$0.0}
\expandafter\def\csname MBgl/new/lavila/gaa/top5\endcsname{$-$1.2}
\expandafter\def\csname MBgh/new/lavila/gaa/top5\endcsname{+1.3}
\expandafter\def\csname MBg/new/lavila/gaa/top1\endcsname{$-$2.3}
\expandafter\def\csname MBgl/new/lavila/gaa/top1\endcsname{$-$3.9}
\expandafter\def\csname MBgh/new/lavila/gaa/top1\endcsname{$-$1.0}
\expandafter\def\csname MBg/new/lavila/gaa/macro_f1\endcsname{+0.1}
\expandafter\def\csname MBgl/new/lavila/gaa/macro_f1\endcsname{$-$1.1}
\expandafter\def\csname MBgh/new/lavila/gaa/macro_f1\endcsname{+1.1}
\expandafter\def\csname MBg/new/sec5/all/top5\endcsname{+10.3}
\expandafter\def\csname MBgl/new/sec5/all/top5\endcsname{+6.1}
\expandafter\def\csname MBgh/new/sec5/all/top5\endcsname{+14.3}
\expandafter\def\csname MBg/new/sec5/all/top1\endcsname{+11.6}
\expandafter\def\csname MBgl/new/sec5/all/top1\endcsname{+8.4}
\expandafter\def\csname MBgh/new/sec5/all/top1\endcsname{+14.4}
\expandafter\def\csname MBg/new/sec5/all/macro_f1\endcsname{+4.5}
\expandafter\def\csname MBgl/new/sec5/all/macro_f1\endcsname{+3.5}
\expandafter\def\csname MBgh/new/sec5/all/macro_f1\endcsname{+5.5}
\expandafter\def\csname MBg/new/lavila/all/top5\endcsname{+2.2}
\expandafter\def\csname MBgl/new/lavila/all/top5\endcsname{$-$0.3}
\expandafter\def\csname MBgh/new/lavila/all/top5\endcsname{+4.5}
\expandafter\def\csname MBg/new/lavila/all/top1\endcsname{+2.2}
\expandafter\def\csname MBgl/new/lavila/all/top1\endcsname{$-$1.1}
\expandafter\def\csname MBgh/new/lavila/all/top1\endcsname{+5.4}
\expandafter\def\csname MBg/new/lavila/all/macro_f1\endcsname{+3.0}
\expandafter\def\csname MBgl/new/lavila/all/macro_f1\endcsname{+1.6}
\expandafter\def\csname MBgh/new/lavila/all/macro_f1\endcsname{+4.2}
\expandafter\def\csname MBg/new/sec5/exo/top5\endcsname{$-$6.5}
\expandafter\def\csname MBgl/new/sec5/exo/top5\endcsname{$-$13.5}
\expandafter\def\csname MBgh/new/sec5/exo/top5\endcsname{+0.1}
\expandafter\def\csname MBg/new/sec5/exo/top1\endcsname{$-$6.3}
\expandafter\def\csname MBgl/new/sec5/exo/top1\endcsname{$-$10.9}
\expandafter\def\csname MBgh/new/sec5/exo/top1\endcsname{$-$1.1}
\expandafter\def\csname MBg/new/sec5/exo/macro_f1\endcsname{$-$2.4}
\expandafter\def\csname MBgl/new/sec5/exo/macro_f1\endcsname{$-$4.9}
\expandafter\def\csname MBgh/new/sec5/exo/macro_f1\endcsname{$-$0.3}
\expandafter\def\csname MBg/new/lavila/exo/top5\endcsname{$-$11.8}
\expandafter\def\csname MBgl/new/lavila/exo/top5\endcsname{$-$18.4}
\expandafter\def\csname MBgh/new/lavila/exo/top5\endcsname{$-$6.1}
\expandafter\def\csname MBg/new/lavila/exo/top1\endcsname{$-$16.2}
\expandafter\def\csname MBgl/new/lavila/exo/top1\endcsname{$-$21.5}
\expandafter\def\csname MBgh/new/lavila/exo/top1\endcsname{$-$10.7}
\expandafter\def\csname MBg/new/lavila/exo/macro_f1\endcsname{$-$3.5}
\expandafter\def\csname MBgl/new/lavila/exo/macro_f1\endcsname{$-$5.4}
\expandafter\def\csname MBgh/new/lavila/exo/macro_f1\endcsname{$-$2.5}
\expandafter\def\csname MBv/new/prior/top5\endcsname{72.3}
\expandafter\def\csname MBv/new/prior/top1\endcsname{25.5}
\expandafter\def\csname MBv/new/prior/macro_f1\endcsname{2.4}
\expandafter\def\csname MBv/new/sec5-ego-prior/top5\endcsname{72.2}
\expandafter\def\csname MBv/new/sec5-ego-prior/top1\endcsname{38.4}
\expandafter\def\csname MBv/new/sec5-ego-prior/macro_f1\endcsname{15.8}
\expandafter\def\csname MBg/new/sec5/prior/top5\endcsname{+13.5}
\expandafter\def\csname MBgl/new/sec5/prior/top5\endcsname{+9.0}
\expandafter\def\csname MBgh/new/sec5/prior/top5\endcsname{+18.8}
\expandafter\def\csname MBg/new/sec5/prior/top1\endcsname{+11.1}
\expandafter\def\csname MBgl/new/sec5/prior/top1\endcsname{+7.5}
\expandafter\def\csname MBgh/new/sec5/prior/top1\endcsname{+14.3}
\expandafter\def\csname MBg/new/sec5/prior/macro_f1\endcsname{$-$0.3}
\expandafter\def\csname MBgl/new/sec5/prior/macro_f1\endcsname{$-$2.0}
\expandafter\def\csname MBgh/new/sec5/prior/macro_f1\endcsname{+1.2}
\newcommand{\SuppBaselineRows}{%
\tgrp{7}{Egocentric video} \\
Omnivore~\cite{Omnivore}\frozen & 71.5 & 42.3 & 19.2 & 64.0 & 33.9 & 14.2 \\
SlowFast~\cite{SlowFast}\frozen & 72.2 & 42.4 & 19.4 & 63.4 & 30.0 & 14.2 \\
CLIP ViT-L/14~\cite{CLIP}\frozen & 73.2 & 43.3 & 19.1 & 66.4 & 35.3 & 14.6 \\
AVT~\cite{AVT}\frozen & 71.8 & 43.2 & 19.6 & 61.0 & 26.4 & 12.1 \\
ImageBind~\cite{ImageBind}\frozen & 70.9 & 41.6 & 18.5 & 59.9 & 30.6 & 13.2 \\
EgoVLP~\cite{EgoVLP}\frozen & 73.7 & 45.5 & 22.7 & 68.3 & 37.2 & \textbf{17.2} \\
LaViLa~\cite{LaViLa}\frozen & \textbf{74.7} & \textbf{46.3} & \textbf{23.3} & \textbf{69.9} & \textbf{39.1} & 17.1 \\
VideoMAE~\cite{VideoMAE}\tuned & 73.1$^\dagger$ & 45.2$^\dagger$ & 22.0$^\dagger$ & 64.7$^\dagger$ & 29.2$^\dagger$ & 13.2$^\dagger$ \\
\tours VideoMAE + CLIP $\star$\frozen & 70.7 & 40.2 & \textbf{23.3} & 58.6 & 27.3 & 16.2 \\
\addlinespace[2pt]
\tgrp{7}{Egocentric video + audio} \\
TIM~\cite{TIM}\frozen\scratch & 77.4 & 44.2 & 18.3 & 71.8 & 35.7 & 12.8 \\
\addlinespace[2pt]
\tgrp{7}{Watch microphone} \\
Aud.\ SlowFast~\cite{AuditorySlowFast}\frozen & 67.2 & 36.4 & 14.3 & 59.3 & 26.9 & 10.7 \\
SSAST~\cite{SSAST}\tuned & \textbf{70.3} & 37.7 & 16.6 & 56.7 & 21.2 & 9.4 \\
ImageBind~\cite{ImageBind}\frozen & 67.7 & 37.3 & 15.4 & \textbf{61.9} & \textbf{30.8} & 12.4 \\
\tours SAMoSA + AST $\star$\frozen\tuned & \textbf{70.3} & \textbf{38.8} & \textbf{18.9} & 61.7 & 28.1 & \textbf{14.1} \\
\addlinespace[2pt]
\tgrp{7}{Glasses microphone} \\
Aud.\ SlowFast~\cite{AuditorySlowFast}\frozen & \textbf{67.7} & \textbf{37.2} & \textbf{14.7} & \textbf{57.4} & \textbf{27.0} & 10.3 \\
\tours AST $\star$\frozen & 58.4 & 26.2 & 14.2 & 49.8 & 21.5 & \textbf{11.3} \\
\addlinespace[2pt]
\tgrp{7}{Watch IMU} \\
DeepConvLSTM~\cite{DeepConvLSTM}\scratch & \textbf{70.3} & 33.4 & 12.7 & \textbf{67.8} & 27.1 & 9.8 \\
ImageBind~\cite{ImageBind}\frozen & 64.2 & \textbf{34.0} & 11.9 & 61.3 & \textbf{29.7} & 9.7 \\
\tours Pose-pre-trained CNN $\star$\frozen\tuned & 63.9 & 29.6 & \textbf{13.7} & 61.0 & 26.4 & \textbf{11.7} \\
\addlinespace[2pt]
\tgrp{7}{Exocentric video (reference)} \\
\tref{Omnivore~\cite{Omnivore}}\frozen & \tref{74.8} & \tref{44.1} & \tref{20.8} & \tref{42.3} & \tref{12.5} & \tref{7.3} \\
\tref{VideoMAE + CLIP $\star$}\frozen & \tref{68.9} & \tref{38.6} & \tref{22.3} & \tref{38.7} & \tref{12.1} & \tref{8.0} \\}
\newcommand{\SuppStrongEgoRows}{%
\tgrp{5}{Added to our egocentric-video model (verb top-5 alone: 70.7 / 58.6\% seen / new)} \\
+ Watch AA & \tpos{+2.1} \tci{+1.0}{+3.4} & \tpos{+2.0} \tci{+0.7}{+3.3} & \tpos{+6.3} \tci{+3.4}{+9.4} & \tpos{+6.8} \tci{+4.3}{+9.0} \\
+ Watch mic + IMU & \tpos{+4.8} \tci{+3.3}{+6.3} & \tpos{+4.8} \tci{+2.9}{+6.3} & \tpos{+8.8} \tci{+5.5}{+12.3} & \tpos{+9.6} \tci{+6.7}{+12.2} \\
+ Whole watch & \tpos{+5.0} \tci{+3.3}{+6.6} & \tpos{+5.2} \tci{+2.8}{+7.3} & \tpos{+10.5} \tci{+6.6}{+14.6} & \tpos{+12.3} \tci{+8.7}{+15.4} \\
+ Glasses AA & \tpos{+2.6} \tci{+1.6}{+3.4} & \tpos{+2.0} \tci{+0.9}{+2.9} & \tpos{+3.4} \tci{+1.7}{+5.3} & \tpos{+2.5} \tci{+0.8}{+3.8} \\
\tours + Watch + glasses & \tpos{+5.9} \tci{+4.3}{+7.5} & \tpos{+5.6} \tci{+3.3}{+7.3} & \tpos{+10.3} \tci{+6.1}{+14.3} & \tpos{+11.6} \tci{+8.4}{+14.4} \\
\tref{+ Exo} & \tpos{+3.1} \tci{+1.7}{+4.7} & \tpos{+3.2} \tci{+1.2}{+5.3} & $-$6.5 \tci{$-$13.5}{+0.1} & \tneg{$-$6.3} \tci{$-$10.9}{$-$1.1} \\
\addlinespace[2pt]
\tgrp{5}{Added to LaViLa\frozen{} (verb top-5 alone: 74.7 / 69.9\% seen / new)} \\
+ Second ego model (ours) & \tpos{+1.5} \tci{+0.8}{+2.2} & +0.4 \tci{$-$2.2}{+2.8} & $-$2.1 \tci{$-$3.9}{+0.3} & $-$0.4 \tci{$-$3.0}{+2.5} \\
+ Watch AA & \tpos{+1.5} \tci{+0.6}{+2.4} & +0.4 \tci{$-$1.0}{+1.8} & \tpos{+1.8} \tci{+0.5}{+3.3} & \tpos{+3.0} \tci{+0.7}{+5.8} \\
+ Watch mic + IMU & \tpos{+3.4} \tci{+2.2}{+4.4} & \tpos{+1.9} \tci{+0.3}{+3.2} & \tpos{+2.7} \tci{+0.9}{+4.6} & \tpos{+2.8} \tci{+0.8}{+4.8} \\
+ Whole watch & \tpos{+3.1} \tci{+1.7}{+4.6} & \tpos{+2.1} \tci{+0.6}{+3.6} & \tpos{+3.4} \tci{+1.1}{+5.6} & \tpos{+4.0} \tci{+1.6}{+6.6} \\
+ Glasses AA & \tpos{+1.5} \tci{+0.5}{+2.4} & $-$0.4 \tci{$-$2.0}{+1.1} & $-$0.0 \tci{$-$1.2}{+1.3} & \tneg{$-$2.3} \tci{$-$3.9}{$-$1.0} \\
\tours + Watch + glasses & \tpos{+3.0} \tci{+1.6}{+4.5} & +1.4 \tci{$-$0.6}{+3.2} & +2.2 \tci{$-$0.3}{+4.5} & +2.2 \tci{$-$1.1}{+5.4} \\
\tref{+ Exo} & \tpos{+2.5} \tci{+1.3}{+3.8} & +1.1 \tci{$-$1.3}{+3.5} & \tneg{$-$11.8} \tci{$-$18.4}{$-$6.1} & \tneg{$-$16.2} \tci{$-$21.5}{$-$10.7} \\}

\newcommand{\ExamplesN}{80}
\newcommand{\ExamplesNew}{41}
\newcommand{\ExamplesParticipants}{14}
\newcommand{\ExamplesVerbs}{15}

\newcommand{\ExamplesHandsNew}{13}
\newcommand{\ExamplesHandsRange}{1--20}

\newcommand{\ExamplesAaNew}{15}
\newcommand{\ExamplesAaRange}{21--50}

\newcommand{\ExamplesMotionNew}{4}
\newcommand{\ExamplesMotionRange}{51--55}

\newcommand{\ExamplesBothNew}{6}
\newcommand{\ExamplesBothRange}{56--75}

\newcommand{\ExamplesMissNew}{3}
\newcommand{\ExamplesMissRange}{76--80}

\newcommand{\SuppComputeRows}{%
\tbase Ego video & CLIP ViT-L/14 ($\times$10 frames), VideoMAE-B & 389.5 & 991 & 100.0 & -- & -- \\
\tref{+ exo video} & \tref{CLIP ViT-L/14 ($\times$10 frames), VideoMAE-B} & \tref{389.5} & \tref{991} & \tref{100.0} & \tref{+3.1} & \tref{$-$6.5} \\
\addlinespace[2pt]
\tgrp{7}{Wearable sensors added to egocentric video} \\
+ watch AA & FastViT-T8 ($\times$3 views), 2 frozen FastViT-T8 & 10.4 & 0.9 & 0.092 & \tpos{+2.1} & \tpos{+6.2} \\
+ watch IMU & 1D CNN ($\times$3 views), frozen 1D CNN & 0.24 & 0.011 & 0.001 & \tpos{+2.2} & \tpos{+5.1} \\
+ watch mic & SAMoSA trunk, frozen SAMoSA trunk, AST & 94.9 & 132 & 13.3 & \tpos{+4.1} & \tpos{+6.0} \\
+ glasses AA & as the watch AA & 10.4 & 0.9 & 0.092 & \tpos{+2.6} & \tpos{+3.4} \\
+ glasses mic & AST & 85.7 & 131 & 13.2 & +0.9 & +0.9 \\
\addlinespace[2pt]
\tgrp{7}{Sets of wearable sensors} \\
+ watch AA + glasses AA &  & 20.7 & 1.8 & 0.2 & \tpos{+4.0} & \tpos{+7.9} \\
+ whole watch &  & 105.5 & 133 & 13.4 & \tpos{+5.0} & \tpos{+10.5} \\
\tours + watch + glasses &  & 201.6 & 265 & 26.7 & \tpos{+5.8} & \tpos{+10.3} \\}
\newcommand{\CompAAPct}{0.18}

\newcommand{\CompAAParams}{10.4}
\newcommand{\CompAAGMAC}{0.91}
\newcommand{\CompCamGMAC}{991}
\newcommand{\CompCamParams}{390}
\newcommand{\CompASTGMAC}{131}

\newcommand{\CompWGNoASTPct}{0.3}

\usepackage{colortbl}
\definecolor{tshade}{gray}{0.93}
\newcommand{\tbase}{\rowcolor{tshade}}
\definecolor{tgrpc}{HTML}{2C3E7F}
\newcommand{\tgrp}[2]{\multicolumn{#1}{@{}l}{\rule{0pt}{2.5ex}\scriptsize\bfseries\color{tgrpc}#2}}
\newcommand{\tref}[1]{\textcolor{black!48}{#1}}
\newcommand{\tsd}[1]{{\scriptsize\,$\pm$\,#1}}
\definecolor{tours}{HTML}{E6F0FB}
\definecolor{tposc}{HTML}{1F7A6D}
\definecolor{tnegc}{HTML}{B8443F}
\newcommand{\tours}{\rowcolor{tours}}
\newcommand{\tpos}[1]{\textbf{\textcolor{tposc}{#1}}}
\newcommand{\tneg}[1]{\textbf{\textcolor{tnegc}{#1}}}
\newcommand{\tci}[2]{{\scriptsize\textcolor{black!55}{[#1, #2]}}}
\usepackage{multirow}
\newcommand{\tgroup}[2]{\multirow{#1}{*}{\rotatebox[origin=c]{90}{\scriptsize\bfseries\color{tgrpc}#2}}}
\newcommand{\tform}[1]{\,{\color{black!50}$#1$}}
\makeatletter
  {\end{minipage}\par\addvspace{8pt}}
\makeatother

\usepackage{titletoc}
\titlecontents{lsection}[1.6em]{\addvspace{0.45em}\bfseries}{\contentslabel{1.6em}}{\hspace*{-1.6em}}{\hfill\contentspage}
\titlecontents{lsubsection}[3.9em]{}{\contentslabel{2.3em}}{\hspace*{-2.3em}}{\titlerule*[0.6pc]{.}\contentspage}

\definecolor{cvprblue}{rgb}{0.21,0.49,0.74}
\usepackage[pagebackref,breaklinks,colorlinks,allcolors=cvprblue]{hyperref}
\hypersetup{pdftitle={EgoCARBS: Egocentric Video Meets Active Acoustics for Kitchen Action Understanding},
  pdfauthor={Saif Mahmud, Vineet Parikh, Bharath Hariharan, Cheng Zhang}}

\title{EgoCARBS: Egocentric Video Meets Active Acoustics for Kitchen Action Understanding}

\author{%
Saif Mahmud\textsuperscript{1}\qquad
Vineet Parikh\textsuperscript{2}\qquad
Bharath Hariharan\textsuperscript{1}\qquad
Cheng Zhang\textsuperscript{1}\\[0.4em]
\textsuperscript{1}Cornell University\qquad
\textsuperscript{2}Amazon
}

\begin{document}
\twocolumn[{%
\renewcommand\twocolumn[1][]{#1}%
\maketitle
\vspace{-3.2em}
\begin{center}\small
\texttt{sm2446@cornell.edu}\\[0.75em]
Project page: \href{https://saif-mahmud.github.io/projects/ego-carbs/}{\texttt{saif-mahmud.github.io/projects/ego-carbs}}
\end{center}
\vspace{0.95em}
\begin{center}
\centering
\captionsetup{type=figure}
\includegraphics[width=\textwidth]{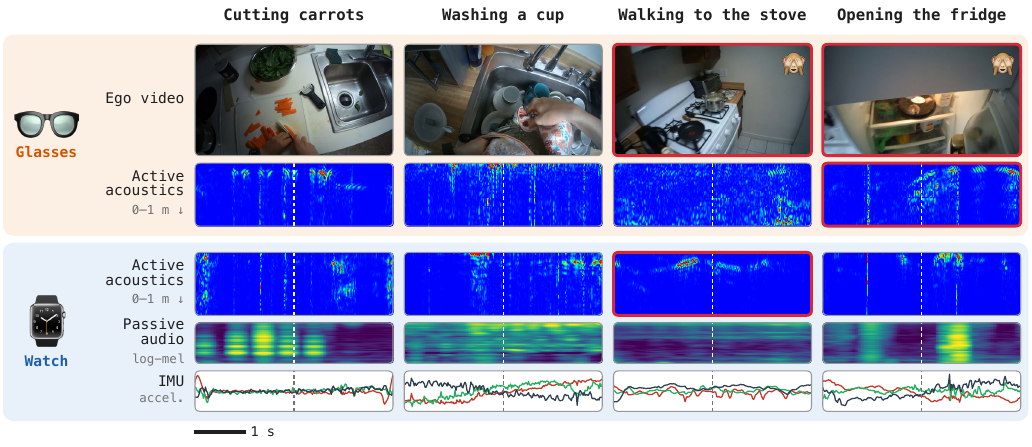}
\captionof{figure}{\textbf{EgoCARBS} records egocentric video and active acoustics (sonar) from camera glasses, and active acoustics, passive audio and IMU from a smartwatch, during unscripted home cooking.
Each column is one moment: the egocentric frame and 4\,s of every other stream around it (dashed line).
Active acoustics images how echoes change over distance (0--1\,m, top to bottom) and time.
When the hands are out of the camera's view (red), the echoes still capture the arm reaching for the stove (watch) and opening the fridge (glasses).}
\label{fig:teaser}
\end{center}

}]
\begin{abstract}
Egocentric video records what a person looks at, but often misses what their body does: the hands leave the frame, and arm motion goes unseen.
An exocentric camera recovers this context, but it must be installed wherever the activity happens.
Active acoustic sensing can recover it from the wearables themselves: a device's speaker emits inaudible chirps, and its microphone senses the motion of the hands, arms, and nearby objects from their echoes.
Yet no public dataset pairs it with egocentric video.
We present EgoCARBS, with two deliverables.
The first is a multimodal dataset and benchmark: 15 participants cook unscripted meals in their own kitchens wearing camera glasses and an off-the-shelf smartwatch that both sense with active acoustics, synchronized with egocentric and exocentric video and the watch's audio and IMU, with 4,019 verb--noun action annotations in 11.4~hours.
The second is a study of what the wearables add to egocentric video.
On verb recognition, they add at least as much as an exocentric camera for participants seen in training, and even more for new participants, for whom the exocentric camera does not help.
Active acoustics alone provides most of this gain, and helps most where the camera does not see the hands.
Trimmed action recognition, detection, and anticipation remain open.
We will release the data, our data collection and annotation tools, and pre-trained models.
\end{abstract}

\section{Introduction}
\label{sec:intro}

Understanding what a person is doing from the devices they wear underpins context-aware assistance in AR/VR, robotics, and assistive technology: a system that recognizes, localizes, and anticipates actions in daily tasks such as cooking can offer timely guidance, track progress, or log behavior.
Large egocentric video datasets such as EPIC-Kitchens~\cite{EK55, EK100} and Ego4D~\cite{Ego4d} made these problems measurable at scale and established verb--noun action recognition, temporal action detection, and action anticipation as standard benchmarks.

A head-mounted camera, however, sees where the wearer looks, and people do not always look where they act: a cook turns a stove knob while watching the pan, or opens the fridge without looking at the handle (\cref{fig:teaser}).
Ego-exo datasets~\cite{EgoExo4d, Assembly101, EgoLife, EPFLSmartKitchen} add third-person cameras that see the body and both hands, and together with the egocentric view they give the most complete visual record of an activity; we call egocentric plus exocentric video the \emph{ego-exo reference}.
But this reference cannot travel with the wearer: an exocentric camera must be installed wherever the activity takes place.
The modalities that wearables already record, sound and motion~\cite{EPICFusion, EPICSounds, IMUVideoMAE}, capture only part of an action: a smartwatch's IMU measures the motion of one wrist, and a microphone helps only when the action makes a sound.
What is missing is a wearable modality that senses the hands, the arms, and nearby objects when the camera cannot see them and the action is silent.

\textbf{Active acoustic sensing} is such a modality: it turns a wearable's speaker and microphone into a short-range sonar.
Many times per second, the speaker plays a \emph{chirp}, a brief sweep of tones too high for most people to hear, and the microphone records its echoes.
Each echo's delay gives the distance of the surface that reflected it, so stacking the chirps over time forms an \emph{echo profile}, an image of echo strength over distance and time in which a moving hand or arm leaves a trace (\cref{fig:sensing}).
The sensing reaches about one meter, enough to cover both hands and the objects they handle; a smartwatch needs only software, and camera glasses a small add-on board.
The echo profile complements video. The camera sees what the objects are, but only when they are in its view; the echoes carry no appearance and cannot tell a tomato from an onion, but they capture the movement of the hands and arms whether or not the camera sees them.

Research in human--computer interaction (HCI) has used this principle on glasses, wristbands, rings, and smartwatches to track hand pose~\cite{EchoWrist, RingAPose, WatchHand}, facial movements~\cite{EarIO}, and body pose~\cite{PoseSonic, EchoMotion}, and to recognize silent speech~\cite{EchoSpeech}, eating~\cite{MunchSonic}, and everyday activities~\cite{ActSonic}.
Yet active acoustics has not been studied as an input alongside egocentric video: where the two have met, video only labeled the echoes~\cite{ActSonic} or echoes only triggered the camera~\cite{EchoGuide}, and no public dataset records both.
To our knowledge, only two wearable active acoustics datasets are public~\cite{WatchHand, GrabNGo}; both were recorded in a lab, for hand pose and microgestures, and neither contains everyday activities or egocentric video.

We introduce \textbf{EgoCARBS} (\textbf{Ego}centric \textbf{C}amera + \textbf{A}coustics \textbf{R}ecording \textbf{B}enchmark and \textbf{S}ensing)%
, which has two deliverables.

The first is \textbf{a multimodal dataset and benchmark}: to our knowledge, the first public dataset to pair wearable active acoustics with egocentric and exocentric video, and the first public wearable active acoustics dataset recorded in the wild.
Fifteen participants cook meals of their choice in their own kitchens, without a script or cues.
They wear the two most widely worn form factors, glasses and a watch, and both sense with active acoustics (\cref{fig:hardware}): the camera glasses carry our small acoustic board, and an off-the-shelf smartwatch runs our sensing app, which also records passive audio and IMU.
A MacBook Pro's camera records the exocentric view for annotation and for the ego-exo reference.
Following EPIC-Kitchens~\cite{EK100}, we annotate 4,019 temporally localized verb--noun actions in 11.4~hours and define benchmarks for action recognition, detection, and anticipation, on participants seen in training and on new ones.
We provide a multimodal baseline and reference results for existing video, audio, and IMU models, among them Omnivore~\cite{Omnivore}, ImageBind~\cite{ImageBind}, EgoVLP~\cite{EgoVLP}, LaViLa~\cite{LaViLa}, TIM~\cite{TIM}, SSAST~\cite{SSAST}, and DeepConvLSTM~\cite{DeepConvLSTM}.

The second is \textbf{a study of what active acoustics adds to egocentric video}, in three research questions (\cref{sec:exp_results}).
We report verb top-5 and top-1 accuracy, as EPIC-Kitchens does~\cite{EK100}, and macro F1, which weighs every verb equally; the answers below give top-5.
\begin{list}{}{\setlength{\leftmargin}{2.3em}\setlength{\labelwidth}{2.0em}\setlength{\labelsep}{0.3em}\setlength{\itemsep}{2pt}\setlength{\topsep}{3pt}\setlength{\parsep}{0pt}}
    \item[\textbf{RQ1}] \textbf{How much of the gap between egocentric video and the ego-exo reference do the wearables close?}
    For participants seen in training, all of it: the watch and the glasses raise egocentric video from \RSEgoVFive\% to \RSEgoWGVFive\%, above the \RSEgoExoVFive\% of adding the exocentric camera.
    The watch's active acoustics alone closes about two-thirds of the gap, and helps most where no hand is visible to the camera.
    \item[\textbf{RQ2}] \textbf{Does active acoustics add information beyond the watch's passive audio and IMU?}
    For new participants, yes: it improves a model that already uses both; for seen participants, its added gain is small.
    Its value depends on timing: when the echoes of each moment are replaced by echoes from another moment of the same person, the gain disappears, so the gain is tied to the moment of the action, not to the person or the kitchen.
    \item[\textbf{RQ3}] \textbf{Do the gains hold for new participants and for a stronger video model?}
    Yes: for new participants the wearables raise egocentric video from \RNEgoVFive\% to \RNEgoWGVFive\%, whereas the exocentric camera, placed differently in every home, does not help.
    Added to LaViLa, the strongest egocentric model we evaluate, the watch still adds \MBg{seen}{lavila/watch}{top5} points for seen and \MBg{new}{lavila/watch}{top5} for new participants.
\end{list}
These results show that the modality helps; they leave the benchmark far from solved. Trimmed actions, temporal detection, anticipation, and rare verbs remain open.

We will release the de-identified data and annotations, our data collection and annotation tools for an off-the-shelf smartwatch and camera glasses, and the benchmark code with pre-trained models.

\section{Related Work}
\label{sec:related_work}

EgoCARBS connects two lines of work that have developed apart, egocentric video datasets and wearable active acoustic sensing, through multimodal egocentric learning, which lies between them.

\paragraph{Egocentric and ego-exo video.}
Kitchen activities have long been a testbed for action understanding~\cite{50Salads, Breakfast, CMUMMAC, delpretoLiu2022actionSense, EGTEA}.
EPIC-Kitchens~\cite{EK55, EK100} scaled unscripted egocentric recording to participants' own kitchens and introduced the verb--noun annotations and the recognition, detection, and anticipation benchmarks that have become standard~\cite{RULSTM, AVT}; HD-EPIC~\cite{HDEPIC} adds far more detailed annotations with Project Aria glasses~\cite{ProjectAria}, and Ego4D~\cite{Ego4d} and EgoLife~\cite{EgoLife} span daily life beyond the kitchen.
Other datasets observe the wearer's body with exocentric cameras~\cite{CharadesEgo, Assembly101, EgoExoLearn, EgoExo4d, EPFLSmartKitchen} or with gaze, hand tracking, and motion capture~\cite{HoloAssist, Nymeria}.
We adopt the EPIC-Kitchens protocol and take the ego-exo setting as our reference, and add what none of these datasets has: sensing of the wearer's hands and body from the wearables themselves.

\paragraph{Multimodal egocentric learning.}
Multimodal models pair egocentric video with body-worn sensors to cover what the camera misses, and face a trade-off: the signals every wearable carries capture only part of an action, and richer signals need special equipment.
Passive audio complements egocentric video~\cite{EPICFusion, EPICSounds} but hears an action only if it makes a sound.
IMUs have been recorded with egocentric video~\cite{WEAR, CAPTURE24}, aligned with vision and language~\cite{IMU2CLIP, ImageBind}, and fused with video models~\cite{IMUVideoMAE, COMODO}, and smartwatches combine motion and sound~\cite{Ubicoustics, SAMoSA}; yet an IMU measures only the device that carries it.
Richer body sensing has required instrumenting the person or the space: motion capture, EMG, and tactile gloves~\cite{CMUMMAC, delpretoLiu2022actionSense}, ambient sensors~\cite{Opportunity, MMAct, HARMES}, skin-contact EMG~\cite{emg2qwerty, emg2pose, NeuromotorInterface}, or radio sensing from fixed infrastructure~\cite{MMFi, XRF55, XRFV2}.
Closest in motivation, FEEL~\cite{FEEL} adds hand contact force from sensor gloves.
Active acoustics resolves this trade-off: like passive audio, it needs only the speaker and microphone that a smartwatch already has~\cite{WatchHand}, and like instrumented sensing, it captures the motion of the hands, arms, and nearby objects, even when they make no sound. None of the datasets above includes it.

\paragraph{Active acoustic sensing.}
In computer vision, echoes have been used to learn spatial image representations~\cite{VisualEchoes}, to predict depth with images~\cite{BeyondImage2Depth}, and to estimate 3D human pose from a fixed speaker and microphone~\cite{AcousticPose}; each of these senses from the room.
HCI research moved the sensor onto the body, after early smartphone systems~\cite{FingerIO, LLAP, CAT}.
On glasses and earables, it recognizes silent speech~\cite{EchoSpeech} and sign language~\cite{SonicASL}, tracks the face~\cite{EarIO, EyeEcho}, gaze~\cite{GazeTrak}, and upper-body pose~\cite{PoseSonic}, and recognizes everyday activities~\cite{ActSonic} and dietary actions~\cite{MunchSonic}; on the wrist and hand, it tracks hand pose~\cite{EchoWrist, RingAPose, WatchHand} and recognizes microgestures~\cite{GrabNGo}.
EchoMotion~\cite{EchoMotion} tracks full-body pose from a single off-the-shelf smartwatch, well beyond the hand.
Only WatchHand~\cite{WatchHand} and Grab-n-Go~\cite{GrabNGo} release their recordings, both made in a lab for hand pose and microgestures.
Where these systems have met egocentric video, the echoes and the video never entered the same model: ActSonic~\cite{ActSonic}, the closest to our setting, recognizes 27 coarse activities in participants' homes and uses video only to label its echoes, and EchoGuide~\cite{EchoGuide} uses echoes only to trigger video capture.
EgoCARBS pairs the echoes with egocentric and exocentric video as inputs on the same verb--noun actions, recorded in the wild, from both glasses and a watch.

\section{The EgoCARBS Sensing System}
\label{sec:system}

\begin{figure*}[t]
\centering
\includegraphics[width=\textwidth]{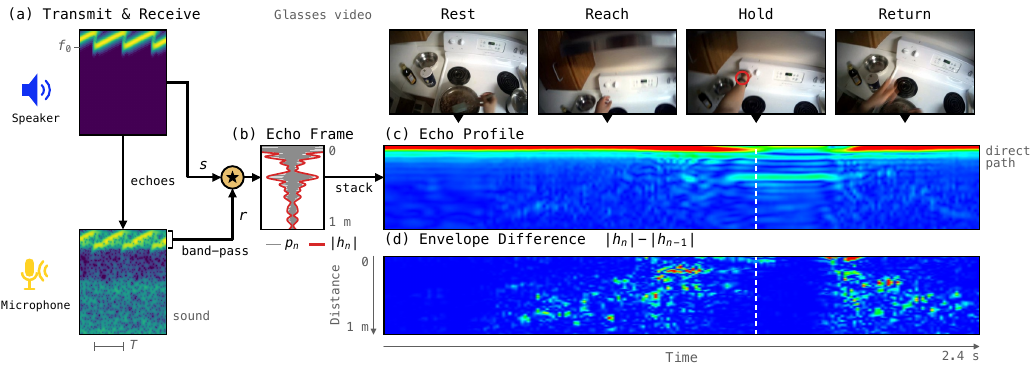}
\caption{\textbf{Active acoustic sensing}, the same pipeline on both wearables, shown on the watch as its hand reaches to a stove knob.
(a)~The speaker plays a chirp $s$ of period $T$; the microphone records its echoes and the surrounding sound.
(b)~The recording, band-passed to $r$ and cross-correlated with $s$ ($\star$), gives an echo frame $p_n$ and its envelope $|h_n|$.
(c)~Stacked envelopes form the echo profile, aligned with the glasses' video: the reach and the return scatter it, and the hold steadies it. The dashed line marks (b).
(d)~The envelope difference shows only what moved (shown squared for contrast).}
\label{fig:sensing}
\end{figure*}

\begin{figure}[t]
\centering
\includegraphics[width=\columnwidth]{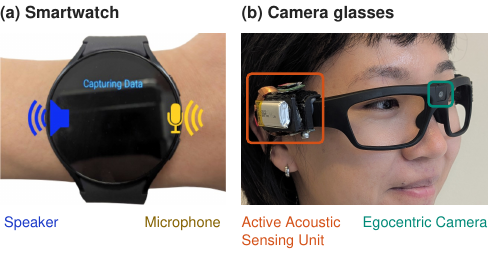}
\caption{\textbf{The two wearables.}
(a)~The smartwatch uses its built-in speaker and microphone for active acoustic sensing.
(b)~The glasses record egocentric video through the bridge camera and carry an acoustic sensing unit on the temple.}
\label{fig:hardware}
\end{figure}

\begin{table}[t]
\centering
\caption{Recorded streams. AA: active acoustics.}
\label{tab:streams}
\footnotesize
\setlength{\tabcolsep}{3pt}
\begin{tabular}{@{}llll@{}}
\toprule
Stream & Device & Rate & Details \\
\midrule
AA            & Watch       & 48\,kHz & 18--21\,kHz chirp, 80/s \\
Passive audio & Watch       & 16\,kHz & 64-band log-mel \\
IMU           & Watch       & 50\,Hz  & accel., gyro., magnetometer \\
AA            & Glasses     & 50\,kHz & 21.5--24.5\,kHz chirp, 83.3/s \\
Ego video     & Glasses     & 30\,fps & 2560$\times$1440, 135$^\circ$ lens \\
Ego audio     & Glasses     & 32\,kHz & mono \\
Exo video     & MacBook Pro & 30\,fps & 1280$\times$720 \\
\bottomrule
\end{tabular}
\end{table}

EgoCARBS records each session with three devices (\cref{fig:hardware,tab:streams}): camera glasses, which record the egocentric video and carry an active acoustic sensor; an off-the-shelf smartwatch, which senses with active acoustics and also records passive audio and motion; and a MacBook Pro, whose camera records the exocentric reference and whose clock aligns all streams.

\subsection{Active acoustic sensing}
\label{sec:system_signal}

Both wearables run the same pipeline, which \cref{fig:sensing} follows from the transmitted chirp to the input of our models; only the frequency band and sampling rate differ (\cref{sec:system_hardware}).

\paragraph{Transmit and receive (\cref{fig:sensing}a).}
The speaker repeatedly plays a linear chirp $s$, a frequency-modulated continuous wave (FMCW) that sweeps a bandwidth $B=3$\,kHz in $N=600$ samples.
The microphone records the chirp's echoes from everything in range, the hands, the forearms, the torso, and the objects they handle, together with the audible sound around it, which becomes the watch's passive audio.

\paragraph{From recording to echo frame (\cref{fig:sensing}a--b).}
We band-pass filter the recording to the chirp band and, as in C-FMCW~\cite{CFMCW}, cross-correlate each chirp period of the filtered signal $r$ with the transmitted chirp:
\begin{equation}
p_n[k] = \sum_{m=0}^{N-1} r[nN+k+m]\,s[m].
\label{eq:echo_frame}
\end{equation}
The echo frame $p_n$ peaks at the delay $k$ of every propagation path.
Because $p_n$ oscillates at the chirp's frequency, we take its envelope $|h_n|$ from the analytic signal $h_n=p_n+j\,\mathcal{H}\{p_n\}$, where $\mathcal{H}$ is the Hilbert transform; the envelope is the strength of the echo at each distance.

\paragraph{Echo profile and envelope difference (\cref{fig:sensing}c--d).}
Stacking the envelopes of successive frames gives the echo profile, an image of echo strength over distance and time.
Static surfaces draw horizontal bands, and a moving hand bends them.
Like the difference of consecutive video frames, the difference of consecutive envelopes,
\begin{equation}
D_n[k] = \lvert h_n[k]\rvert - \lvert h_{n-1}[k]\rvert,
\label{eq:diff}
\end{equation}
cancels static reflections, and with them much of what changes when a device is put on again, while it keeps what moved.
This envelope-difference profile is the input of our acoustic models (\cref{sec:exp_models}).

\paragraph{Range and resolution.}
An echo from a surface at distance $d$ travels $2d$ farther than the direct path from speaker to microphone, so it arrives $k = 2df_s/c$ lags after it, where $f_s$ is the sampling rate and $c\approx343$\,m/s the speed of sound.
One lag thus spans $c/(2f_s)$, 3.6\,mm on the watch ($f_s=48$\,kHz) and 3.4\,mm on the glasses (50\,kHz); two surfaces are resolved when their distances differ by $c/(2B) = 5.7$\,cm; and a frame lasts one chirp period $N/f_s$, 12--12.5\,ms.
Echoes remain measurable to about 1\,m, covering both hands and the objects they handle; our models read the first 200 lags past the direct path, about 0.7\,m.

\subsection{Implementation on the watch and the glasses}
\label{sec:system_hardware}

\paragraph{Smartwatch.}
Our Wear OS app turns an off-the-shelf Samsung Galaxy Watch7 into an active acoustic sensor in software.
It plays an 18--21\,kHz chirp through the built-in speaker, 80 times per second, and records the built-in microphone at 48\,kHz with the operating system's audio processing disabled, since it would distort the echoes.
The watch is worn on the left wrist; because the echoes reach about 1\,m, it also senses the right hand.

\paragraph{Glasses.}
The camera glasses (zShades-2K) record 2560$\times$1440 video at 30\,fps through a 135$^\circ$ lens, but expose no speaker or microphone that an app could drive, so we attach our own acoustic board (\cref{fig:hardware}b).
An nRF52840 microcontroller plays a 21.5--24.5\,kHz chirp, 83.3 times per second, and records the microphone at 50\,kHz to a memory card.
The two sensors use disjoint bands and different repetition rates, so neither leaves a fixed pattern in the other's echo profiles.

\subsection{Passive streams and synchronization}
\label{sec:system_passive}

\paragraph{Passive audio and IMU.}
The audible band of the watch's recording is its passive audio: we low-pass filter it below the chirp, resample it to 16\,kHz, and compute a 64-band log-mel spectrogram (0.6\,s windows every 30\,ms, 10\,Hz--8\,kHz), which we release in place of the raw audio.
The watch's accelerometer, gyroscope, and magnetometer are resampled to 50\,Hz, and the glasses' 32\,kHz audio serves as a second microphone.

\paragraph{Synchronization.}
\label{sec:system_sync}
All streams share the MacBook Pro's clock, which also times the exocentric video.
The watch streams its audio with clock heartbeats, and a clap at the start of each session pins its offset.
The glasses' camera and acoustic board run on their own clocks, so we align each to the watch through the kitchen's own sounds: every 30\,s, we find the time shift at which the two recordings match best~\cite{GCCPHAT}, and fit a line through the shifts.
The fits leave residuals of \SyncRmsMin--\SyncRmsMax\,ms rms (median \SyncRmsMedian\,ms).

\section{The EgoCARBS Dataset}
\label{sec:dataset}

EgoCARBS holds one session from each of 15 participants cooking in their own kitchens; its annotations and splits define the benchmark, and its exocentric reference and hand-visibility labels support the study.

\subsection{Recording}
\label{sec:dataset_recording}

\paragraph{Participants and sessions.}
Each participant cooks a dish of their choice in the kitchen of their own home and ends the recording when done; sessions last 22--68\,min (mean 45\,min).
Nothing is scripted or cued: the researcher does not prompt, correct, or converse with the participant while recording.
Participants were 19--54 years old (median 26); nine were women and five men, one preferred not to say, and 14 are right-handed.

\paragraph{Devices and wearability.}
Before recording, we fit the watch and tilt the camera glasses until the participant's outstretched hands sit near the center of the frame.
The laptop's camera frames the upper body and hands at the main work area.
After the session, participants rated wearing the devices at a median of 4 on a 5-point comfort scale, and 13 of the 15 heard no sound from the system.

\paragraph{Privacy.}
The study was approved by our institutional review board; all participants gave informed consent, were compensated, and may withdraw their session. Before release, faces and other identifiers are blurred in both videos, speech is removed from the audio, and the active acoustics is released as echo profiles only.

\subsection{Annotation}
\label{sec:dataset_annotation}

\paragraph{Action segments.}
Following the EPIC-Kitchens protocol~\cite{EK100}, each action is a temporally localized segment with a start and end time, a verb, and the object acted on.
Google's Gemini 3.8 Flash, used only as an annotation tool, first proposes segments from the synchronized egocentric and exocentric video; two annotators then review and correct every segment in our fork of VIDAT~\cite{VIDAT}, which shows every synchronized video and sensing stream on one timeline.

\paragraph{Vocabulary and classes.}
The annotations hold \VocabVerbs{} verbs and \VocabNames{} object names, and \VocabNamesMapped{} of the names map to \VocabEkNouns{} noun classes of EPIC-Kitchens-100~\cite{EK100}, so that models can be pre-trained on or evaluated against it.
A verb becomes a benchmark class when at least \VocabMinParticipants{} participants perform it for at least \VocabMinSeconds\,s in total; the resulting \VocabVerbClasses{} verb classes cover \VocabVerbKeptPct\% of the annotated time.
The same rule selects \VocabNounClasses{} noun classes and \VocabActionClasses{} verb--noun actions (\cref{sec:exp_tasks}).

\paragraph{Hand visibility.}
To test whether the wearables help where the camera does not see the hands (\cref{sec:exp_results}), we mark each 1\,s window by how many of three egocentric frames spread over it show a hand, as detected by MediaPipe Hands~\cite{MediaPipeHands}.

\subsection{Statistics}
\label{sec:dataset_stats}
\begin{figure}[t]
\centering
\includegraphics[width=\linewidth]{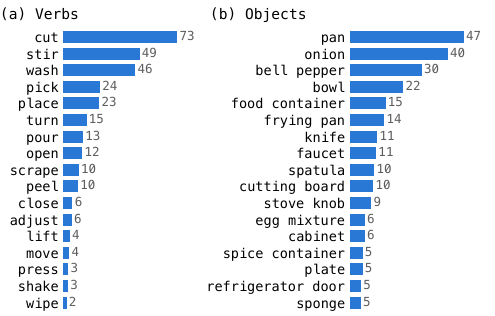}
\caption{Annotated minutes per class: (a) the benchmark verbs and (b) the most annotated objects.}
\label{fig:dataset_stats}
\end{figure}

The recordings total 11.4\,h and hold 4,019 action segments, 5.9 per minute.
Segments are short (median 2.4\,s, 90\% under 10.5\,s); time between annotated segments is not scored in recognition and is background in detection.
The classes are long-tailed (\cref{fig:dataset_stats}): cut, stir, and wash make up 55\% of the annotated verb time, while the rarest verbs have 2--3\,min each in total.
The camera often misses the hands: in \RSHandShareNone\% of the annotated 1\,s windows, no egocentric frame shows a hand, and in another \RSHandShareSome\% only some do.

\subsection{Splits}
\label{sec:dataset_splits}
Every task is evaluated on two splits.
For \textbf{seen participants}, following EPIC-Kitchens~\cite{EK100}, each session is cut into 20 time slices that rotate through five folds, so that every participant appears in training and test.
For \textbf{new participants}, each participant is held out in turn (15 folds), which tests whether a model transfers to a new person, kitchen, and device fit, as in prior active acoustics work~\cite{ActSonic, MunchSonic}.

\section{Benchmarks and Experiments}
\label{sec:experiments}

Our experiments serve the paper's two deliverables: the study of what active acoustics adds to egocentric video (\cref{sec:exp_results}), then the benchmark and its open tasks (\cref{sec:exp_benchmark}).

\begin{figure*}[t]
\centering
\includegraphics[width=\textwidth]{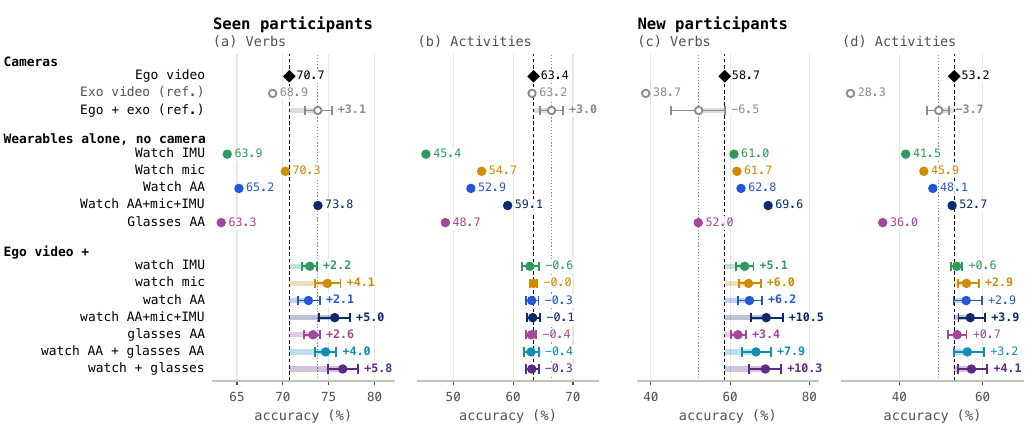}
\caption{\textbf{From egocentric video toward egocentric plus exocentric video}, on 1\,s windows, for seen (a, b) and new participants (c, d): top-5 accuracy over the \VocabVerbClasses{} verbs (a, c) and top-1 accuracy over the five activities (b, d).
``Ego video +'' rows show the gain over ego video with its 95\% interval (bold: excludes zero). Dashed line: ego video; dotted: ego + exo. AA: active acoustics.}
\label{fig:spectrum}
\end{figure*}

\subsection{Tasks and metrics}
\label{sec:exp_tasks}
\begin{itemize}
    \item \textbf{Window recognition.} Every 1\,s window, at a 0.5\,s stride, that overlaps an annotated segment by at least 0.5\,s is classified into the \VocabVerbClasses{} verbs and, separately, into five coarser activities that group verbs (knife work, washing, stirring, transport, access). This is our primary task, because it does not assume that an action's boundaries are given; models may use context around the window, so the evaluation is offline.
    \item \textbf{Trimmed action recognition}, following EPIC-Kitchens-100~\cite{EK100}: each annotated segment by verb, noun (\VocabNounClasses{} classes), and action (\VocabActionClasses{} classes).
    \item \textbf{Temporal action detection} of the five activities in the untrimmed session, by an MS-TCN~\cite{MSTCN}, scored by mAP over temporal IoU thresholds of 0.1--0.5.
    \item \textbf{Action anticipation} of the next verb from the 2\,s that end 1\,s before it starts, by class-mean top-5 recall~\cite{EK100}; only the microphones' 3\,s audio features reach closer.
\end{itemize}
Recognition reports top-5 and top-1 accuracy, as EPIC-Kitchens does, and macro F1, the mean of the per-class F1 scores.
Because verbs are long-tailed (\cref{fig:dataset_stats}), every classifier is trained with class weights, and macro F1 shows how a model does on the rare verbs as well as the frequent ones. Every gain comes with a 95\% interval from 1,000 resamples of the test participants.
No model receives the exocentric video except in the rows marked \emph{reference}, since a deployed system has none.

\subsection{A multimodal baseline}
\label{sec:exp_models}
To score any combination of modalities, we build one baseline in which every modality has its own classifier and the classifiers are fused late: a setup sums the calibrated log-probabilities of its modalities with equal weights, none fitted per combination, and a modality that did not record a window adds nothing there.
Each classifier starts from an existing pre-trained model where one exists: frozen VideoMAE~\cite{VideoMAE} and CLIP~\cite{CLIP} features for egocentric video (and the exocentric reference), and SAMoSA's sound model~\cite{SAMoSA} and AST~\cite{AST} for passive audio.
No pre-trained model exists for wearable echo profiles, so for active acoustics we pre-train a FastViT-T8~\cite{FastViT} on the envelope-difference profile ourselves, to predict full-body pose from echoes on a separate 12.9\,h motion-capture corpus recorded with a smartwatch; we will release these pre-trained weights.
The IMU's 1D CNN is pre-trained the same way on that corpus's IMU; no EgoCARBS data is used in pre-training.
The supplementary material gives architectures and training details, and top-1 accuracy and macro F1 for every setup.

\subsection{What active acoustics adds}
\label{sec:exp_results}

\begin{figure}[t]
\centering
\includegraphics[width=\linewidth]{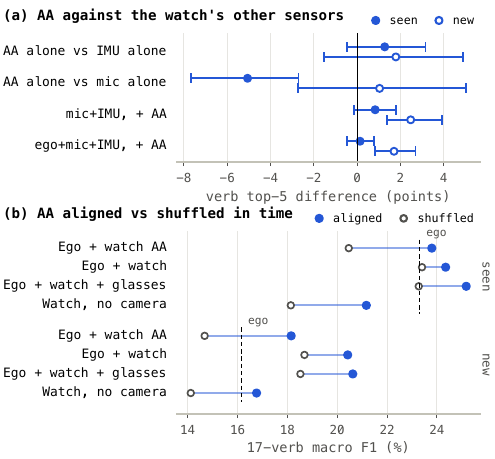}
\caption{\textbf{What active acoustics (AA) adds beyond the watch's other sensors.}
(a)~Verb top-5 differences: AA against the IMU and the microphone, and AA added to both, without and with ego video.
(b)~Verb macro F1 with AA aligned, and with each window's AA output replaced by that of another window of the same participant; dashed: ego video alone.
Whiskers: 95\% intervals.}
\label{fig:aa}
\end{figure}

\begin{figure}[t]
\centering
\includegraphics[width=\linewidth]{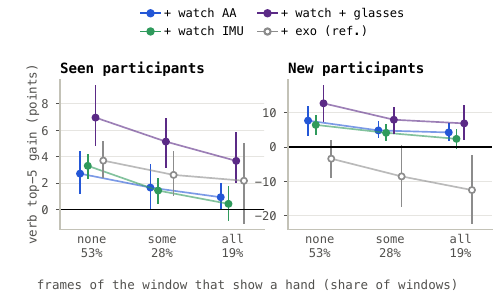}
\caption{\textbf{Gains where the camera does not see the hands}: verb top-5 gain over ego video, by how many of three egocentric frames of a window show a hand. Whiskers: 95\% intervals.}
\label{fig:hands}
\end{figure}

\begin{table*}[t]
\centering
\caption{\textbf{Beyond 1\,s windows.} (a)~Trimmed action recognition (EPIC-Kitchens-100's protocol), accuracy (\%) over \RSSegN{} segments; AA: watch and glasses. (b)~Temporal detection of the five activities (mAP, mean of 3 seeds) and next-verb anticipation (Ant., class-mean top-5 recall). Bold: best setup without the exocentric camera.}
\label{tab:segments}
\footnotesize
\setlength{\tabcolsep}{2.3pt}
\begin{minipage}[t]{0.66\textwidth}
\centering
\begin{tabular}[t]{@{}lcccccccc@{}}
\toprule
& \multicolumn{4}{c}{Seen participants} & \multicolumn{4}{c}{New participants} \\
\cmidrule(lr){2-5}\cmidrule(l){6-9}
(a) Input & Verb@1 & Verb@5 & Noun@1 & Act.@1 & Verb@1 & Verb@5 & Noun@1 & Act.@1 \\
\midrule
\ResultsSegmentRowsMain
\bottomrule
\end{tabular}
\end{minipage}\hfill
\begin{minipage}[t]{0.33\textwidth}
\centering
\setlength{\tabcolsep}{1.7pt}
\begin{tabular}[t]{@{}lcccc@{}}
\toprule
& \multicolumn{2}{c}{Seen} & \multicolumn{2}{c}{New} \\
\cmidrule(lr){2-3}\cmidrule(l){4-5}
(b) Input & mAP$\uparrow$ & Ant.$\uparrow$ & mAP$\uparrow$ & Ant.$\uparrow$ \\
\midrule
\ResultsDetectRowsMain
\bottomrule
\end{tabular}
\end{minipage}
\end{table*}

\paragraph{Active acoustics alone recognizes actions.}
From the watch's echoes alone, with no camera, microphone, or IMU (\cref{fig:spectrum}, ``Wearables alone''), the baseline recognizes the verb of a 1\,s window at \RSAAVFive\% top-5 accuracy for seen participants, against \RSEgoVFive\% for egocentric video, and at \RNAAVFive\% for new participants, where it exceeds egocentric video (\RNEgoVFive\%) in accuracy, though not in macro F1 (\RNAAVF\% against \RNEgoVF\%).

\paragraph{RQ1: the wearables close the gap to the ego-exo reference.}
For seen participants, adding the watch and the glasses raises egocentric video's verb top-5 from \RSEgoVFive\% to \RSEgoWGVFive\% (\RSEgoWGGainVFive{} points), above the \RSEgoExoVFive\% of adding the exocentric camera (\RSEgoExoGainVFive; \cref{fig:spectrum}a).
In macro F1, the wearables and the exocentric camera gain alike (\RSEgoWGGainVFOne{} and \RSEgoExoGainVFOne).
Active acoustics carries much of the accuracy gain: the watch's alone adds \RSEgoAAGainVFive, and that of both wearables, with no microphone or IMU, \RSEgoAAGAAGainVFive.
The gain concentrates where the camera does not see the hands (\cref{fig:hands}): in the \RSHandShareNone\% of test windows where no egocentric frame shows a hand, the watch's active acoustics adds \RSHandAANone{} verb top-5 for seen participants, against \RSHandAAAll{} where every frame shows a hand, and for new participants \RNHandAANone{} against \RNHandAAAll.
Hand detections are only a proxy for visibility.

\paragraph{RQ2: active acoustics adds what the watch's other sensors miss.}
For new participants, adding active acoustics to the watch's microphone and IMU improves the camera-free watch (\RNAAonMicIMUVFive), and still adds to egocentric video, microphone, and IMU (\RNAAonEgoMicIMUVFive; \cref{fig:aa}a).
For seen participants both top-5 gains are small and not significant, although the camera-free watch gains in macro F1 (\RSAAonMicIMUVF).
The gain is tied to the moment of the action (\cref{fig:aa}b): when the echoes of each window are replaced by those of another window of the same participant, which keeps the person, the kitchen, and their verb frequencies but breaks the timing, every setup with active acoustics drops in macro F1, and egocentric video plus the watch's active acoustics falls below egocentric video alone.

\begin{table}[!tb]
\centering
\caption{\textbf{Existing models per modality} on window recognition: verb top-5 and top-1 accuracy and macro F1 (\%), on the \MBSeenWindows{} test windows of each split. $\star$: our baseline's models, as in \cref{fig:spectrum}. Bold: best of each group.}
\label{tab:baselines}
\footnotesize
\setlength{\tabcolsep}{2.4pt}
\begin{tabular}{@{}lcccccc@{}}
\toprule
& \multicolumn{3}{c}{Seen participants} & \multicolumn{3}{c}{New participants} \\
\cmidrule(lr){2-4}\cmidrule(l){5-7}
Model & Top-5 & Top-1 & F1 & Top-5 & Top-1 & F1 \\
\midrule
\multicolumn{7}{@{}l}{\textit{Egocentric video}} \\
Omnivore~\cite{Omnivore} & \MBv{seen}{omnivore_probe}{top5} & \MBv{seen}{omnivore_probe}{top1} & \MBv{seen}{omnivore_probe}{macro_f1} & \MBv{new}{omnivore_probe}{top5} & \MBv{new}{omnivore_probe}{top1} & \MBv{new}{omnivore_probe}{macro_f1} \\
ImageBind~\cite{ImageBind} & \MBv{seen}{imagebind_vision_probe}{top5} & \MBv{seen}{imagebind_vision_probe}{top1} & \MBv{seen}{imagebind_vision_probe}{macro_f1} & \MBv{new}{imagebind_vision_probe}{top5} & \MBv{new}{imagebind_vision_probe}{top1} & \MBv{new}{imagebind_vision_probe}{macro_f1} \\
EgoVLP~\cite{EgoVLP} & \MBv{seen}{egovlp_probe}{top5} & \MBv{seen}{egovlp_probe}{top1} & \MBv{seen}{egovlp_probe}{macro_f1} & \MBv{new}{egovlp_probe}{top5} & \MBv{new}{egovlp_probe}{top1} & \textbf{\MBv{new}{egovlp_probe}{macro_f1}} \\
LaViLa~\cite{LaViLa} & \textbf{\MBv{seen}{lavila_probe}{top5}} & \textbf{\MBv{seen}{lavila_probe}{top1}} & \textbf{\MBv{seen}{lavila_probe}{macro_f1}} & \textbf{\MBv{new}{lavila_probe}{top5}} & \textbf{\MBv{new}{lavila_probe}{top1}} & \MBv{new}{lavila_probe}{macro_f1} \\
VideoMAE + CLIP $\star$ & \RSEgoVFive & \RSEgoVOne & \textbf{\RSEgoVF} & \RNEgoVFive & \RNEgoVOne & \RNEgoVF \\
\midrule
\multicolumn{7}{@{}l}{\textit{Egocentric video + audio}} \\
TIM~\cite{TIM} & \MBv{seen}{tim}{top5} & \MBv{seen}{tim}{top1} & \MBv{seen}{tim}{macro_f1} & \MBv{new}{tim}{top5} & \MBv{new}{tim}{top1} & \MBv{new}{tim}{macro_f1} \\
\midrule
\multicolumn{7}{@{}l}{\textit{Watch microphone}} \\
SSAST~\cite{SSAST} & \textbf{\MBv{seen}{ssast}{top5}} & \MBv{seen}{ssast}{top1} & \MBv{seen}{ssast}{macro_f1} & \MBv{new}{ssast}{top5} & \MBv{new}{ssast}{top1} & \MBv{new}{ssast}{macro_f1} \\
ImageBind~\cite{ImageBind} & \MBv{seen}{imagebind_audio_probe}{top5} & \MBv{seen}{imagebind_audio_probe}{top1} & \MBv{seen}{imagebind_audio_probe}{macro_f1} & \textbf{\MBv{new}{imagebind_audio_probe}{top5}} & \textbf{\MBv{new}{imagebind_audio_probe}{top1}} & \MBv{new}{imagebind_audio_probe}{macro_f1} \\
SAMoSA + AST $\star$ & \textbf{\RSMicVFive} & \textbf{\RSMicVOne} & \textbf{\RSMicVF} & \RNMicVFive & \RNMicVOne & \textbf{\RNMicVF} \\
\midrule
\multicolumn{7}{@{}l}{\textit{Watch IMU}} \\
DeepConvLSTM~\cite{DeepConvLSTM} & \textbf{\MBv{seen}{deepconvlstm}{top5}} & \MBv{seen}{deepconvlstm}{top1} & \MBv{seen}{deepconvlstm}{macro_f1} & \textbf{\MBv{new}{deepconvlstm}{top5}} & \MBv{new}{deepconvlstm}{top1} & \MBv{new}{deepconvlstm}{macro_f1} \\
ImageBind~\cite{ImageBind} & \MBv{seen}{imagebind_imu_probe}{top5} & \textbf{\MBv{seen}{imagebind_imu_probe}{top1}} & \MBv{seen}{imagebind_imu_probe}{macro_f1} & \MBv{new}{imagebind_imu_probe}{top5} & \textbf{\MBv{new}{imagebind_imu_probe}{top1}} & \MBv{new}{imagebind_imu_probe}{macro_f1} \\
Pose-pre-trained CNN $\star$ & \RSIMUVFive & \RSIMUVOne & \textbf{\RSIMUVF} & \RNIMUVFive & \RNIMUVOne & \textbf{\RNIMUVF} \\
\midrule
\multicolumn{7}{@{}l}{\textit{Egocentric video + watch + glasses (all modalities)}} \\
Our baseline, fused & \RSEgoWGVFive & \RSEgoWGVOne & \RSEgoWGVF & \RNEgoWGVFive & \RNEgoWGVOne & \RNEgoWGVF \\
\bottomrule
\end{tabular}
\end{table}

\paragraph{RQ3: the gains hold for new participants and on a stronger video model.}
For new participants, the wearables raise egocentric video from \RNEgoVFive\% to \RNEgoWGVFive\% verb top-5 (\RNEgoWGGainVFive; \cref{fig:spectrum}c) and from \RNEgoVF\% to \RNEgoWGVF\% macro F1 (\RNEgoWGGainVFOne).
The exocentric camera, placed differently in every home, does not help (\RNEgoExoGainVFive): alone it is far below egocentric video, and equal weights cannot discount it.
The gains are not an artifact of the baseline's video model: added by the same rule to LaViLa~\cite{LaViLa}, the strongest egocentric model we evaluate (\cref{tab:baselines}), the watch adds \MBg{seen}{lavila/watch}{top5} [\MBgl{seen}{lavila/watch}{top5}, \MBgh{seen}{lavila/watch}{top5}] top-5 for seen and \MBg{new}{lavila/watch}{top5} [\MBgl{new}{lavila/watch}{top5}, \MBgh{new}{lavila/watch}{top5}] for new participants, and the watch and the glasses add \MBg{seen}{lavila/all}{macro_f1} and \MBg{new}{lavila/all}{macro_f1} macro F1, both with intervals above zero.
More video does not do the same: a second egocentric video model adds \MBg{seen}{lavila/ego2}{top5} and \MBg{new}{lavila/ego2}{top5} top-5, and the exocentric camera \MBg{seen}{lavila/exo}{top5} and \MBg{new}{lavila/exo}{top5}.

\subsection{Benchmark results}
\label{sec:exp_benchmark}

\paragraph{Existing models per modality (\cref{tab:baselines}).}
We train existing video, audio-visual, audio, and IMU models on the benchmark's folds.
\emph{Egocentric pre-training transfers best to new people:} for new participants LaViLa reaches \MBv{new}{lavila_probe}{top5}\% verb top-5, against \MBv{new}{omnivore_probe}{top5}\% for Omnivore and \MBv{new}{imagebind_vision_probe}{top5}\% for ImageBind.
\emph{Adding modalities matters as much as the video model:} our fused baseline has the highest macro F1 on both splits.
\emph{Rare verbs remain open:} the four rarest, \emph{wipe}, \emph{press}, \emph{shake}, and \emph{move}, stay at about 10\% F1 or below for egocentric video, with or without the watch.

\paragraph{Trimmed actions are far from solved (\cref{tab:segments}a).}
Given an action's boundaries and its seconds of context, egocentric video recognizes the verb at \RSSegEgoVFive\% top-5 for seen participants, and the wearables add under a point: once the boundaries are known, video suffices for the verb.
The full action, verb and noun together, remains hard: the best setup without the exocentric camera reaches \RSSegAllAOne\% top-1 for seen participants and \RNSegAllAOne\% for new ones.

\paragraph{Detection and anticipation remain open (\cref{tab:segments}b).}
Temporal detection reaches at most \RSDetExo{} mAP for seen participants and only a few points for new ones, where the camera-free watch (\RNDetWatch) does as well as any setup with video and better than egocentric video alone (\RNDetEgo).
For anticipation, egocentric video alone reaches \RSAntEgo\% class-mean top-5 recall for seen and \RNAntEgo\% for new participants, and no wearable improves it significantly.
Both remain open challenges that EgoCARBS makes measurable.

\section{Conclusion}
\label{sec:conclusion}

We presented EgoCARBS, to our knowledge the first public dataset and benchmark that pairs wearable active acoustic sensing with egocentric and exocentric video.
Its central lesson is that the context an egocentric camera misses need not come from a camera in the room: wearable echoes sense motion without appearance, the complement of video, and travel with the wearer, so they keep helping for new people in new kitchens and on top of a strong egocentric model.
Because any wearable with a speaker and a microphone can sense this way, the same signal is within reach of earbuds and AR glasses, not only watches.
Trimmed actions, detection, anticipation, and rare verbs remain open, and invite models that follow echoes, video, and motion over time.

\paragraph{Limitations and future work.}
EgoCARBS spans 15 participants, each recorded in one kitchen, so the new-participant split changes the person and the kitchen together; our collection tools let it grow to more people and repeated sessions.
Its reference is one exocentric camera, read by frozen features and fused with equal weights; a multi-camera rig or a learned fusion would set a higher one.
Our baseline scores 1\,s windows offline with late fusion; causal, jointly trained models and self-supervised pre-training on unlabeled echoes are next steps.

\section*{LLM Usage Statement}
\label{sec:llm_statement}
We used LLMs in the research and in the writing of this paper.
\textbf{Claude Code} (Anthropic's coding agent, running Claude models) assisted with writing the experiment and model code and running the baselines (\cref{sec:experiments}); with improving our existing annotation tools; with providing initial annotations; with identifying and verifying the contemporary and competitive baselines to run, and checking related work against the cited papers; and with writing and editing the prose of every section.
\textbf{Google Gemini} (3.8 Flash) provided initial temporally localized action annotations from the synchronized egocentric and exocentric video, which we then reviewed (\cref{sec:dataset_annotation}).
We did not use LLMs to design or build the sensing hardware, or to run the user study and data collection.
The authors reviewed all LLM output and take full responsibility for the content of the paper.

{
    \small
    \setlength{\bibsep}{0pt plus 0.5pt}
    \bibliographystyle{ieeenat_fullname}
    \bibliography{main,supp}
}

\clearpage
\setcounter{section}{0}\setcounter{figure}{0}\setcounter{table}{0}\setcounter{equation}{0}
\renewcommand{\thesection}{\Alph{section}}
\renewcommand{\thefigure}{S\arabic{figure}}
\renewcommand{\thetable}{S\arabic{table}}
\renewcommand{\theequation}{S\arabic{equation}}
\renewcommand{\theHsection}{supp.\Alph{section}}
\renewcommand{\theHfigure}{S\arabic{figure}}
\renewcommand{\theHtable}{S\arabic{table}}
\renewcommand{\theHequation}{S\arabic{equation}}
\renewcommand{\arraystretch}{1.12}
\renewcommand{\topfraction}{0.95}\renewcommand{\dbltopfraction}{0.8}
\renewcommand{\bottomfraction}{0.6}\renewcommand{\textfraction}{0.05}
\renewcommand{\floatpagefraction}{0.85}\renewcommand{\dblfloatpagefraction}{0.7}
\makeatletter\setlength{\@fptop}{0pt}\setlength{\@dblfptop}{0pt}\makeatother
\setcounter{topnumber}{4}\setcounter{dbltopnumber}{4}\setcounter{totalnumber}{6}
\maketitlesupplementary
\startcontents[supp]

\noindent A head-worn camera sees what its wearer looks at, but not always what the hands do: they work below its view, behind a pot or an arm, or in motions too alike for a glimpse to tell apart.
EgoCARBS asks whether the devices people already wear can fill that gap.
A smartwatch and camera glasses become short-range sonars: each plays inaudible chirps and hears their echoes off the hands, the arms and the objects they hold.
EgoCARBS delivers a dataset and benchmark that pair this active acoustic sensing with egocentric and exocentric video, passive audio and motion, and a study of what the wearables add to egocentric video.

This supplement follows the data from the sensor to the result.
Sections~A to C document the dataset and benchmark: how each wearable stream is sensed and brought onto one clock (A), what is released (B), and how the benchmark is scored and its baseline built, with what it costs to run (C).
Sections~D and E document the study: the full evidence behind each of the main paper's research questions (D), and \ExamplesN{} examples that show every stream around one action (E).
Numbers without the prefix S refer to the main paper.

\vspace{0.6em}{\footnotesize\noindent\textbf{Contents}\par\vspace{0.3em}\printcontents[supp]{l}{1}{\setcounter{tocdepth}{2}}}

\section{Sensing system}
\label{sec:suppl_sensing}

\begin{figure*}[t]
\centering
\includegraphics[width=\textwidth]{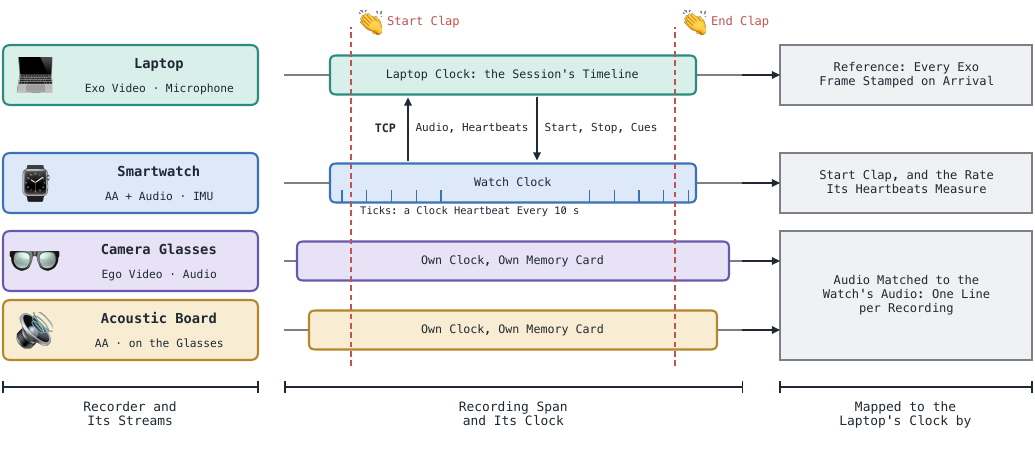}
\caption{\textbf{Four recorders, one clock.} Each lane is a recorder: its streams, its recording span and the clock that times it. The laptop owns the timeline and the watch streams to it over TCP; the glasses and their board record to their own cards and are aligned afterwards through their audio. Two claps bracket the session (\cref{sec:suppl_sync}).}
\label{fig:suppl_collection}
\end{figure*}

Beside egocentric video, the two wearables record three kinds of stream: active acoustics on the watch and on the glasses, passive audio from the same microphones, and the watch's IMU (\cref{tab:suppl_aa_params}).
Each is a different view of the same action: the echoes follow the hands and arms within about a metre, the audio hears what the action sounds like, and the IMU measures how one wrist moves and turns.
This section first derives what each stream measures (\cref{sec:suppl_theory,sec:suppl_passive,sec:suppl_imu}) and how the two sonars share one space without disturbing each other (\cref{sec:suppl_signal}).
It then describes how each wearable runs them (\cref{sec:suppl_watch,sec:suppl_board}), and how four recorders with four clocks are brought onto one timeline, so that a window cut at the same times shows the same moment in every stream (\cref{sec:suppl_laptop,sec:suppl_sync}; \cref{fig:suppl_collection}).

\subsection{Active acoustic sensing}
\label{sec:suppl_theory}

Active acoustic sensing turns a device's speaker and microphone into a short-range sonar: the speaker plays a known sound, every surface nearby reflects it, and the delay of each echo gives the distance of the surface that made it.
Repeated many times a second, this yields an echo profile, an image of echo strength over distance and time in which a moving hand leaves a trace (\cref{fig:sensing}).
Following C-FMCW~\cite{CFMCW}, we trace one chirp from the speaker to the envelope-difference profile the models read, and derive how finely the profile places a reflector and how fast a motion it can follow.
Both wearables run the same steps with the same chirp length and bandwidth, each in its own frequency band.

\begin{table}[t]
\centering
\caption{\textbf{Sensing parameters of the two wearables.} A value centred across both columns is shared. Formulas from \cref{sec:suppl_theory}; $c=343$\,m/s.}
\label{tab:suppl_aa_params}
\footnotesize
\setlength{\tabcolsep}{3.4pt}
\renewcommand{\arraystretch}{1.08}
\begin{tabular}{@{}c l c c@{}}
\toprule
& & \textbf{Smartwatch} & \textbf{Glasses} \\
\midrule
\tgroup{11}{Active acoustics}
& Speaker, microphone & built in & our board \\
& Chirp band & 18--21\,kHz & 21.5--24.5\,kHz \\
& Bandwidth \tform{B} & \multicolumn{2}{c}{3\,kHz} \\
& Samples per chirp \tform{N} & \multicolumn{2}{c}{600} \\
& Sampling rate \tform{f_s} & 48\,kHz & 50\,kHz \\
& Chirp period \tform{T=N/f_s} & 12.5\,ms & 12\,ms \\
& Band-pass passband & 17.8--21.2\,kHz & 21.5--24.5\,kHz \\
& Processing gain \tform{BT} & 16\,dB & 16\,dB \\
& Range resolution \tform{c/2f_s} & 3.57\,mm & 3.43\,mm \\
& Span read \tform{200\,c/2f_s} & 0.71\,m & 0.69\,m \\
& Phase period \tform{c/2f_c} & 8.8\,mm & 7.5\,mm \\
\midrule
\tgroup{4}{Passive audio}
& Microphone & built in & our board \\
& Resampling to 16\,kHz & $\div 3$ & $\times 8/25$ \\
& Log-mel bands & \multicolumn{2}{c}{64, 10\,Hz--8\,kHz} \\
& Window, hop & \multicolumn{2}{c}{0.6\,s, 30\,ms} \\
\midrule
\tgroup{3}{IMU}
& Sensors & accel., gyro., magn. & -- \\
& Rate & 100\,Hz $\to$ 50\,Hz & -- \\
& Orientation & 6-axis Madgwick~\cite{Madgwick} & -- \\
\bottomrule
\end{tabular}
\end{table}

\paragraph{Transmitted chirp.}
The speaker repeats a linear chirp, a tone whose frequency sweeps upward at a constant rate, with no gap between repetitions.
Within one period $T=N/f_s$, its frequency rises from $f_0$ to $f_0+B$:
\begin{equation}
s(t)=\cos\!\Big(2\pi\big(f_0 t+\tfrac{\mu}{2}t^2\big)\Big),\quad 0\le t<T,\quad \mu=B/T.
\label{eq:suppl_chirp}
\end{equation}
Both wearables use $N=600$ samples and $B=3$\,kHz:
\begin{itemize}
\item \emph{Watch}: 18--21\,kHz from its own speaker at $f_s=48$\,kHz, so $T=12.5$\,ms (80 chirps per second);
\item \emph{Glasses}: 21.5--24.5\,kHz from our board at $f_s=50$\,kHz, so $T=12$\,ms (83.3 chirps per second).
\end{itemize}
The bands lie above most adults' hearing, so the sensing is silent to the wearer, and below each device's Nyquist limit, so its microphone can record them.
They do not overlap, so both devices transmit at once (\cref{sec:suppl_signal}).
One chirp per period also fixes how often the scene is measured: 80 and 83.3 echo frames per second, fast enough to follow the strokes of cutting or stirring.

\paragraph{Received signal.}
A reflector at distance $R_i$ lengthens the sound's path by about $2R_i$, so its echo arrives $\tau_i$ after the direct path from speaker to microphone:
\begin{equation}
x(t)=\sum_i a_i\,s(t-\tau_0-\tau_i)+\nu(t),\qquad \tau_i=\frac{2R_i}{c},
\label{eq:suppl_received}
\end{equation}
with $\tau_0$ the delay of the direct path, the sound that reaches the microphone through the device and the air without reflecting; $a_i$ the echo's attenuation; $c=343$\,m/s; and $\nu$ the room's own sound.
Every echo is a copy of the same chirp, so only its delay and strength carry information: the delay says how far the reflector is, the strength how large and how reflective it is.

\paragraph{Band-pass filter.}
Kitchen sound is far louder below the chirp band than within it, so an order-8 Butterworth filter $g$ first keeps only the chirp band, leaving the audible sound to the passive stream:
\begin{equation}
r=g*x,\qquad \lvert G(f)\rvert^2=\frac{1}{1+\big(\frac{f^2-f_c^2}{f\,W}\big)^{16}},
\label{eq:suppl_bandpass}
\end{equation}
with passband $[f_\text{lo},f_\text{hi}]$, width $W=f_\text{hi}-f_\text{lo}$ and centre $f_c=\sqrt{f_\text{lo}f_\text{hi}}$:
\begin{itemize}
\item \emph{Watch}: 17.8--21.2\,kHz, 0.2\,kHz beyond its sweep on each side, so the band edges keep their strength;
\item \emph{Glasses}: 21.5--24.5\,kHz, exactly its sweep, which alone attenuates the watch's chirp 0.5\,kHz below by 25\,dB.
\end{itemize}
The filter delays every echo equally, so the spacing between echoes, which encodes distance, is intact.

\paragraph{Cross-correlation.}
Each chirp period of $r$ is correlated with the transmitted chirp (\cref{eq:echo_frame}):
\begin{equation}
p_n[k]=\sum_{m=0}^{N-1} r[nN+k+m]\,s[m],\qquad k=0,\dots,N-1.
\label{eq:suppl_xcorr}
\end{equation}
Correlating with the known chirp is a matched filter: it compresses each 12\,ms echo into a sharp peak at its lag, of height proportional to $a_i$, and lifts it above uncorrelated noise by the time--bandwidth product $BT$, 37.5 on the watch and 36 on the glasses, about 16\,dB.
This gain is what lets a quiet wearable speaker be heard over a noisy kitchen.
The output $p_n$ is the echo frame of chirp $n$: one column of the profile, echo strength against lag.

\paragraph{Range and resolution.}
As in C-FMCW~\cite{CFMCW}, a peak's lag is its delay counted in samples, $\text{Lag}_i=\tau_i f_s$, which gives the reflector's distance, its resolution and the profile's span:
\begin{equation}
R_i=\frac{c\,\text{Lag}_i}{2f_s},\quad
\delta R=\frac{c\,\delta\text{Lag}}{2f_s}=\frac{c}{2f_s},\quad
R_\text{max}=\frac{c\,N}{2f_s}.
\label{eq:suppl_range}
\end{equation}
Two numbers describe the profile's precision, and they differ.
The sampling rate sets how finely a single reflector is placed: one lag is 3.57\,mm on the watch and 3.43\,mm on the glasses, within spans of 2.14 and 2.06\,m.
The bandwidth sets how close two reflectors can be and still be told apart: surfaces less than $c/(2B)=5.7$\,cm apart merge into one peak, as in any FMCW sonar~\cite{CFMCW}.
The hand, the forearm and a held object are therefore separate in the profile, while the fingers of one hand are not.
Echoes stay measurable to about 1\,m, which covers both hands and the objects they handle.

\paragraph{Direct-path alignment.}
Where the direct path falls in a frame depends on when recording started relative to playback, so it differs between recordings.
Being the strongest peak, and fixed while the device is worn, it is easy to find, and we shift every frame to put it at lag 300:
\begin{equation}
k_0=\arg\max_k\sum_{n}\lvert p_n[k]\rvert,\qquad
d(k)=\frac{c\,(k-300)}{2f_s}.
\label{eq:suppl_direct}
\end{equation}
Lag $k$ is then a fixed distance $d(k)$ from the device; the models read lags 300--499, the first 0.71\,m (watch) and 0.69\,m (glasses).

\paragraph{Envelope and phase.}
The echo frame $p_n$ still oscillates at the chirp's carrier frequency, so a reflector that moves by a fraction of a wavelength can flip its sign.
The analytic signal separates this oscillation from the echo's strength:
\begin{equation}
\begin{aligned}
h_n&=p_n+j\,\mathcal{H}\{p_n\}=\lvert h_n\rvert\,e^{j\varphi_n},\\
\mathcal{F}\big\{\mathcal{H}\{p\}\big\}(f)&=-j\,\mathrm{sgn}(f)\,\mathcal{F}\{p\}(f).
\end{aligned}
\label{eq:suppl_envelope}
\end{equation}
The envelope $\lvert h_n\rvert$ is the echo strength at each distance, free of the carrier, and $\mathrm{Re}\,h_n=p_n$, so nothing is lost.
The phase $\varphi_n$ is finer but fragile: it turns once per $c/(2f_c)$ of motion, 8.8\,mm on the watch and 7.5\,mm on the glasses, and so wraps between chirps above about 0.35 and 0.31\,m/s, speeds hands often reach while cooking.
The models therefore read the envelope, which changes smoothly at any speed.
$\mathcal{H}$ runs along the whole stream rather than within each frame, so frame boundaries never cut the carrier.

\paragraph{Envelope difference.}
A static scene returns the same envelope every chirp, so the difference of consecutive envelopes (\cref{eq:diff}),
\begin{equation}
D_n[k]=\lvert h_n[k]\rvert-\lvert h_{n-1}[k]\rvert,
\label{eq:suppl_diff}
\end{equation}
is zero for every static reflector, and nonzero only where something moved.
This removes what dominates the envelope but says nothing about the action: the direct path, the counter, the wearer's own body and the device's fit on the wrist or head.
A hand at 1\,m/s moves 1.2\,cm, a few lags, between frames, so its trace in the profile is continuous rather than a scatter of jumps.
Its Doppler shift reads as a small, consistent extra delay, $vf_cT/B$ (about 4\,cm at 0.5\,m/s), which shifts a moving trace without breaking it and becomes part of the motion pattern.

\paragraph{Scaling.}
Each recording's envelope difference is divided by one number, its standard deviation over the lags the models read:
\begin{equation}
\tilde D_n[k]=\frac{D_n[k]}{\sigma},\qquad \sigma=\operatorname{std}_{n,\;300\le k<500} D_n[k].
\label{eq:suppl_scale}
\end{equation}
The scaling removes what differs between recordings, such as speaker volume and how the device sits, while one scale per recording keeps the relative strength of its moments.
Normalizing each window on its own would instead stretch a still moment to the level of a stir.

\subsection{Passive acoustic sensing}
\label{sec:suppl_passive}

Below the chirp band, the same microphone hears the kitchen: running water, cutlery, a knife on a board, oil in a pan.
This passive audio tells what an action sounds like, and so complements the echoes, which tell how the hands move.
We release it, and the models read it, as a log-mel spectrogram, the input of SAMoSA's smartwatch sound model~\cite{SAMoSA}; it is released in place of the raw audio, with speech removed (\cref{sec:dataset_recording}).
The watch uses its own microphone.
The glasses carry two, the camera's (32\,kHz, \cref{sec:system}) and the board's; the models read the board's, recorded on the same clock as the glasses' echoes, and the camera's aligns the video (\cref{sec:suppl_sync}).

\paragraph{Low-pass filter and resampling.}
We remove the chirp and everything above 8\,kHz with a linear-phase anti-aliasing filter $h_\text{lp}$ and resample to 16\,kHz, by $\div 3$ on the watch and $\times 8/25$ on the glasses:
\begin{equation}
x_{16}[m]=\big(h_\text{lp}*x\big)\!\left[\tfrac{f_s}{16\,\text{kHz}}\,m\right].
\label{eq:suppl_resample}
\end{equation}
Without the filter, the 18--21\,kHz chirp would fold onto 2--5\,kHz, and the passive stream would partly copy the active one.

\paragraph{Log-mel spectrogram.}
The spectrogram follows SAMoSA~\cite{SAMoSA}, so its pre-trained sound model reads our audio exactly as it was trained.
A short-time Fourier transform (periodic Hann window of 0.6\,s every 30\,ms) gives magnitudes $\lvert X_t(f)\rvert$, which 64 triangular filters $W_b$, evenly spaced in mel from 10\,Hz to 8\,kHz, pool into bands:
\begin{equation}
\begin{aligned}
M_t[b]&=\log\Big(\textstyle\sum_f W_b(f)\,\lvert X_t(f)\rvert+0.001\Big),\\
\text{mel}(f)&=1127\,\ln\!\big(1+f/700\,\text{Hz}\big),
\end{aligned}
\label{eq:suppl_logmel}
\end{equation}
at 33.3 frames per second.
SAMoSA~\cite{SAMoSA} reads 96 frames (2.88\,s) around each window; AST~\cite{AST} computes its own spectrogram from 3\,s of $x_{16}$, and its frozen features are released (\cref{sec:suppl_release}).

\subsection{Inertial sensing}
\label{sec:suppl_imu}

The IMU measures how the wrist moves: its accelerometer $\mathbf a$ senses linear acceleration plus gravity, its gyroscope $\boldsymbol\omega$ the rate of turn, and its magnetometer the surrounding magnetic field.
The watch records all three at 100\,Hz on its own clock (\cref{sec:suppl_watch}); we resample them to 50\,Hz on the laptop's clock (\cref{sec:suppl_sync}).

\paragraph{Orientation.}
Raw acceleration mixes the wrist's motion with gravity, so the same pour reads differently with the palm up or down.
A Madgwick filter~\cite{Madgwick} resolves this by tracking the wrist's orientation $\mathbf q$: it integrates the gyroscope, which is precise over short spans but drifts, and corrects the drift toward the orientation in which gravity points along the measured acceleration,
\begin{equation}
\mathbf q_{t+1}=\mathbf q_t+\Delta t\Big(\tfrac12\,\mathbf q_t\otimes(0,\boldsymbol\omega_t)-\beta\,\tfrac{\nabla f}{\lVert\nabla f\rVert}\Big),
\label{eq:suppl_madgwick}
\end{equation}
renormalized, with $\Delta t=20$\,ms, $\beta=0.1$, and $f$ the misfit between gravity rotated into the watch frame and $\mathbf a_t/\lVert\mathbf a_t\rVert$.
The magnetometer stays out of the fusion, since a kitchen's appliances and pans bend the field near the wrist, but it is released.

\paragraph{Model input.}
The models read the acceleration divided by 10\,m/s$^2$ and the orientation as a rotation vector divided by $\pi$: six channels of similar scale, 6$\times$50 values per 1\,s window, which tell both how the wrist moves and how it is held.

\subsection{Two active acoustic sensors in one space}
\label{sec:suppl_signal}

Both wearables chirp at once, a few tens of centimetres apart, and each microphone hears the other device as well as its own.
This subsection shows why neither disturbs the other, and how the same property lets the laptop check the glasses' board during recording.
Each device is described by one configuration file, shared by recording and processing, and processing refuses a recording whose chirp does not match it, so a profile is never computed with the wrong chirp.

\paragraph{Interference.}
The bands are disjoint, but the watch's chirp leaks into the glasses' band about 38\,dB down.
Because the two chirps repeat at incommensurate rates (80 and 83.3 per second), the other device's chirp lands at a different lag in every frame.
It therefore appears as faint speckle spread over the profile, rather than a fixed line that a model could mistake for a surface.

\paragraph{Board monitor.}
The board has no link to the laptop, but the watch hears it.
Its repetition rate lets the laptop confirm live that the board is transmitting: in the watch's audio, energy repeating at 83.3\,Hz stands 12--17\,dB above its neighbourhood when the board is on and $-$4 to $-$6\,dB when it is off.
The rate, not the energy in the board's band, is the reliable sign, because the watch's own chirp also leaks into that band.

\paragraph{In-band sound.}
Broadband sound that reaches into the chirp band, such as running water, appears as speckle across all lags, and so does the wrist's own motion.
The models receive the profile unfiltered and learn from data what separates the two, so that the sound of a running tap remains available as a cue to washing.

\subsection{Smartwatch implementation}
\label{sec:suppl_watch}

The watch needs no hardware change: our Wear OS app (Kotlin, Android 14) on a Samsung Galaxy Watch7 records all three wrist streams from the speaker, microphone and IMU it already has.
Two requirements shape the app: the chirp must leave the speaker with exact timing, and the microphone must record it unaltered.

\paragraph{Transmit.}
The app tiles the chirp into a 1\,s buffer of 80 chirps, seamless since 48{,}000 is a multiple of 600, and hands it to the audio hardware once, in a static loop; writing it in a streaming loop adds timing jitter that smears echoes across lags.

\paragraph{Receive.}
The microphone records at 48\,kHz, 16-bit, from Android's \texttt{UNPROCESSED} source, with noise suppression, gain control and echo cancellation disabled, since each alters the chirp.
On the wrist, the result is a clean signal: the correlation peak stands 56 times its median at exactly 600-sample spacing, and the chirp band 34\,dB above 8--17\,kHz.

\paragraph{IMU.}
The sensors are read at the fastest rate the watch allows, kept at 100\,Hz, and stamped with the watch's clock (\cref{sec:suppl_imu}).

\paragraph{Session control.}
One TCP link carries the audio live, a clock heartbeat every 10\,s that pairs the watch's sample count with its time (\cref{sec:suppl_sync}), and the IMU at the end; the laptop sends start, stop and cues.
The participant runs the session from the watch, but the laptop decides: recording begins only when the laptop answers Start, so every recorder it controls brackets the same interval.
The watch prompts only the session's bookends, each with a message and a vibration: a countdown and the start clap, and after Stop, the final clap; the cooking in between is never prompted.
The vibration matters most at the end, when the participant may be across the kitchen from the laptop.
Sessions are open-ended, with no script or timed blocks, and the back gesture is disabled so that a brush of the wrist cannot end one.

\subsection{Camera glasses implementation}
\label{sec:suppl_board}

The zShades-2K glasses record video and 32\,kHz audio to their own card but expose no speaker or microphone that software can drive, so active acoustics comes from our board on the temple (\cref{fig:hardware}b).
Its nRF52840 microcontroller plays the chirp and records the microphone at 50\,kHz, 16-bit, through a direct-memory-access ring of 90 blocks of 12\,ms, one block per chirp, which it copies to a microSD card every 9.72\,s.

\paragraph{Timekeeping.}
The board has no link to the laptop, so its only timeline is its sample index, and every sample must stay in its place.
The firmware therefore counts the blocks the ring completes against those it writes, so the stored stream keeps exactly one block per chirp period.

\subsection{Recording setup}
\label{sec:suppl_laptop}

A MacBook Pro runs the collection app: it receives the watch, records the exocentric camera (1280$\times$720, 30\,fps) and its own microphone, and shows the operator every stream's state, the board monitor included, so every stream is confirmed live during the session.
Each exocentric frame is stamped with the laptop's time on arrival and every frame is kept, so the frame index is the session's timeline, and the annotations, drawn on this video, share it.
We record with \texttt{mp4v}, which keeps every frame, and convert to H.264 afterwards with the frame count checked.

\subsection{Synchronization}
\label{sec:suppl_sync}

Four recorders run on four clocks (\cref{fig:suppl_collection}): the laptop, the watch, the glasses' camera and the board.
Clocks that nominally agree still drift apart by tens of parts per million, about 0.1\,s per hour, and the glasses and the board start whenever they are switched on.
We map every stream to the laptop's clock, the clock of the exocentric video and of the annotations (\cref{sec:system_sync}).
The watch audio carries both its active acoustics and its passive audio, so one mapping serves both, and, with its rate measured by the heartbeats, it serves as the session's audio clock.

\paragraph{Watch.}
Each heartbeat pairs the watch's sample count with the laptop's arrival time.
The median slope between them gives the watch's true sample rate, at most 79\,ppm from nominal (0.17\,s over 36\,min); the median ignores the heartbeats that a busy network delays.
The IMU is stamped differently, with the watch's wall-clock time, which ran 0.07--4.3\,s ahead of the laptop's and drifted; we follow it through the heartbeats' upper envelope and pin it at the start clap's spike in the IMU.

\paragraph{Claps.}
A clap is a sharp event that every stream records: a bang in the audio, a spike in the IMU and the hands meeting in the video.
The start clap is detected in the laptop's microphone, the exocentric video, the watch audio and the IMU, each search confined near the previous one (the watch within 0.15\,s of the microphone, the IMU within 0.3\,s of the watch), so that every stream takes the same clap rather than a nearby knock.
The end clap checks the mapping at the other end of the session.
It is projected from the start clap through the watch audio, and accepted only when two detectors agree on it and it is at least a quarter as loud as the start clap.

\paragraph{Glasses and board.}
The glasses and the board record to their own cards with no link to the laptop, but they hear the same kitchen as the watch.
We align their audio to the watch's with GCC-PHAT~\cite{GCCPHAT}, a cross-correlation that whitens the spectrum to sharpen the peak, over 20\,s windows every 30\,s, and fit one line $t_\text{laptop}=a\,t+b$ robustly to the delays: the slope $a$ absorbs the clock's drift, the offset $b$ its start.
Across the sessions, a median \SyncInlierMedian\% of the windows fit one line with \SyncRmsMin--\SyncRmsMax\,ms rms residual (median \SyncRmsMedian\,ms), well under one 100\,ms frame of the models' video, and the clocks ran \SyncGlassesPpmMin--\SyncGlassesPpmMax\,ppm (glasses) and \SyncBoardPpmMin--\SyncBoardPpmMax\,ppm (board) from the laptop's.

\paragraph{Camera latency.}
Audio places a recorder's sound on the laptop's clock, but not its pictures: exocentric frames are stamped 16--89\,ms after the light reaches the camera, and the glasses' video trails its own audio by 15--200\,ms.
We therefore render every session as one synchronized video, with every stream on the laptop's clock, and check, and where needed pin, the frame on which the hands meet at the clap in both views.
The same video is what annotators label (\cref{sec:suppl_rel_tools}).

\section{Dataset release}
\label{sec:suppl_release}

We release EgoCARBS in three parts, each for a different use: the dataset, to train and test new models (\cref{sec:suppl_rel_data}); the benchmark, to compare against ours on the same terms (\cref{sec:suppl_rel_bench}); and the tools that recorded and processed it, to record more (\cref{sec:suppl_rel_tools}).
Every stream is released on one clock, and every wearable stream in the form the models read, so a user needs none of our hardware to start.

\begin{table}[t]
\centering
\caption{\textbf{Released streams} per session, all on the session's clock (\cref{sec:suppl_sync}). Audio is released only as log-mel spectrograms.}
\label{tab:suppl_release}
\footnotesize
\setlength{\tabcolsep}{4pt}
\begin{tabular}{@{}l l l@{}}
\toprule
Stream & Released as & Rate, size \\
\midrule
\tgrp{3}{Smartwatch} \\
Active acoustics & echo profile (\cref{sec:suppl_theory}) & 80/s, 600 lags \\
Passive audio & log-mel (\cref{sec:suppl_passive}) & 33.3/s, 64 bands \\
IMU & accel., gyro., magnet. & 50\,Hz \\
\tgrp{3}{Camera glasses} \\
Egocentric video & faces blurred & 30\,fps, 2560$\times$1440 \\
Active acoustics & echo profile (\cref{sec:suppl_theory}) & 83.3/s, 600 lags \\
Passive audio & log-mel (\cref{sec:suppl_passive}) & 33.3/s, 64 bands \\
\tgrp{3}{Laptop} \\
Exocentric video & faces blurred & 30\,fps, 1280$\times$720 \\
\bottomrule
\end{tabular}
\end{table}

\subsection{Dataset}
\label{sec:suppl_rel_data}

\paragraph{Recordings.}
11.4\,h of unscripted cooking by 15 participants, one session each in their own kitchen (\cref{sec:dataset}), with the streams of \cref{tab:suppl_release}.
Every stream comes with its timestamps on the session's clock (\cref{sec:suppl_sync}), so a window cut at the same times shows the same moment in every stream.

\paragraph{Annotations.}
4,019 action segments with their times, verbs and objects as annotated, and their verb, noun and action classes (\cref{sec:dataset_annotation}).
The vocabulary maps every annotated name to a class and to EPIC-Kitchens-100's~\cite{EK100}, so models can be pre-trained on or compared with it.
The hand visibility of every 1\,s window, from three egocentric frames, supports the analysis of \cref{tab:suppl_visibility}.

\subsection{Benchmark}
\label{sec:suppl_rel_bench}

\paragraph{Tasks, splits and metrics.}
Window recognition, trimmed action recognition, temporal detection and anticipation, each for seen participants and for new ones, with the window grid, folds and scoring code (\cref{sec:suppl_protocol}), so that a new model is scored on exactly the windows and splits used here.

\paragraph{Baselines.}
The code of every model in \cref{sec:suppl_models}, its checkpoints for every fold and seed, the frozen features of every window, and every prediction behind \cref{sec:experiments} and \cref{sec:suppl_results}.
One command remakes every number, figure and table from the predictions in about an hour on a CPU.

\paragraph{Pre-trained wearable encoders.}
The echo and IMU encoders we pre-trained without EgoCARBS data (\cref{sec:suppl_pretrain}), the first public encoders for wearable echo profiles, as a starting point for active acoustic models that do not need a corpus of their own.

\subsection{Tools}
\label{sec:suppl_rel_tools}

\begin{itemize}
\item \emph{Sensing}: the Wear OS app that turns a commercial smartwatch into a sonar that also records audio and IMU (\cref{sec:suppl_watch}), and the firmware of the glasses' board (\cref{sec:suppl_board}).
\item \emph{Collection and processing}: the collection app (\cref{sec:suppl_laptop}), the processing of every wearable stream, configured per device (\cref{sec:suppl_theory,sec:suppl_passive,sec:suppl_imu}), and the synchronization of every recorder (\cref{sec:suppl_sync}), so that new sessions, devices and participants extend the same benchmark.
\item \emph{Annotation}: our multi-stream extension of VIDAT~\cite{VIDAT}, which shows every stream of a session on one timeline while annotators review the proposed segments.
\end{itemize}

\subsection{Privacy and intended use}
\label{sec:suppl_rel_terms}

\paragraph{Privacy.}
The study was approved by our institutional review board; every participant gave informed consent, was compensated, and may withdraw their session (\cref{sec:dataset_recording}).
Faces and other identifiers are blurred in both videos, speech is removed from the audio, which is released as log-mel spectrograms, and active acoustics is released as echo profiles only.

\paragraph{Intended use.}
EgoCARBS is for research on multimodal action understanding and wearable sensing, not for identifying or monitoring people.
Its split for new participants measures how far a model carries to people and kitchens it has never seen.

\section{Benchmark protocol and models}
\label{sec:suppl_models}

\begin{figure*}[t]
\centering
\includegraphics[width=\textwidth]{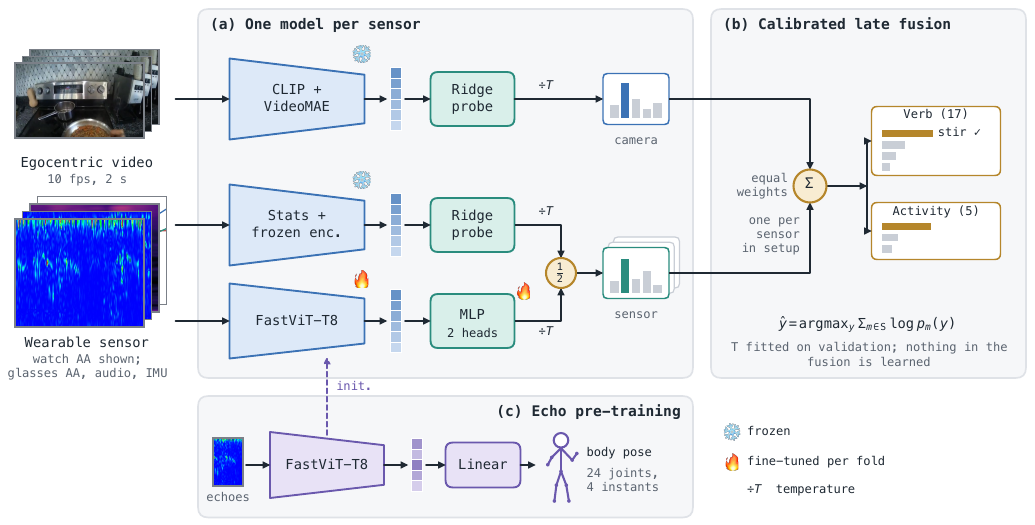}
\caption{\textbf{The baseline on 1\,s windows.} (a)~Every sensor has its own model: a camera, a ridge probe on frozen CLIP~\cite{CLIP} and VideoMAE~\cite{VideoMAE} features; a wearable sensor (the watch's active acoustics shown), a ridge probe on statistics and frozen encoders and a fine-tuned network. (b)~Logits are divided by a temperature $T$ fitted on validation, a sensor's two models are averaged, and a setup sums its sensors' log-probabilities with equal weights (\cref{eq:suppl_fusion}). (c)~The echo encoder is pre-trained on body pose from a separate motion-capture corpus. Bar charts are schematic.}
\label{fig:suppl_model}
\end{figure*}

The study asks what a sensor adds to egocentric video, and the answer is only as trustworthy as the comparison behind it.
The baseline (\cref{sec:exp_models}) is therefore built to make that comparison clean: every sensor has its own model, all trained by one recipe, and a setup, a set of sensors such as egocentric video with the watch, sums its sensors' calibrated outputs with equal weights (\cref{fig:suppl_model}).
Adding a sensor to a setup adds one term to the sum and changes nothing else, so no sensor's gain can come from fusion weights tuned in its favour.
This section gives the protocol that scores every model (\cref{sec:suppl_protocol}), the pre-trained models the baseline starts from (\cref{sec:suppl_pretrain}), the window model behind the main results (\cref{sec:suppl_window_models}), the models of the other tasks (\cref{sec:suppl_task_models}), the published models we compare against (\cref{sec:suppl_published}), and what each sensor costs to run (\cref{sec:suppl_compute}).

\subsection{Evaluation protocol}
\label{sec:suppl_protocol}

\paragraph{Windows and labels.}
The main task recognizes the verb of each 1\,s window, the delay at which an assistant would have to respond.
Each session is cut into 1\,s windows every 0.5\,s.
A window takes the label of the annotated segment it overlaps most, provided the overlap is at least 0.5\,s, and is scored if that segment's verb is one of the \VocabVerbClasses{} classes, the verbs that enough participants perform for long enough to train and test on (\cref{sec:dataset_annotation}).
The five activities group the verbs into coarser goals: knife work (cut, peel), washing (wash), stirring (stir), transport (pick, place, move, lift, adjust) and access (open, close, turn).

\paragraph{Folds.}
Two splits ask two questions.
\emph{Seen participants} asks how well a model recognizes a person and kitchen it has trained on, at other moments.
Each session is cut into 20 slices in time, which rotate through five folds (slice $k$ in fold $k \bmod 5$); fold $f$ is tested on its slices, validated on fold $f+1$'s and trained on the rest, so every participant appears in every fold and every slice is tested once.
\emph{New participants} asks how well a model carries to a person and kitchen it has never seen: one participant is tested at a time, trained on the others, and every tenth slice of the others validates.
In both, windows within 10\,s of a slice boundary are left out, so no training window shares context with a test window.
Every choice, from hyperparameters to which features a probe reads, is made on validation, and test windows are scored once.

\paragraph{Intervals.}
Participants differ more than moments do, so uncertainty is measured over participants: a gain's 95\% interval comes from 1,000 resamples of the test participants, with each resample applied to both setups compared, so the interval measures the gain itself rather than the spread of either setup.

\subsection{Pre-trained models}
\label{sec:suppl_pretrain}

With 15 participants, no stream has enough labelled data to train a large model from scratch, so every stream starts from a pre-trained model (\cref{tab:suppl_pretrained}), either frozen and read by a ridge probe (\frozen) or fine-tuned on each fold (\tuned).
Cameras and microphones use published models.
No public model reads wearable echo profiles, so we pre-train the echo and IMU encoders ourselves, on two smartwatch corpora from other studies, never on EgoCARBS data, and release them.
For the annotation, not the benchmark, MediaPipe Hands~\cite{MediaPipeHands} marks egocentric frames that show a hand, and Gemini~\cite{Gemini} proposes the segments that annotators review (\cref{sec:dataset_annotation}).

\begin{table*}[t]
\centering
\caption{\textbf{Pre-trained models of the baseline}, by the sensor they read. \frozen: frozen, read by a ridge probe; \tuned: fine-tuned on each fold. Tinted: pre-trained by us without EgoCARBS data, and released.}
\label{tab:suppl_pretrained}
\footnotesize
\setlength{\tabcolsep}{7pt}
\begin{tabular}{@{}l l l l c@{}}
\toprule
Model & Pre-training data & Objective & Input & Use \\
\midrule
\tgrp{5}{Egocentric and exocentric video} \\
CLIP ViT-L/14~\cite{CLIP} & image--text pairs & contrastive & frames at 10\,fps & \frozen \\
VideoMAE-B~\cite{VideoMAE} & Something-Something v2~\cite{SSv2} & action labels & 16 frames over 2\,s & \frozen \\
TSM~\cite{TSM} & EPIC-Kitchens-100~\cite{EK100} & action labels & trimmed segments & \tuned \\
\tgrp{5}{Watch and glasses microphones} \\
AST~\cite{AST} & AudioSet~\cite{AudioSet} & sound events & 3\,s of audio & \frozen \\
SAMoSA sound model~\cite{SAMoSA} (watch) & smartwatch audio & activities & log-mel, 2.88\,s & \frozen\,\tuned \\
\tgrp{5}{Watch and glasses active acoustics} \\
\tours FastViT-T8~\cite{FastViT} & ImageNet~\cite{ImageNet}, then 12.9\,h of echoes & body pose & echo profile, 1\,s & \frozen\,\tuned \\
\tours FastViT-T8~\cite{FastViT} & ImageNet~\cite{ImageNet}, then 10.3\,h of echoes & temporal contrast & echo profile, 1\,s & \frozen \\
\tgrp{5}{Watch IMU} \\
\tours 1D CNN & 12.9\,h of IMU & body pose & 1\,s at 50\,Hz & \frozen\,\tuned \\
\bottomrule
\end{tabular}
\end{table*}

\paragraph{Pose regression.}
Pose is a natural target for an echo encoder: the profile records how the arms and hands move, and pose describes exactly that.
A motion-capture corpus pairs 12.9\,h of a smartwatch's echoes (22 participants, 542 sessions, an 18--23.5\,kHz chirp) with full-body pose.
We compute its envelope-difference profile exactly as for EgoCARBS (\cref{sec:suppl_theory}) and cut 1\,s windows of lags 300--499 (200$\times$80).
A FastViT-T8~\cite{FastViT} with one input channel, initialized from ImageNet~\cite{ImageNet}, regresses the 24 body joints (pelvis-relative, heading removed) at four instants of the window under a mean per-joint distance loss, trained with AdamW~\cite{AdamW} (learning rate $10^{-3}$, weight decay 0.05) for 100 epochs of 50,000 windows; on two held-out participants it places the joints within 10.6\,cm on average.
From the same corpus's watch IMU, a three-layer 1D CNN (32, 64, 128 channels; 128-d output) over 1\,s of acceleration and orientation (6$\times$50) regresses the same pose (9.96\,cm).

\paragraph{Temporal contrast.}
A second echo encoder learns without labels, from 10.3\,h of echoes recorded with our watch and chirp in another study.
Two windows of one recording at most 2\,s apart form a positive pair and the rest of the batch serve as negatives, under InfoNCE~\cite{InfoNCE} at temperature 0.1, so the encoder learns what stays the same over a few seconds of one activity and what separates different moments.

\paragraph{Published encoders.}
For video, CLIP ViT-L/14~\cite{CLIP} reads every frame at 10\,fps, averaged over the window, and VideoMAE-B~\cite{VideoMAE}, fine-tuned on Something-Something v2~\cite{SSv2}, reads 16 frames of a 2\,s clip centred on it.
For audio, resampled to 16\,kHz (\cref{sec:suppl_passive}), AST~\cite{AST} gives its 768-d embedding and 527 AudioSet~\cite{AudioSet} class probabilities, and SAMoSA's~\cite{SAMoSA} convolutional trunk its pooled 1024-d features of the 2.88\,s log-mel.

\subsection{Window recognition model}
\label{sec:suppl_window_models}

Each sensor has its own classifier, and a setup $\mathcal{S}$ fuses them late, at the level of class probabilities (\cref{fig:suppl_model}):
\begin{equation}
\begin{aligned}
\hat y &= \arg\max_y \textstyle\sum_{m\in\mathcal{S}} \log p_m(y), \\
\log p_m &= \tfrac12\big(\log p_m^\text{probe} + \log p_m^\text{net}\big),
\end{aligned}
\label{eq:suppl_fusion}
\end{equation}
where each $p$ is a softmax of temperature-scaled logits, and a camera has only its probe.
A wearable sensor has two models because they complement each other: the probe is stable with few labels, and the network learns structure in the signal beyond fixed features (\cref{tab:suppl_components}).

\paragraph{Camera: ridge probe \frozen.}
A camera's probe reads its frozen CLIP~\cite{CLIP} and VideoMAE~\cite{VideoMAE} features: CLIP describes what is in view, frame by frame, and VideoMAE how it moves over 2\,s.

\paragraph{Wearable sensor: ridge probe \frozen.}
Each wearable sensor's probe reads two kinds of feature: hand-crafted statistics of the signal, which need no training, and the frozen encoders of \cref{sec:suppl_pretrain}, which carry what pre-training learned.
For active acoustics, it reads the envelope's and envelope difference's mean and spread in 10 range bands, motion-energy statistics, both pre-trained echo encoders, and, where validation prefers them, pooled range-Doppler images (how fast the reflectors at each distance move) and envelope-modulation images (the rhythm of each distance's echo strength); for the IMU, per-axis statistics, band powers and its encoder; for the microphones, log-mel statistics, AST~\cite{AST}, and SAMoSA~\cite{SAMoSA} on the watch.
Blocks are standardized on training windows, and wearable blocks wider than 256 are reduced to 256 principal components.
The probe is a class-balanced ridge regression solved in closed form, so it has no training schedule to tune, and its strength is chosen on validation macro F1.
Since an action usually lasts longer than one window, it may also read the mean features of the $\pm K$ windows around, with $K\in\{0,2,4,8\}$ chosen on validation.

\paragraph{Wearable sensor: fine-tuned network \tuned.}
Each wearable sensor also has a network: its pre-trained encoder (FastViT-T8~\cite{FastViT} for both active acoustic sensors, the 1D CNN for the IMU, SAMoSA's~\cite{SAMoSA} trunk for the watch microphone), an MLP (256 units, GELU, dropout 0.3), and verb and activity heads trained together.
The echo and IMU encoders read the window and its neighbours 1\,s before and after, so that a stroke cut by the window's edge is still seen whole.
Training uses AdamW~\cite{AdamW} (encoder $10^{-4}$, heads $10^{-3}$, weight decay 0.05) with a one-cycle schedule for 12 epochs in batches of 128, label smoothing 0.1, square-root inverse-frequency class weights, and augmentation (echoes: shifts of 5 lags and 8 frames, gain 0.8--1.25; IMU: gain 0.9--1.1; audio: one frequency and one time mask); we keep the best validation epoch of two seeds.

\paragraph{Calibration and fusion.}
Models differ in how confident they are, and an overconfident model would dominate an equal-weight sum.
Every model's logits are therefore divided by a temperature fitted on validation by log-likelihood, which makes its probabilities honest without changing its ranking of classes, and setups sum the calibrated log-probabilities with equal weights (\cref{eq:suppl_fusion}).
With nothing in the fusion fitted, a comparison between setups cannot ride on noise in tuned weights.
\Cref{tab:suppl_components} shows what each part contributes.

\subsection{Models for the other tasks}
\label{sec:suppl_task_models}

The benchmark's other tasks reuse the window models' features wherever they can, so that their results reflect the same sensors.

\paragraph{Trimmed action recognition.}
Given a whole annotated segment, the task names its verb, noun and action, as in EPIC-Kitchens-100~\cite{EK100}.
Each setup's model is the geometric mean of four members:
(i)~a segment network over the setup's window features pooled over the segment and the 4\,s around it;
(ii)~per-sensor logistic-regression probes on pooled segment features, fused as on windows;
(iii)~EPIC-Kitchens-100's TSM~\cite{TSM,EK100}, fine-tuned per fold from its released weights with its classifier mapped to our classes;
(iv)~a kernel ridge probe on VideoMAE~\cite{VideoMAE} segment features.
The action is the verb--noun pair with the highest product of probabilities.

\paragraph{Temporal detection.}
Detection finds where each activity starts and ends in an untrimmed session.
An MS-TCN~\cite{MSTCN} labels every 0.5\,s step with one of the five activities or background from the frozen window features of the setup's sensors, each sensor group with its own input projection and dropped at random during training; four configurations are compared on validation mAP, and the chosen one is trained with three seeds.

\paragraph{Anticipation.}
Anticipation names the next action before it starts, from what the sensors recorded just before.
Each sensor's probe reads its mean window features over the 2\,s that end 1\,s before an action, and probes are fused as on windows.

\subsection{Published baselines}
\label{sec:suppl_published}

Published models show where the baseline stands and give the benchmark reference numbers.
For \cref{tab:baselines,tab:suppl_baselines}, we trained published models for every sensor but active acoustics, for which none exists, on the benchmark's folds and scored them on its test windows, exactly as the baseline is scored.
Frozen backbones are read by a linear head trained full-batch with AdamW~\cite{AdamW}; VideoMAE~\cite{VideoMAE} and SSAST~\cite{SSAST} are fine-tuned, DeepConvLSTM~\cite{DeepConvLSTM} is trained from scratch, and TIM~\cite{TIM} is trained on frozen Omnivore~\cite{Omnivore} and Auditory SlowFast~\cite{AuditorySlowFast} features with 20\,s of context.
ImageBind's~\cite{ImageBind} IMU encoder receives 10\,s of acceleration and rotation rate in physical units, upsampled from 50 to 200\,Hz.

\subsection{Model size and compute}
\label{sec:suppl_compute}

A wearable sensor is worth adding only if it is cheap next to the camera, since a wearable's battery and processor are small.
\Cref{tab:suppl_compute} lists, per sensor, the models the baseline runs on a 1\,s window, their parameters and multiply-accumulates (MACs), and the verb top-5 gain the sensor brings to egocentric video.

\begin{table*}[t]
\centering
\caption{\textbf{Model size and compute per sensor} on one 1\,s window, beside the sensor's verb top-5 gain over egocentric video (points). Coloured: the 95\% interval excludes zero. Grey: the exocentric reference.}
\label{tab:suppl_compute}
\footnotesize
\setlength{\tabcolsep}{6pt}
\begin{tabular}{@{}l l r r r c c@{}}
\toprule
& & & & & \multicolumn{2}{c}{Verb top-5 gain} \\
\cmidrule(l){6-7}
Input & Models on each window & Params (M) & GMACs & \% of ego & Seen & New \\
\midrule
\SuppComputeRows
\bottomrule
\end{tabular}
\end{table*}

\paragraph{Active acoustics costs a thousandth of the camera.}
Egocentric video takes \CompCamParams\,M parameters and \CompCamGMAC{} GMACs per window, mostly CLIP ViT-L/14~\cite{CLIP} on ten frames.
The watch's active acoustics takes \CompAAParams\,M parameters and \CompAAGMAC{} GMACs, on FastViT-T8~\cite{FastViT}, a network designed for mobile devices.
Both active acoustic sensors together cost \CompAAPct\% of the camera's operations, and add more to it than a second, exocentric camera does: $+$4.0 against $+$3.1 points of verb top-5 for seen participants, and $+$7.9 against $-$6.5 for new ones, where the second camera lowers accuracy at another \CompCamGMAC{} GMACs (\cref{tab:suppl_compute}).

\paragraph{The microphones' cost is AST.}
Each microphone's probe also reads AST~\cite{AST}, \CompASTGMAC{} GMACs per window; everything else the wearables run takes \CompWGNoASTPct\% of the camera's operations.

\section{Additional results}
\label{sec:suppl_results}

The main paper answers three research questions with verb top-5 accuracy (\cref{sec:exp_results}).
This section gives the full evidence behind each answer, on every metric and every combination of sensors: the wearables close the gap to the ego-exo reference (RQ1, \cref{sec:suppl_res_rqone}), active acoustics adds what the watch's audio and IMU miss (RQ2, \cref{sec:suppl_res_rqtwo}), and the gains hold for new participants and a stronger video model (RQ3, \cref{sec:suppl_res_rqthree}).
It ends with the benchmark's other tasks, which remain open (\cref{sec:suppl_res_tasks}).

\paragraph{Reading the tables.}
AA is active acoustics and mic a device's passive audio; the whole watch is its AA, mic and IMU.
Grey rows are egocentric video, against which every gain is measured; blue rows are the full system, egocentric video with the watch and the glasses; grey type marks the exocentric reference, which no deployed system has.
Numbers are percentages or points, with 95\% participant-bootstrap intervals (\cref{sec:suppl_protocol}).

\subsection{RQ1: Gap to the ego-exo reference}
\label{sec:suppl_res_rqone}

The ego-exo reference adds a room camera to egocentric video: the most complete visual record, but one that cannot travel with the wearer.
RQ1 asks how much of its gain the wearables recover, and where.

\begin{table*}[t]
\begin{minipage}[t]{\columnwidth}
\centering
\caption{\textbf{Every input set on 1\,s windows}: verb top-1 accuracy and macro F1, and five-activity macro F1 (\%). Bold: a setup added to egocentric video whose gain over it has an interval above zero.}
\label{tab:suppl_windows}
\footnotesize
\setlength{\tabcolsep}{3.4pt}
\begin{tabular}{@{}lcccccc@{}}
\toprule
& \multicolumn{3}{c}{Seen participants} & \multicolumn{3}{c}{New participants} \\
\cmidrule(lr){2-4}\cmidrule(l){5-7}
& \multicolumn{2}{c}{Verb} & Act. & \multicolumn{2}{c}{Verb} & Act. \\
\cmidrule(lr){2-3}\cmidrule(lr){4-4}\cmidrule(lr){5-6}\cmidrule(l){7-7}
Input & Top-1 & F1 & F1 & Top-1 & F1 & F1 \\
\midrule
\SuppWindowRows
\bottomrule
\end{tabular}
\end{minipage}\hfill
\begin{minipage}[t]{\columnwidth}
\centering
\caption{\textbf{Gain by hand visibility.} Verb accuracy (\%) of egocentric video and each setup's gain over it (points), by how many of three egocentric frames of a window show a hand: none (\RSHandShareNone\% of test windows), some (\RSHandShareSome\%) or all (\RSHandShareAll\%). Coloured: the interval excludes zero.}
\label{tab:suppl_visibility}
\footnotesize
\setlength{\tabcolsep}{3.6pt}
\begin{tabular}{@{}lcccccc@{}}
\toprule
& \multicolumn{3}{c}{Seen participants} & \multicolumn{3}{c}{New participants} \\
\cmidrule(lr){2-4}\cmidrule(l){5-7}
Hands visible in & None & Some & All & None & Some & All \\
\midrule
\SuppVisibilityRows
\bottomrule
\end{tabular}
\end{minipage}
\end{table*}

\begin{figure*}[t]
\centering
\includegraphics[width=\textwidth]{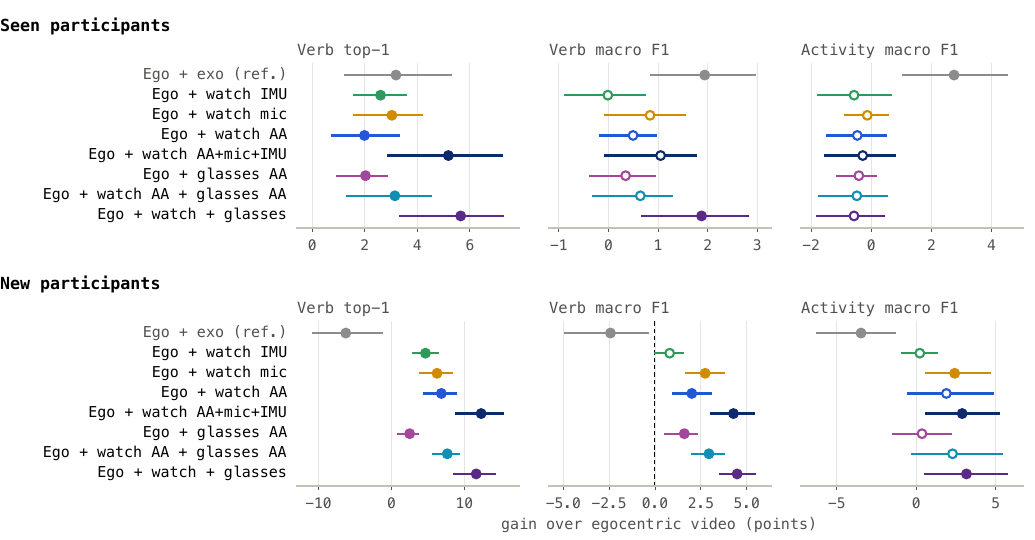}
\caption{\textbf{Gain over egocentric video on further window metrics}: verb top-1 accuracy, verb macro F1 and activity macro F1, for seen and new participants. Whiskers: 95\% intervals; filled: the interval excludes zero. The exocentric reference is grey.}
\label{fig:suppl_gains}
\end{figure*}

\paragraph{The headline holds on every verb metric.}
In verb top-5 accuracy (\cref{fig:spectrum}), the watch and the glasses raise egocentric video from \RSEgoVFive\% to \RSEgoWGVFive\% for seen participants, above the \RSEgoExoVFive\% of the ego-exo reference; the watch's active acoustics alone adds \RSEgoAAGainVFiveShort{} of the reference's \RSEgoExoGainVFiveShort{} points, about two-thirds of the gap.
\Cref{tab:suppl_windows,fig:suppl_gains} add top-1 accuracy and macro F1: the watch and the glasses raise verb top-1 by \RSEgoWGGainVOneShort{} points for seen participants and \RNEgoWGGainVOneShort{} for new ones, and verb macro F1, which weighs rare verbs as much as common ones, by \RSEgoWGGainVFOne{} and \RNEgoWGGainVFOne.
The exocentric camera adds less for seen participants (\RSEgoExoGainVOneShort{} top-1) and lowers both metrics for new ones (\RNEgoExoGainVOneShort), so on every verb metric the wearables match or exceed a second camera.
The watch's active acoustics alone raises top-1 in both splits (\RSEgoAAGainVOneShort{} and \RNEgoAAGainVOneShort).
For the five coarse activities, which the camera can often tell from the scene alone, egocentric video is already strong for seen participants (\RSEgoActOne\% top-1), and the watch and the glasses add \RNEgoWGGainActOne{} points of top-1 for new ones.

\paragraph{Every wearable adds to egocentric video.}
For new participants, every wearable setup in \cref{tab:suppl_windows}, from a single sensor to both wearables, raises egocentric video's verb top-1 with an interval above zero; for seen participants, every setup but the glasses' microphone does.
The gains carry over to the ego-exo reference itself: adding the whole watch to egocentric and exocentric video raises verb top-1 from 43.3\% to 46.9\% for seen participants and from 21.0\% to 32.6\% for new ones.

\paragraph{The watch alone rivals the camera.}
Without any camera, the whole watch already exceeds egocentric video in verb top-1 accuracy, 43.1\% against 40.2\% for seen participants and 37.6\% against 27.3\% for new ones, and the watch's active acoustics alone reaches 31.7\% for new participants, also above egocentric video (\cref{tab:suppl_windows}).
In macro F1, which weighs rare verbs equally, the whole watch matches egocentric video for new participants (16.8\% against 16.2\%).
The two are complementary: together they exceed either alone on every verb metric, in both splits.

\paragraph{The gain is largest where the camera does not see the hands.}
If the wearables help because they sense the hands when the camera does not, their gain should be largest when no hand is in view.
\Cref{tab:suppl_visibility} tests this, extending \cref{fig:hands} to every setup, by splitting the test windows by how many of three egocentric frames show a hand.
In the \RSHandShareNone\% of test windows where no egocentric frame shows a hand, the watch and the glasses add \RSHandWGNone{} points of verb top-5 for seen participants and \RNHandWGNone{} for new ones, about twice their gain where every frame shows a hand (\RSHandWGAll{} and \RNHandWGAll).
In both splits, every wearable setup gains most where no frame shows a hand, the watch's active acoustics alone included.
In verb top-1 (\cref{tab:suppl_visibility}b), the gain of the watch and the glasses for new participants is as large where every frame shows a hand (+11.6) as where none does (+11.7): for a person the model has never seen, the wearables help whether or not the hands are in view.

\subsection{RQ2: Active acoustics beyond audio and IMU}
\label{sec:suppl_res_rqtwo}

A smartwatch already records audio and motion, so active acoustics is worth adding only if it carries information they do not.
RQ2 tests this three ways: by adding it to the other watch sensors, by breaking its timing, and by asking which verbs it helps.

\begin{table*}[t]
\begin{minipage}[t]{\columnwidth}
\centering
\caption{\textbf{Time-shuffle control.} Macro F1 (\%) with active acoustics aligned (Al.) and shuffled among the same participant's windows (Sh., 10 permutations), and the change (points). Coloured: the interval excludes zero.}
\label{tab:suppl_shuffle}
\footnotesize
\setlength{\tabcolsep}{3.4pt}
\begin{tabular}{@{}lcccccc@{}}
\toprule
& \multicolumn{3}{c}{Seen participants} & \multicolumn{3}{c}{New participants} \\
\cmidrule(lr){2-4}\cmidrule(l){5-7}
Input & Al. & Sh. & $\Delta$ & Al. & Sh. & $\Delta$ \\
\midrule
\SuppShuffleRows
\bottomrule
\end{tabular}
\end{minipage}\hfill
\begin{minipage}[t]{\columnwidth}
\centering
\caption{\textbf{The two models of each wearable sensor}, alone and averaged, on 1\,s windows: verb top-5, top-1 and macro F1 (\%). \frozen: frozen features; \tuned: fine-tuned. Bold: best of each sensor.}
\label{tab:suppl_components}
\footnotesize
\setlength{\tabcolsep}{3.4pt}
\begin{tabular}{@{}lcccccc@{}}
\toprule
& \multicolumn{3}{c}{Seen participants} & \multicolumn{3}{c}{New participants} \\
\cmidrule(lr){2-4}\cmidrule(l){5-7}
Model & Top-5 & Top-1 & F1 & Top-5 & Top-1 & F1 \\
\midrule
\SuppComponentRows
\bottomrule
\end{tabular}
\end{minipage}
\end{table*}

\paragraph{It adds to the microphone and the IMU.}
Without a camera, adding active acoustics to the watch's microphone and IMU raises verb top-1 by \RSAAonMicIMUVOne{} points for seen participants and \RNAAonMicIMUVOne{} for new ones; with egocentric video present, it still adds \RNAAonEgoMicIMUVOne{} for new participants (\cref{tab:suppl_windows}).

\paragraph{It is tied to the moment, not the person.}
A model could gain from active acoustics without reading the action, by recognizing the person or the kitchen from how their echoes look.
Replacing each window's active acoustics with that of another window of the same participant keeps the person, the kitchen and the verb frequencies, and breaks only the link to the moment (\cref{fig:aa}b).
Macro F1 then falls in every row of \cref{tab:suppl_shuffle}, for verbs and activities, seen and new participants, every drop with an interval above zero: what active acoustics adds is what happens in that second.
The drop is largest without a camera, where active acoustics carries more of each decision: 7.0 and 7.2 points of activity macro F1 for the whole watch, for seen and new participants.

\begin{figure*}[t]
\centering
\includegraphics[width=\textwidth]{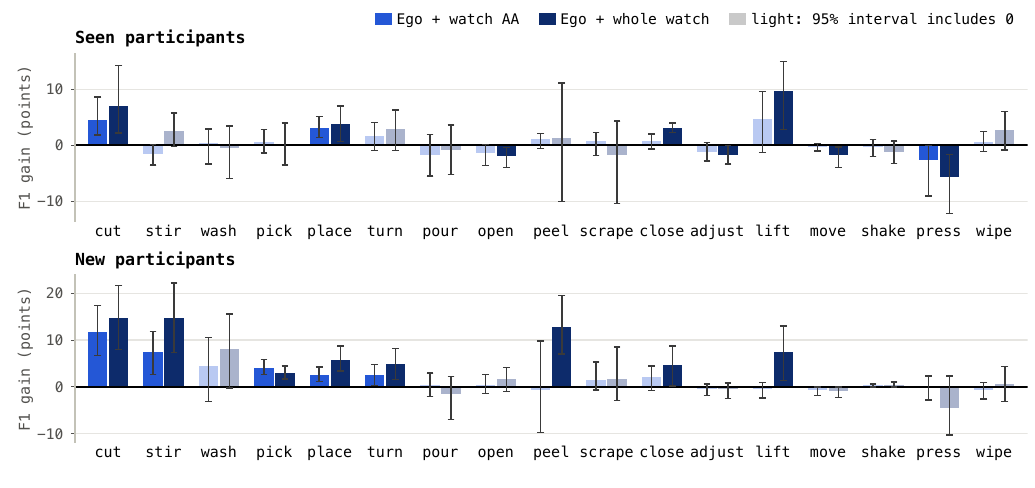}
\caption{\textbf{Per-verb F1 gain} of egocentric video + the watch's active acoustics (top) and + the whole watch (bottom), over egocentric video, on 1\,s windows; verbs ordered by frequency. Dark bars: the 95\% interval excludes zero.}
\label{fig:suppl_perverb}
\end{figure*}

\paragraph{It helps the verbs the hands perform.}
For new participants, the watch's active acoustics gains significantly on cut, stir, pick, place and turn, and the whole watch adds peel, lift and close (\cref{fig:suppl_perverb}): actions of the hand at the object, which a head-worn camera often sees from above or not at all.

\paragraph{Its signal is in the echo profile.}
The baseline reads active acoustics two ways (\cref{sec:suppl_window_models}), and comparing them shows where its information lies.
The network that reads the echo profile directly beats the probe on statistics and frozen encoders by about 5 points of verb top-5 for seen and 11 for new participants (\cref{tab:suppl_components}): the profile carries structure that end-to-end models find, and that better active acoustic models can exploit further.
Averaging the two models improves macro F1 for every sensor, most for the watch's microphone with new participants, by 2.9 points of top-5 and 2.3 of macro F1 over its network alone.

\subsection{RQ3: New participants and a stronger video model}
\label{sec:suppl_res_rqthree}

RQ3 asks whether the gains generalize, in two directions: to people and kitchens held out entirely, and on top of the strongest published egocentric-video model.

\paragraph{The gains grow for new participants.}
Every setup with the watch gains more for new participants than for seen ones (\cref{tab:suppl_windows,fig:suppl_gains}), while the exocentric camera, placed differently in every home, does not help.
A new kitchen changes what the camera sees more than how the hands move to cut or stir, consistent with the wearables mattering most where egocentric video is weakest.

\paragraph{The baseline holds against published models.}
\Cref{tab:suppl_baselines} extends \cref{tab:baselines} to every published model we ran (\cref{sec:suppl_published}), on all \MBParticipants{} participants and \MBSeenWindows{} test windows per split.
For new participants, LaViLa's~\cite{LaViLa} egocentric pre-training transfers best among video models (\MBv{new}{lavila_probe}{top5}\% verb top-5), which is why we test the wearables on top of it below.
For the watch's microphone and IMU, the baseline's model has the highest macro F1 of all the models for that sensor, in both splits.
Among the IMU models, DeepConvLSTM~\cite{DeepConvLSTM}, trained from scratch, reaches the highest top-5 accuracy, while the baseline's pose-pre-trained CNN leads in macro F1, the metric that counts the rare verbs.
The strongest published model overall is TIM~\cite{TIM}, which reads egocentric video together with audio: 77.4\% verb top-5 for seen and 71.8\% for new participants, above every video-only model, itself evidence that a second modality beside the camera pays off.

\paragraph{The wearables add to the strongest video model.}
Added to LaViLa~\cite{LaViLa} by the same equal-weight rule (\cref{tab:suppl_strong_ego}), the watch adds \MBg{seen}{lavila/watch}{top5} points of verb top-5 for seen and \MBg{new}{lavila/watch}{top5} for new participants, and its active acoustics alone \MBg{seen}{lavila/aa}{top5} and \MBg{new}{lavila/aa}{top5}.
More video does not do the same: a second egocentric model adds \MBg{seen}{lavila/ego2}{top5} and \MBg{new}{lavila/ego2}{top5}, and the exocentric camera \MBg{new}{lavila/exo}{top5} for new participants.
\Cref{tab:suppl_strong_ego} also shows how consistent the gains are.
Added to the baseline's own egocentric-video model, every wearable setup, from the watch's active acoustics alone to both wearables, gains in verb top-5 and top-1, for seen and new participants, each with an interval above zero.
Added to LaViLa, the watch's microphone and IMU and the whole watch gain in all four columns.
The wearables bring information that more video, or better video, does not.

\begin{table*}[p]
\centering
\renewcommand{\arraystretch}{1.0}
\caption{\textbf{The wearables added to a stronger egocentric-video model.} Gain in \VocabVerbClasses{}-verb top-5 and top-1 accuracy (points, with its 95\% interval) from adding each setup's baseline models to the baseline's own egocentric-video model and to LaViLa, by the same calibrated, equal-weight sum; windows and models as in \cref{tab:suppl_baselines}. Coloured: the interval excludes zero. ``Second ego model'' adds the baseline's egocentric-video model to LaViLa: what any second video model adds.}
\label{tab:suppl_strong_ego}
\footnotesize
\setlength{\tabcolsep}{7pt}
\begin{tabular}{@{}lcccc@{}}
\toprule
& \multicolumn{2}{c}{Seen participants} & \multicolumn{2}{c}{New participants} \\
\cmidrule(lr){2-3}\cmidrule(l){4-5}
Added & Verb top-5 & Verb top-1 & Verb top-5 & Verb top-1 \\
\midrule
\SuppStrongEgoRows
\bottomrule
\end{tabular}
\par\vspace{8pt}
\begin{minipage}[t]{\columnwidth}
\centering
\caption{\textbf{Published models per sensor} on 1\,s windows, on the benchmark's folds and labels: \VocabVerbClasses{}-verb top-5 and top-1 accuracy and macro F1 (\%) over all \MBParticipants{} participants, mean of 2 seeds ($^\dagger$one seed). \frozen: frozen, with a linear probe; \tuned: fine-tuned; \scratch: trained from scratch (TIM on frozen features). $\star$ (tinted): the baseline's model for that input. Bold: best of each input.}
\label{tab:suppl_baselines}
\footnotesize
\setlength{\tabcolsep}{2.7pt}
\begin{tabular}{@{}lcccccc@{}}
\toprule
& \multicolumn{3}{c}{Seen participants} & \multicolumn{3}{c}{New participants} \\
\cmidrule(lr){2-4}\cmidrule(l){5-7}
Model & Top-5 & Top-1 & F1 & Top-5 & Top-1 & F1 \\
\midrule
\SuppBaselineRows
\bottomrule
\end{tabular}
\end{minipage}\hfill
\begin{minipage}[t]{\columnwidth}
\centering
\caption{\textbf{Trimmed action recognition}, EPIC-Kitchens-100's protocol~\cite{EK100}: top-1 and top-5 accuracy (\%) of verb, noun and action over \RSSegN{} segments. AA: active acoustics of the watch and the glasses. Bold: best setup with egocentric video and without the exocentric camera.}
\label{tab:suppl_segments}
\footnotesize
\setlength{\tabcolsep}{3.2pt}
\begin{tabular}{@{}lcccccc@{}}
\toprule
& \multicolumn{2}{c}{Verb} & \multicolumn{2}{c}{Noun} & \multicolumn{2}{c}{Action} \\
\cmidrule(lr){2-3}\cmidrule(lr){4-5}\cmidrule(l){6-7}
Input & @1 & @5 & @1 & @5 & @1 & @5 \\
\midrule
\SuppSegmentRows
\bottomrule
\end{tabular}
\end{minipage}
\end{table*}

\subsection{Other benchmark tasks}
\label{sec:suppl_res_tasks}

Beyond window recognition, the benchmark defines three tasks that follow EPIC-Kitchens-100~\cite{EK100} (\cref{sec:suppl_task_models}).
Trimmed action recognition names the verb, noun and action of a given segment; temporal detection finds where each activity starts and ends in an untrimmed session; and anticipation names the next action 1\,s before it starts.
Each asks more of a model than a 1\,s window does, either a longer context or a prediction of what has not happened yet, and we give reference results for every setup.

\begin{table*}[t]
\begin{minipage}[t]{\columnwidth}
\centering
\caption{\textbf{Temporal detection}: mAP over temporal IoU 0.1--0.5 and segment F1 (\%), mean $\pm$ standard deviation over 3 seeds.}
\label{tab:suppl_detect}
\footnotesize
\setlength{\tabcolsep}{3.4pt}
\begin{tabular}{@{}lcccc@{}}
\toprule
& \multicolumn{2}{c}{Seen participants} & \multicolumn{2}{c}{New participants} \\
\cmidrule(lr){2-3}\cmidrule(l){4-5}
Input & mAP & F1 & mAP & F1 \\
\midrule
\SuppDetectRows
\bottomrule
\end{tabular}
\end{minipage}\hfill
\begin{minipage}[t]{\columnwidth}
\centering
\caption{\textbf{Anticipation} of the next action 1\,s before it starts: class-mean top-5 recall (\%) of verb, noun and action, over \RSAntN{} actions. Wearables alone predict only the verb.}
\label{tab:suppl_anticipation}
\footnotesize
\setlength{\tabcolsep}{3.6pt}
\begin{tabular}{@{}lcccccc@{}}
\toprule
& \multicolumn{3}{c}{Seen participants} & \multicolumn{3}{c}{New participants} \\
\cmidrule(lr){2-4}\cmidrule(l){5-7}
Input & Verb & Noun & Act. & Verb & Noun & Act. \\
\midrule
\SuppAnticipationRows
\bottomrule
\end{tabular}
\end{minipage}
\end{table*}

\paragraph{Trimmed actions.}
Without any camera, the watch and the glasses recognize the verb of a trimmed action at \RSSegWearVFive\% top-5 for seen and \RNSegWearVFive\% for new participants, above the exocentric camera alone for new participants on verbs, nouns and actions (\cref{tab:suppl_segments}).
Nouns come from the camera, which sees what the objects are, while the wearables sense how the hands move; adding them keeps noun accuracy at or above that of egocentric video alone, in both splits.
For new participants, the watch and the glasses raise verb, noun and action top-1 alike (\cref{tab:suppl_segments}).
With a whole segment, egocentric video also has many frames in which to see the hands, so the wearables' contribution is largest at the 1\,s scale of \cref{sec:suppl_res_rqone}, where a decision cannot wait for the hands to come into view.

\paragraph{Detection and anticipation.}
Both are open challenges that the benchmark poses for every input, cameras included (\cref{tab:suppl_detect,tab:suppl_anticipation}): detection mAP stays below 14\% for the activities, and no setup's class-mean top-5 recall of the next action exceeds 11\%.
Detection must place boundaries in sessions where activities interleave and pause without warning, and it is far harder for the 17 verbs than for the five activities, whose segments are longer.
Anticipation must name an action from what precedes it alone.
In segment F1 for new participants, the camera-free watch (29.6\%) also exceeds egocentric video (24.5\%) and every setup with video (\cref{tab:suppl_detect}).
For new participants, the camera-free watch already detects the activities as well as any setup with video, and better than egocentric video alone (\cref{tab:suppl_detect}).
Both call for models that reason over time across sensors, which the synchronized streams of EgoCARBS make possible to study.

\section{Qualitative examples}
\label{sec:suppl_examples}

\begin{figure*}[p]
\centering
\includegraphics[width=\textwidth]{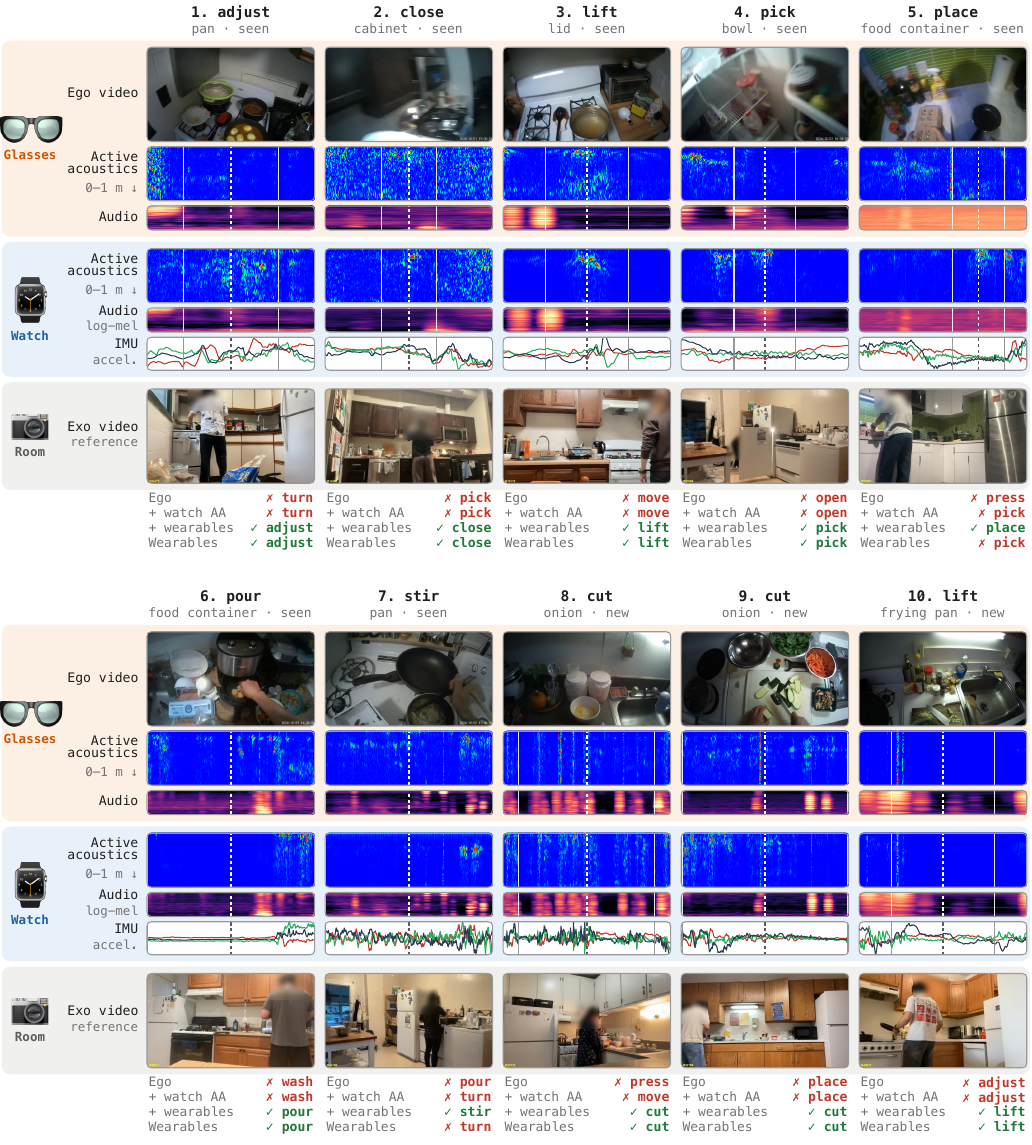}
\caption{\textbf{Examples 1--10: Hands out of view: egocentric video wrong, the wearables right.} Each column is one action. Rows: the glasses' egocentric frame, active acoustics and audio; the watch's active acoustics, audio and IMU; the exocentric frame. Solid lines bound the action, the dashed line marks the frame. Below: the verb named by egocentric video (Ego), with the watch's active acoustics (+ watch AA), with the watch and the glasses (+ wearables), and by the wearables alone; green right, red wrong. Groups: \cref{tab:suppl_examples}.}
\label{fig:suppl_examples_1}
\label{fig:suppl_ex_hands}
\end{figure*}

\begin{figure*}[p]
\centering
\includegraphics[width=\textwidth]{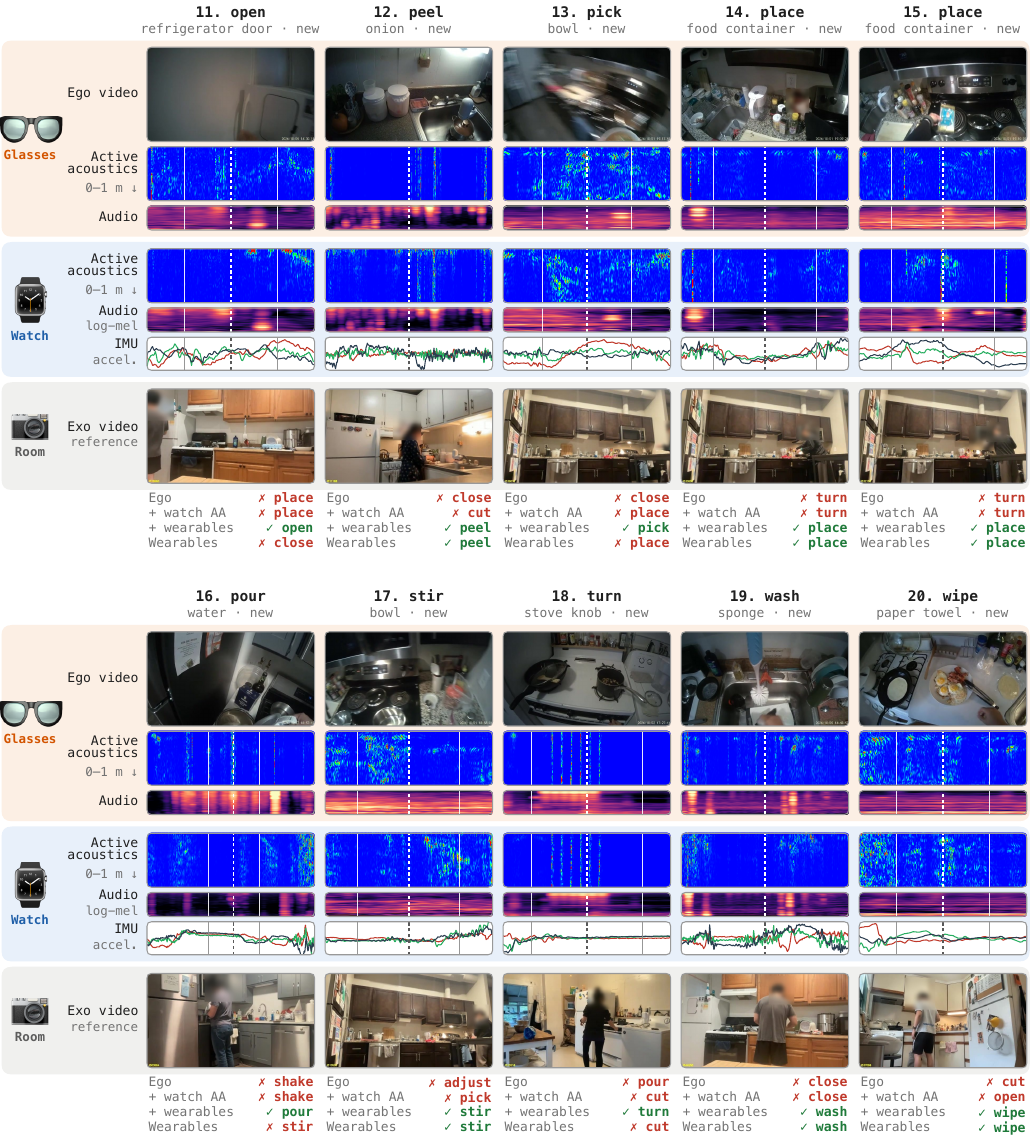}
\caption{\textbf{Examples 11--20: Hands out of view: egocentric video wrong, the wearables right.} See \cref{sec:suppl_examples}.}
\label{fig:suppl_examples_2}
\end{figure*}

\begin{figure*}[p]
\centering
\includegraphics[width=\textwidth]{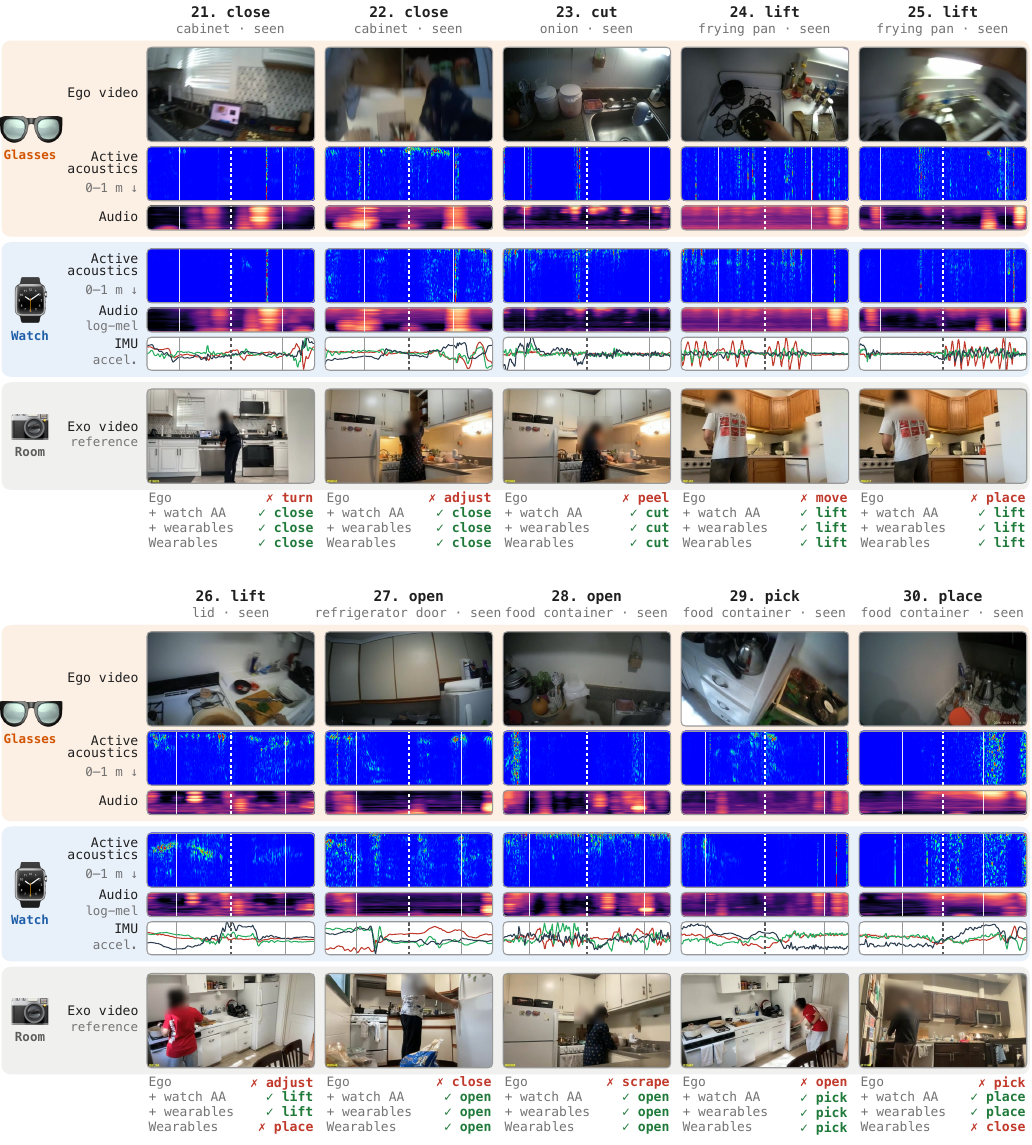}
\caption{\textbf{Examples 21--30: Active acoustics: egocentric video wrong, + watch AA right.} See \cref{sec:suppl_examples}.}
\label{fig:suppl_examples_3}
\label{fig:suppl_ex_aa}
\end{figure*}

\begin{figure*}[p]
\centering
\includegraphics[width=\textwidth]{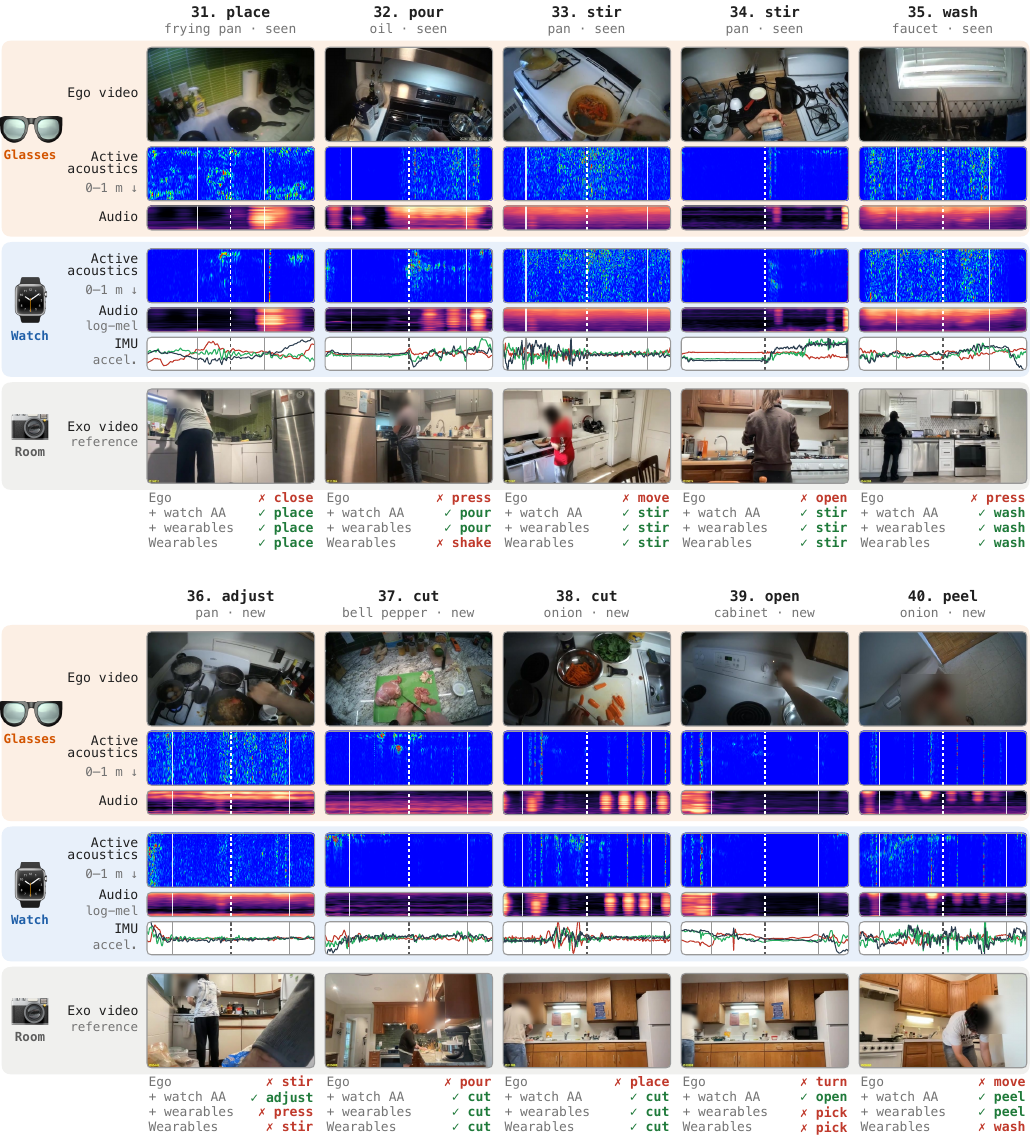}
\caption{\textbf{Examples 31--40: Active acoustics: egocentric video wrong, + watch AA right.} See \cref{sec:suppl_examples}.}
\label{fig:suppl_examples_4}
\end{figure*}

\begin{figure*}[p]
\centering
\includegraphics[width=\textwidth]{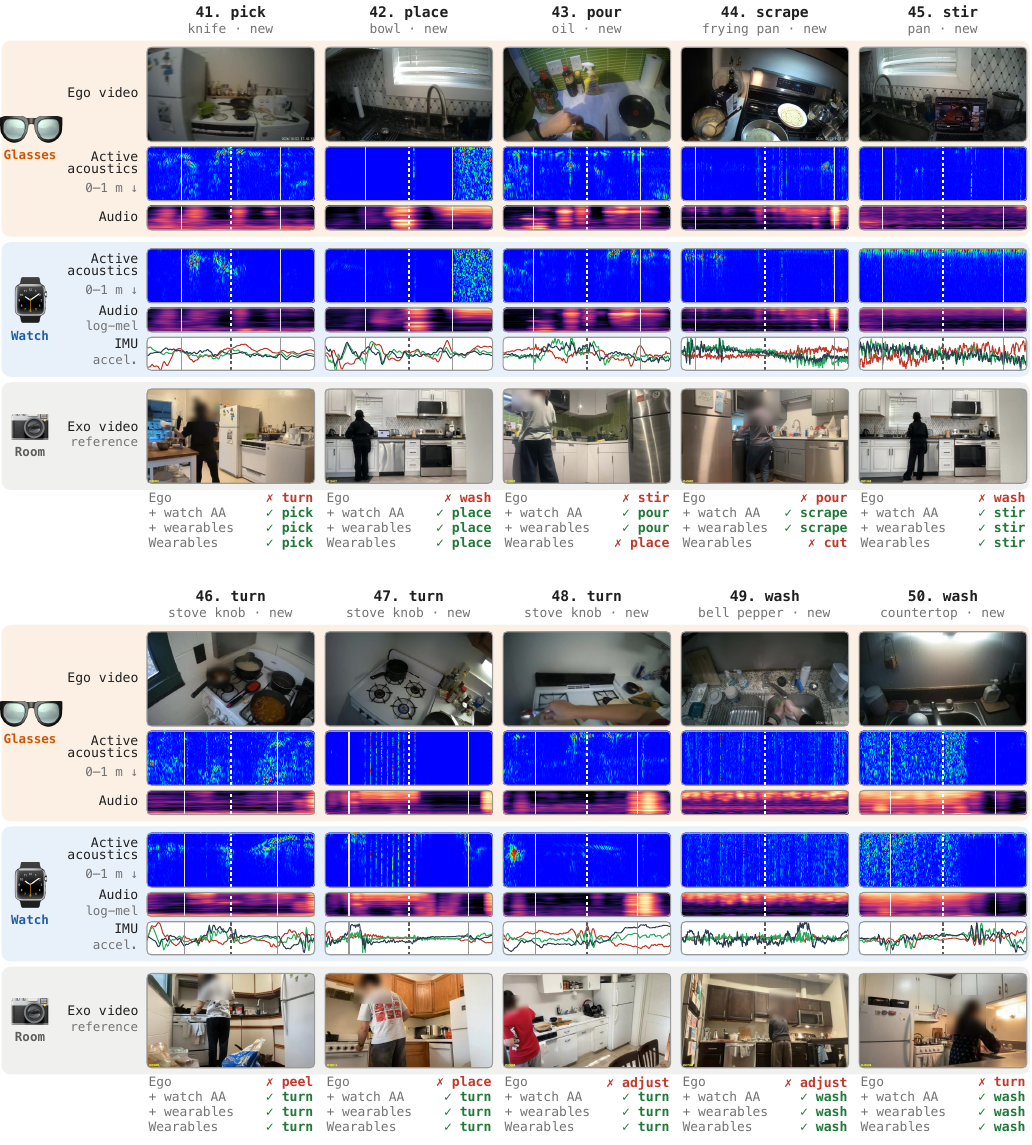}
\caption{\textbf{Examples 41--50: Active acoustics: egocentric video wrong, + watch AA right.} See \cref{sec:suppl_examples}.}
\label{fig:suppl_examples_5}
\end{figure*}

\begin{figure*}[p]
\centering
\includegraphics[width=\textwidth]{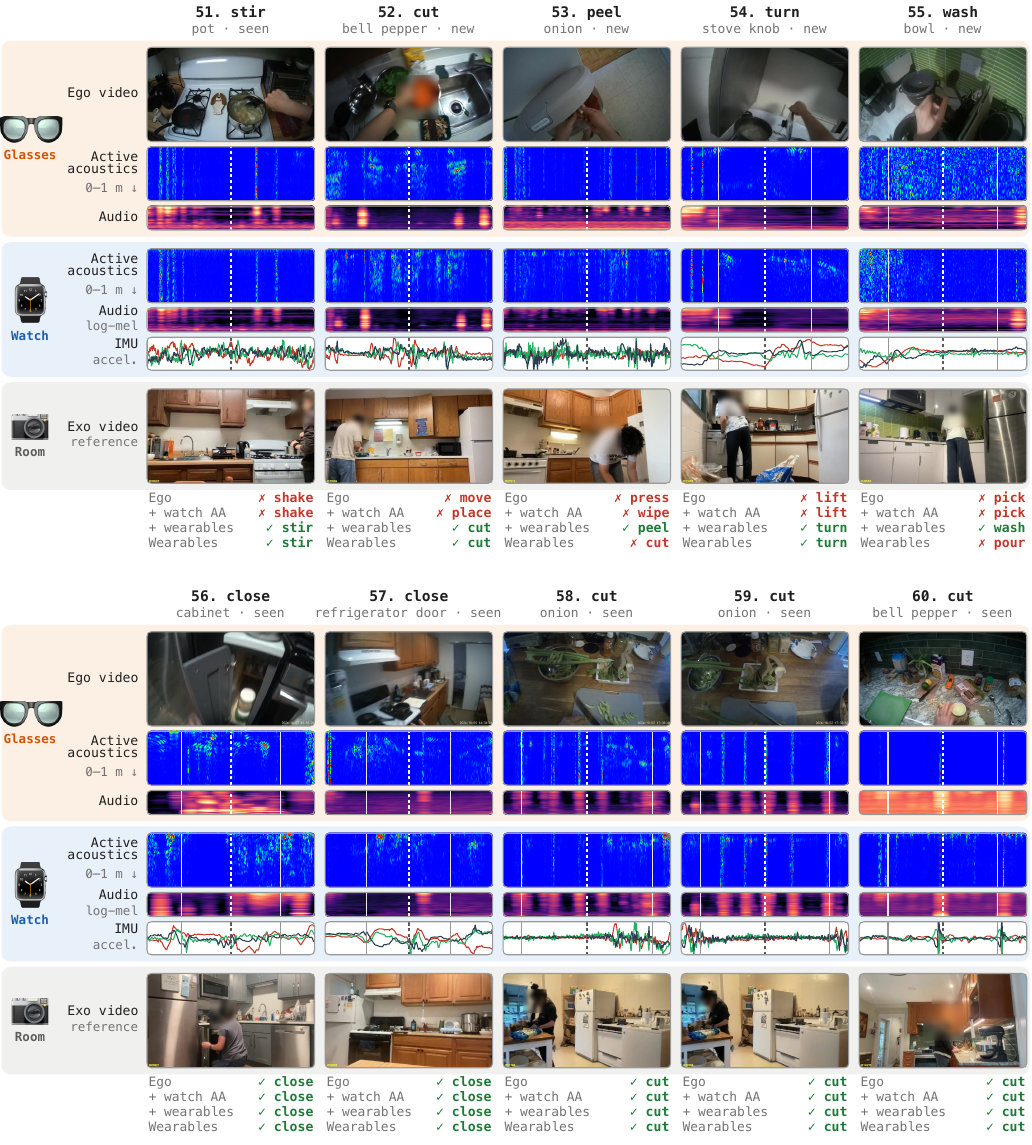}
\caption{\textbf{Examples 51--55: Hands in view: egocentric video wrong, the wearables right. Examples 56--60: Both right: egocentric video and the wearables.} See \cref{sec:suppl_examples}.}
\label{fig:suppl_examples_6}
\label{fig:suppl_ex_motion}
\label{fig:suppl_ex_both}
\end{figure*}

\begin{figure*}[p]
\centering
\includegraphics[width=\textwidth]{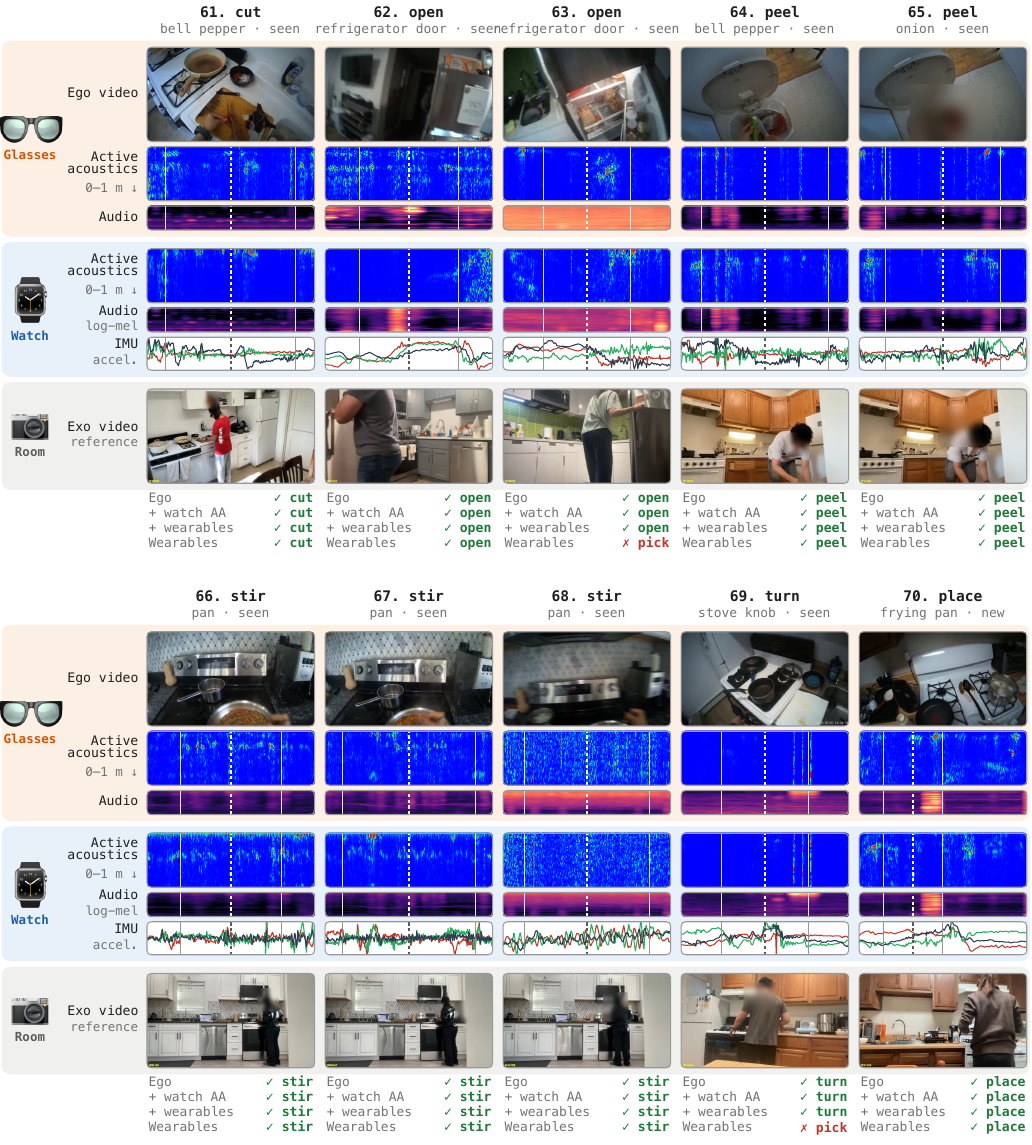}
\caption{\textbf{Examples 61--70: Both right: egocentric video and the wearables.} See \cref{sec:suppl_examples}.}
\label{fig:suppl_examples_7}
\end{figure*}

\begin{figure*}[p]
\centering
\includegraphics[width=\textwidth]{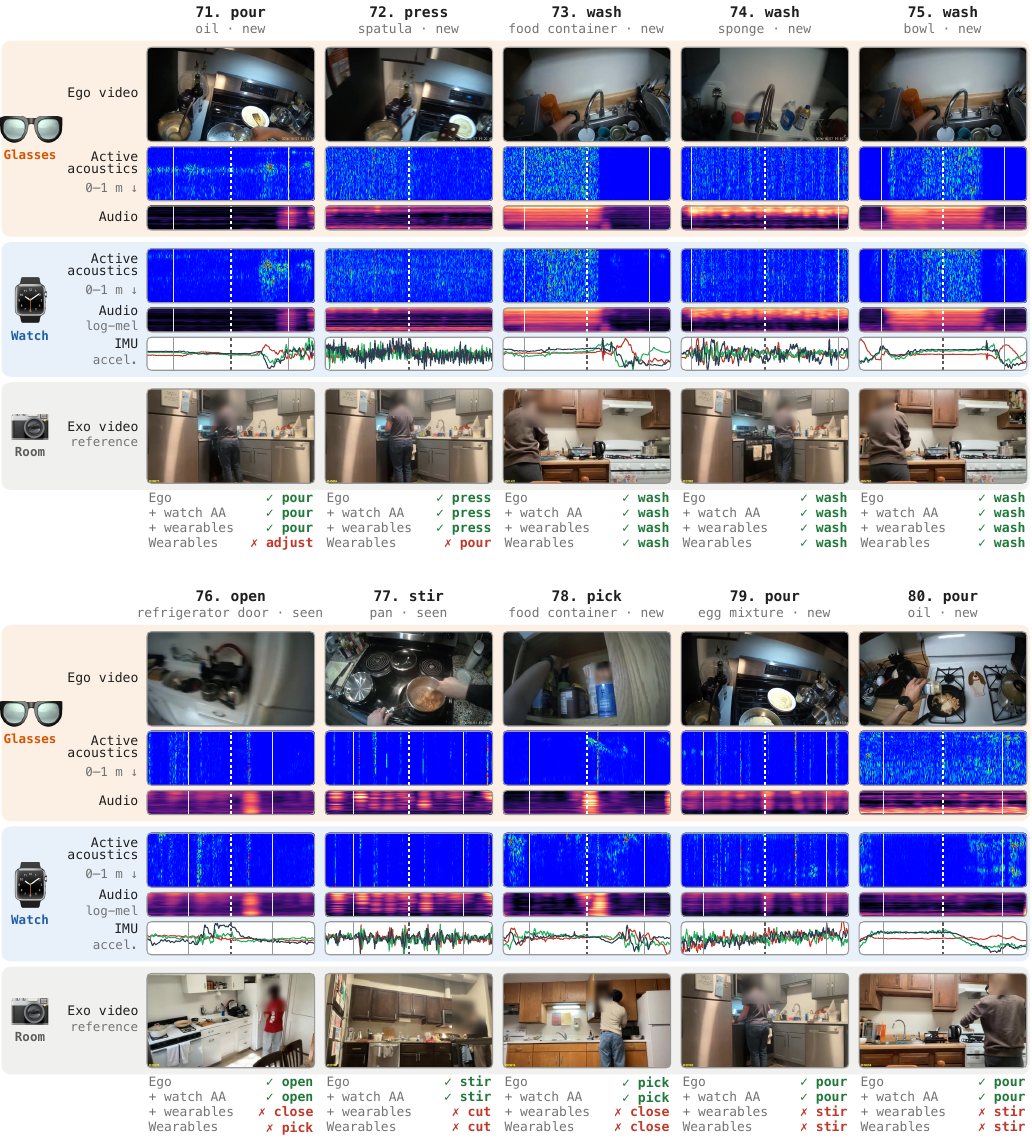}
\caption{\textbf{Examples 71--75: Both right: egocentric video and the wearables. Examples 76--80: Failure cases: egocentric video right, the wearables wrong.} See \cref{sec:suppl_examples}.}
\label{fig:suppl_examples_8}
\label{fig:suppl_ex_miss}
\end{figure*}

Eight pages (\cref{fig:suppl_examples_1,fig:suppl_examples_2,fig:suppl_examples_3,fig:suppl_examples_4,fig:suppl_examples_5,fig:suppl_examples_6,fig:suppl_examples_7,fig:suppl_examples_8}) show \ExamplesN{} test actions with every stream, from \ExamplesParticipants{} participants and \ExamplesVerbs{} verbs, \ExamplesNew{} of them by participants held out in training.
They show the paper's theme one action at a time: the camera sees what the objects are, and the echoes follow the hands whether or not the camera sees them.
\Cref{tab:suppl_examples} groups them by which setups name the verb, one group per case of \cref{sec:suppl_results}, failures included.
How often each case occurs is measured in \cref{sec:suppl_results}.

\begin{table}[t]
\centering
\caption{\textbf{Groups of examples}, by which setups name the verb: egocentric video (Ego), with the watch's active acoustics (+AA), and with the watch and the glasses (+W+G, ``+ wearables'' in the figures). \cmark: right; \xmark: wrong; --: either.}
\label{tab:suppl_examples}
\footnotesize
\setlength{\tabcolsep}{3pt}
\begin{tabular}{@{}l l c c c c@{}}
\toprule
Group & Examples (new) & Ego & +AA & +W+G & Hands seen \\
\midrule
Hands out of view & \ExamplesHandsRange{} (\ExamplesHandsNew) & \xmark & \xmark & \cmark & not always \\
Active acoustics & \ExamplesAaRange{} (\ExamplesAaNew) & \xmark & \cmark & -- & -- \\
Hands in view & \ExamplesMotionRange{} (\ExamplesMotionNew) & \xmark & \xmark & \cmark & always \\
Both right & \ExamplesBothRange{} (\ExamplesBothNew) & \cmark & -- & \cmark & -- \\
Failure cases & \ExamplesMissRange{} (\ExamplesMissNew) & \cmark & -- & \xmark & -- \\
\bottomrule
\end{tabular}
\end{table}

\paragraph{Layout.}
Each column is one action, with every stream around it, grouped by device: the glasses' egocentric frame, active acoustics and audio; the watch's active acoustics, audio and IMU; and the exocentric frame, for reference.
Echo profiles span 0 to 1\,m from the device, top to bottom, and time from left to right, so a moving hand leaves a trace and a still scene stays uniform.
Solid lines bound the action, and a dashed line marks the moment of the frames shown.
Below each column are the verbs that each setup names, the majority over the action's 1\,s windows, green when right.
Faces are blurred.

\paragraph{Reading a column.}
The egocentric frame shows what the camera had when it chose its verb: whether the hands were in view, and what objects were.
The echo panels below it show whether the hands moved during the action and where: a trace near the top of a panel is close to the device, and a regular pattern is a repeated stroke.
The audio and the IMU add what the action sounds like and how the wrist turned.
The labels then compare the setups, so each column reads as one decision made with and without the wearables.

\paragraph{What each stream shows.}
The streams make different actions visible.
Washing fills the echo profiles with speckle across every distance, the running water's sound in the chirp band (\cref{sec:suppl_signal}), and brightens the audio (Examples~73 to 75).
Cutting draws regular vertical stripes in the echo profiles and fast, even oscillation in the IMU (Examples~8, 58 and 59), and pressing with a spatula does the same in the IMU (Example~72).
Lifting a frying pan shows instead as a few large swings of the wrist (Examples~24 and 25).
Together, the streams cover what a single egocentric frame leaves open: how the hands move, and what the action sounds like.

\paragraph{Hands out of view (\cref{fig:suppl_examples_1,fig:suppl_examples_2}).}
The camera misses the hands and names a verb that fits the scene in view, often placing, picking or turning; the echoes still change with the hands, and the wearables name the right verb.
In Example~1, the frame shows the pan but not the hand on its handle, and egocentric video names turn where the wearables name adjust.
In Example~8, a new participant cuts an onion below the camera's view: egocentric video names press, while the watch's IMU and the glasses' echoes show the knife's regular strokes, and with the wearables the verb is cut.
This is where the wearables gain most (\cref{tab:suppl_visibility}).

\paragraph{Active acoustics (\cref{fig:suppl_examples_3,fig:suppl_examples_4,fig:suppl_examples_5}).}
Adding only the watch's active acoustics to egocentric video turns a wrong verb into the right one.
Its profile shows what tells the verbs apart: the regular strokes of cutting or stirring, the reach and withdrawal of placing, the swing of an arm closing a door.
In Examples~21 and 22, the camera sees a cabinet and names turn or adjust; with the watch's active acoustics, both become close.
In Examples~24 and 25, egocentric video names move and place for a frying pan being lifted, and the watch's active acoustics names lift.

\newpage   %
\paragraph{Hands in view (\cref{fig:suppl_ex_motion}).}
The camera sees the hands at a pan, a board or a sink, yet whether they stir, cut, peel or wash is in their motion, which the echoes and the IMU record 80 and 50 times a second.
In Example~51, the hands are in view at a pot, and egocentric video names shake; with the wearables, stir.
In Example~53, they peel an onion that egocentric video takes for pressing, and in Example~55 they hold a bowl in the sink: egocentric video names pick, and the wearables wash.

\paragraph{Both right (\cref{fig:suppl_examples_6,fig:suppl_examples_7,fig:suppl_examples_8}).}
From repetitive actions such as cutting to brief ones such as opening a door, the wearables keep what the camera already recognizes: adding them does not trade the camera's successes for new ones.
In Examples~58 and 59, the steady strokes of cutting appear in the IMU and in both devices' echoes, and every setup names cut.

\paragraph{Failure cases (\cref{fig:suppl_ex_miss}).}
Here the full system is wrong where egocentric video is right.
In every one of them, egocentric video with the watch's active acoustics is still right: the error comes from the other wearable streams naming a neighbouring action, which the equal-weight fusion follows.
In Examples~79 and 80, pouring into a pan, the wearables name stir, the other action performed at the same pan; in Examples~76 and 78, at a refrigerator door and a food container, they name close.
A fusion that weighs each sensor by its confidence in context is a direct way to improve on the baseline.

\clearpage   %

\end{document}